\documentclass[journal]{IEEEtran}
\usepackage{ifpdf}
\usepackage{cite}
\ifCLASSINFOpdf
   \usepackage[pdftex]{graphicx}
\else
\fi
\usepackage{amsmath}
\usepackage{algorithmic}
\usepackage{array}
\usepackage{url}
\newcommand{\draft}[1]{\textcolor{black}{#1}}

\usepackage{booktabs} % for professional tables
\usepackage{amsfonts,amssymb}
\usepackage{subfigure}
\usepackage{color,xcolor}
\usepackage{times}
\usepackage{epsfig}
\usepackage{amsmath}
\usepackage{textcomp}
\usepackage{gensymb}
\usepackage{wasysym}
\usepackage{multirow}
\usepackage{cite}
\usepackage{graphicx} % more modern

\usepackage{algorithm}
\usepackage{algorithmic}
\usepackage{mathrsfs}
\usepackage{url}
\usepackage{lettrine}

\begin{document}
\title{Interpreting and Improving Adversarial Robustness of Deep Neural Networks with Neuron Sensitivity}
%\textsuperscript{*}
\author{
Chongzhi Zhang\textsuperscript{*}, Aishan Liu\textsuperscript{*}, Xianglong~Liu\textsuperscript{\dag}, Yitao Xu, Hang Yu, Yuqing Ma, Tianlin Li

\thanks{* Equal contributions.}
\thanks{C. Zhang, A. Liu, X. Liu, Y. Xu, H. Yu, Y. Ma and T. Li are with the State Key Lab of Software Development Environment, Beihang University, Beijing, China. X. Liu is also with Beijing Advanced Innovation Center for Big Data-Based Precision Medicine, Beihang University, Beijing, China (\dag\ Corresponding author: Xianglong Liu, xlliu@nlsde.buaa.edu.cn).}
\thanks{This work was supported by National Key Research and Development Plan of China (2020AAA0103502), and National Natural Science Foundation of China (62022009 and 61872021).}
}


\markboth{IEEE Transactions on Image Processing,~Vol.~, No.~, December~2020}%
{Shell \MakeLowercase{\textit{Liu et al.}}: Training Robust Deep Neural Networks via Adversarial Noise Propagation}



\maketitle

\begin{abstract}
Deep neural networks (DNNs) are vulnerable to adversarial examples where inputs with imperceptible perturbations mislead DNNs to incorrect results. Despite the potential risk they bring, adversarial examples are also valuable for providing insights into the weakness and blind-spots of DNNs. Thus, the interpretability of a DNN in the adversarial setting aims to explain the rationale behind its decision-making process and makes deeper understanding which results in better practical applications. To address this issue, we try to explain adversarial robustness for deep models from a new perspective of neuron sensitivity which is measured by neuron behavior variation intensity against benign and adversarial examples. In this paper, we first draw the close connection between adversarial robustness and neuron sensitivities, as sensitive neurons make the most non-trivial contributions to model predictions in the adversarial setting. Based on that, we further propose to improve adversarial robustness by stabilizing the behaviors of sensitive neurons. %improve adversarial robustness by constraining the similarities of sensitive neurons between benign and adversarial examples which stabilizes the behaviors of sensitive neurons towards adversarial noises.
Moreover, we demonstrate that state-of-the-art adversarial training methods improve model robustness by reducing neuron sensitivities, which in turn confirms the strong connections between adversarial robustness and neuron sensitivity. Extensive experiments on various datasets demonstrate that our algorithm effectively achieves excellent results. To the best of our knowledge, we are the first to study adversarial robustness using neuron sensitivities.
\end{abstract}


% \fi
\IEEEpeerreviewmaketitle

\section{Introduction}
\IEEEPARstart{R}{cently}, Deep Neural Network (DNNs) have demonstrated remarkable performance in a wide spectrum of areas, including computer vision \cite{DBLP:conf/nips/KrizhevskySH12,chi2017dual,
cai2018learning,fu2019stacked,8902220}, speech recognition \cite{DBLP:journals/taslp/MohamedDH12} and  natural language processing \cite{chen2014fast,DBLP:conf/nips/SutskeverVL14}. Despite the significant achievements, unfortunately, the emergence of adversarial examples \cite{DBLP:journals/corr/GoodfellowSS14,carlini2017towards,dong2018boosting,guo2019simple,8883191}, images containing small perturbations imperceptible to human but extremely misleading to DNNs, casts a cloud over the recent progress of deep learning. Further, adversarial examples also pose potential security threats by attacking or misleading the practical deep learning applications like auto-driving and face recognition system, which may cause pecuniary loss or even people death with severe impairment \cite{DBLP:conf/iclr/KurakinGB17a,Liu2019Perceptual,DBLP:conf/woot/SongEEF0RTPK18}.

\begin{figure}[!htb]
	\centering
	\includegraphics[width=1.0\linewidth]{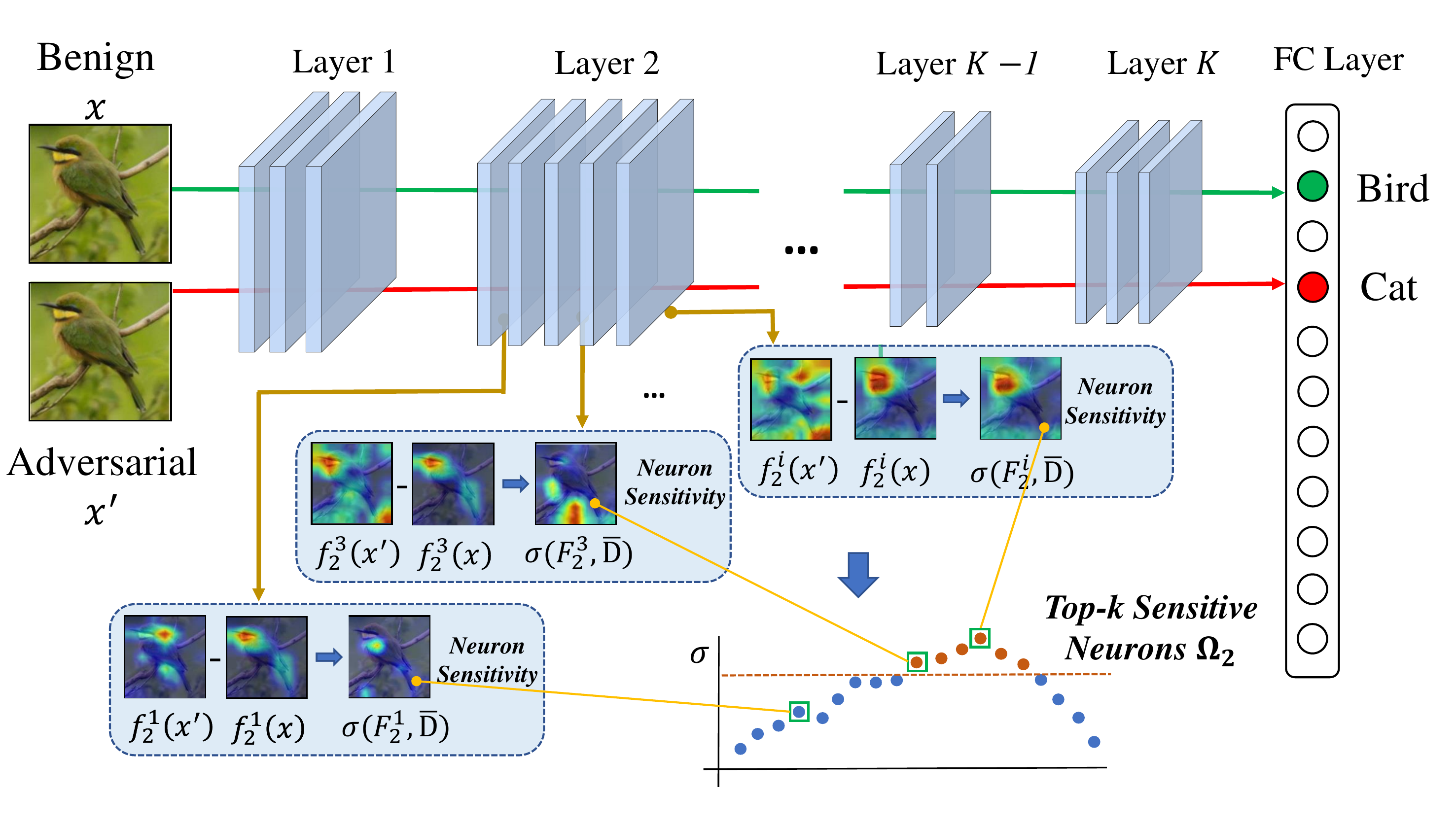}
	\caption{Our framework of computing \emph{Neuron Sensitivity} and selecting \emph{Sensitive Neurons}. \emph{Neuron Sensitivity} is measured by neuron behavior variation intensities against benign example $x$ and the corresponding adversarial example $x^{\prime}$. Neurons with the top-$k$ maximum \emph{Neuron Sensitivity} on specific layers will be selected as \emph{Sensitive Neurons}.}
	\label{fig:illustration of Sensitivity}
\end{figure}

In the past years, plenty of defense methods have been proposed to improve model robustness to adversarial examples, avoiding the potential danger in real world applications. These methods can be roughly categorized into adversarial training \cite{DBLP:journals/corr/GoodfellowSS14,DBLP:conf/iclr/KurakinGB17,DBLP:conf/iclr/MadryMSTV18}, input transformation  \cite{7839189,DBLP:conf/iclr/GuoRCM18,DBLP:conf/iclr/XieWZRY18,8365806}, elaborately designed model architectures \cite{DBLP:conf/sp/PapernotM0JS16,liao2018defense,liu2019training} and adversarial example detection \cite{DBLP:conf/iccv/LuIF17,DBLP:conf/iclr/MetzenGFB17,roth2019odds,DBLP:conf/ndss/Xu0Q18}. Though challenging deep learning, from another point of view, adversarial examples are also valuable and beneficial for understanding the behaviors of DNNs. Due to the myriad of linear and nonlinear operations in the black-box model, interpreting deep learning is an extremely difficult problem in the literature. Thus, adversarial attacks provide us with a new way to explore model weakness and blind-spots, which are valuable to understand their behaviors in the adversarial setting. Further, we can improve model robustness and build stronger models against noises. Several works have been proposed to study model robustness against adversarial examples from the views of denoising and activation scaling \cite{liao2018defense,DBLP:journals/corr/abs-1904-02057}.

However, these methods mainly focused on the pixel space of input images. Few works have focused on analyzing the influence of adversarial perturbations via investigating the behaviors of the model's intermediate layers, which make significant contributions to model robustness and performance.
From a high-level perspective, model robustness to noises can be viewed as a kind of global insensitivity property \cite{DBLP:conf/iclr/TsiprasSETM19}. A deep model can learn insensitive representations towards adversarial examples if the intermediate layers (\emph{i.e.,} neurons) behave stably without too much performance degeneration when encountering such noises. %Thus, by constraining the differences between clean and adversarial pairs in the penultimate or the logit layer of a DNN, several studies have been proposed to defense adversarial examples \cite{DBLP:journals/corr/abs-1803-06373,liao2018defense}. %\aishan{The difference between other works and ours is: ``investigating xxxx intermediate layers''? Sth else? Why this is useful? Why using intermediate layer is better? I think we need more motivations.}
Thus, this paper first tries to interpret model robustness in the adversarial setting by analyzing behavior sensitivities in a neuron-wise way. By studying neuron behaviors in layers sequentially, we reveal insightful clues for model robustness and weakness, which in turn motivate us to introduce a strategy to improve model robustness via stabilizing neuron sensitivities.

Our main contributions can be summarized as follows:
\begin{itemize}
\item \textcolor{black}{We introduce the concept of Neuron Sensitivity that considers the changing intensity of neuron behaviors for adversarial and benign examples. And we are the first to use Neuron Sensitivity as a criterion to measure the stability of a DNN in adversarial settings.}
\item Further, we take the first steps to define \emph{Sensitive Neuron}, a sequence of neurons most sensitive to adversarial examples, which we believe may conduct the most non-trivial contributions to sensitive model behaviors.
\item By stabilizing neuron sensitivities towards benign and adversarial examples, we propose the \emph{Sensitive Neuron Stabilizing} (SNS) method to improve model robustness against adversarial noises. Extensive experiments on CIFAR-10 and ImageNet empirically demonstrate that such a simple technique significantly outperforms state-of-the-art adversarial training strategies.
\item Empirical studies on neuron sensitivities in different layers of different deep models demonstrate that state-of-the-art adversarial training methods improve model robustness mainly by embedding insensitivity to neurons which in turn confirms the significance of neuron sensitivities towards adversarial robustness.
\end{itemize}

The structure of the paper is as follows. Section \ref{Section:related work} introduces the related works. Section \ref{Section:concept} defines the notion of neuron sensitivity and sensitive neuron. Section \ref{Section:Sensitive Neurons and Adversarial Robustness} provides empirical analyses of the strong connections between sensitive neurons and adversarial robustness. Section \ref{Section:Improving Robustness with Sensitive Neurons} explores the reasons why state-of-the-art adversarial training strategies achieve strong robustness from the view of sensitive neurons and propose \emph{Sensitive Neuron Stabilizing} (SNS) for improving model robustness. Section \ref{Section:conclusion} summarizes the whole contributions again and has further discussions on them.

%\aishan{Do we need to add a `related work' section to specify the work that tried to explain and interpret adversarial examples?}

\section{Related Work} \label{Section:related work}
Adversarial examples was first proposed in \cite{DBLP:journals/corr/SzegedyZSBEGF13}. With imperceptible perturbations, an adversarial example $x^\prime$ looks similar to its clean counterpart $x$, but they are able to mislead model $F$:
\begin{align*}
F(x^\prime) \neq y  \quad s.t. \quad \|x-x^\prime\| < \epsilon.
\end{align*}

%\cite{DBLP:journals/corr/GoodfellowSS14,carlini2017towards,dong2018boosting,guo2019simple}
In the past years, great efforts have been devoted to generate adversarial examples in different scenarios \cite{DBLP:journals/corr/GoodfellowSS14,carlini2017towards,DBLP:conf/iclr/MadryMSTV18,dong2018boosting,chen2020boosting,Liu2020Spatiotemporal,Liu2020Biasbased}.

Beyond attacks, a line of work have been proposed to interpret model robustness towards adversarial noises. Understanding adversarial examples provide insights into the weaknesses and blind-spots of DNNs in adversarial settings, which in turn offers us clues to further build robust deep learning models. Dong \emph{et al}. \cite{DBLP:journals/corr/abs-1708-05493} re-examined the internal representations of DNNs using adversarial examples and further improved their interpretability with an adversarial training scheme. Besides, Xu \emph{et al.} \cite{DBLP:journals/corr/abs-1904-02057} attempted to explore the weakness of models under adversarial conditions by analyzing the effects of different regions within a specific image. More recently, Ilyas \emph{et al.} \cite{DBLP:journals/corr/abs-1905-02175} believed that robust features can be extracted with the help of adversarially robust deep models. Wang \emph{et al}. \cite{wang2018interpret} proposed a distillation guided routing method to discover critical data routing paths (CDRPs) in the neural network. The proposed CDRPs mainly preserve the performance and can be applied to adversarial detection problem. This finding demonstrates the close relationship between model performance and adversarial robustness. \draft{In \cite{wang2020interpret}, they further extended the concept of CDRPs and applied it to the class level as critical subnetworks. The novel perspective enabled the investigation of model's layerwise semantic behavior and more accurate visual explanations appearing in the data. Furthermore, based on the properties of sample-specific and class-specific subnetworks, two adversarial example detection methods were proposed. Besides, Hierarchical Attribution Fusion (HAF) technique was also introduced to learn reliable visual explanation saliency in \cite{wang2019learning}. Based on the optimized hierarchical attribution masks, a novel adversarial example detection was proposed.} Rouhani \emph{et al.} \cite{rouhani2017curtail} performed a thorough sensitivity analysis based on a layer-wise spectrum density evaluation of pertinent model parameters during their training phase. After the global flow of whole framework, the legitimacy probability of the input data will be given which can be used to detect adversarial examples. Ma \emph{et al.} \cite{ma2018characterizing} characterized the dimensional properties of adversarial regions via Local Intrinsic Dimensionality (LID). LID assesses the space-filling capability of the region surrounding a reference example, based on the distance distribution of the example to its neighbors. They showed that a potential application of LID is to distinguish adversarial examples. \textcolor{black}{Furthermore, another line of works tried to interpret deep learning models by performing sensitivity analysis \cite{DBLP:journals/dsp/MontavonSM18,sung1998ranking,khan2001classification}. Sensitivity analysis is a feasible approach to identify the most important part of input features \cite{DBLP:journals/dsp/MontavonSM18}. It is based on the model's locally evaluated gradients \cite{sung1998ranking,khan2001classification} or some other local measurements. The heatmaps built by these methods provide a local scope of explanation of the neural network.}
%However, these methods either mainly focused on the pixel space of input images without investigating intermediate layers or failed to have an insight into how the intermediate layers gradually contribute to the misclassification towards adversarial attacks.

\textcolor{black}{In contrast to \cite{wang2018interpret}, the sensitive neurons are selected based on our proposed metric neuron sensitivity by considering the varying intensity of neuron behaviors for adversarial and benign examples. Besides, as for \cite{rouhani2017curtail} and \cite{ma2018characterizing}, our differences are two-folded. Firstly, though we all use internal representations, the calculation methods are entirely different. In \cite{rouhani2017curtail}, they directly use the Spectral Energy Factor preserved by the first Eigenvalue of the model parameters in a layer-wise way. In \cite{ma2018characterizing}, they calculate the Local Intrinsic Dimensionality as the characterization of adversarial regions. Different from them, we use Neuron Sensitivity and Sensitivity Ratio based on the $\ell_1$ norm of discrepancy between adversarial and benign examples in a neuron-wise way. Secondly, we differ in the purpose and motivation. Both \cite{rouhani2017curtail} and \cite{ma2018characterizing} use their criterion to characterize the unique properties of adversarial examples for adversarial detection. However, we concentrate on using Neuron Sensitivity to probe model's flaws (Sensitive Neurons) and further improve adversarial robustness.}

\section{Neuron Sensitivity and Sensitive Neuron} \label{Section:concept}
Prior studies have shown that different neurons play different roles and possess different importance for model prediction even arranged in the same layer \cite{bau2017network,DBLP:journals/corr/abs-1708-05493,DBLP:journals/corr/ZhouKLOT14}. Inspired by this fact, this section studies the adversarial robustness from the view of neuron behaviors.

Given a dataset $\mathbf{D}$ with data sample $x$ $\in$ $\mathcal{X}$ and
label $y$ $\in$ $\mathcal{Y}$, % the corresponding class label.
the deep supervised learning model tries to learn a mapping or classification function $F$: $\mathcal{X}$ $\rightarrow$ $\mathcal{Y}$. The model $F$ consists of $\mathit{L}$ serial layers. For the $l$-th layer $F_l$, where $l=1,\ldots,L$, it contains several neurons, which can also be regarded as a neuron set. We use superscript $m$ to denote the $m$-th neuron, and satisfying $F_l^m \in F_l$. The output of one neuron $F_l^m$ is equivalent to the $m$-th channel of the feature map produced by layer $l$. For model $F$, this paper chooses the popular deep convolutional neural networks (CNNs) for the visual recognition task.

\subsection{Neuron Sensitivity}
The model robustness towards noises can be viewed as a global insensitive behavior showing small losses and consistent predictions under noise conditions. Recall the definition of model robustness to noise in \cite{xu2012robustness}, which are derived from the idea that if two instances are ``similar'' then their test errors are close, too:
\begin{align*}
\forall x_i,x_j \in \mathbf{D}, \  if \  ||x_i-x_j|| < \epsilon  \Rightarrow \Vert \mathcal{L}_F(x_i)-\mathcal{L}_F(x_j) \Vert \leq e,
\end{align*}
where $x_i$ and $x_j$ are samples selected randomly from the same dataset $\mathbf{D}$ and $\mathcal{L}_F(\cdot)$ denotes the loss function. $\|\cdot\|$ is a distance metric to quantify the distance between samples and $e$ denotes a small value.

This fact should hold for the benign sample $x\in \mathbf{D}$  from category $y$ and its adversarial example $x^\prime\in \mathbf{D}^{\prime}$. However, in practice, adversarial examples mislead the non-robust classifier to predict the wrong label.

Intuitively, for a model that owns strong robustness, namely, insensitive to adversarial examples, we expect that the benign sample $x$ and the corresponding adversarial example $x^\prime$ share a similar representation in the hidden layers of the model, leading to similar final predictions as well. Motivated by this intuition, to understand the adversarial robustness of deep models, one can concentrate on the deviation of the feature representation in hidden layers between benign samples and corresponding adversarial examples.

To achieve this goal, we introduce \emph{Neuron Sensitivity} to  interpret the model sensitivity from the view of neurons inside it. Specifically, given a benign example $x_i$, where $i=1,\ldots, \mathit{N}$, from $\mathbf{D}$ and its corresponding adversarial example $x^\prime_i$ from $\mathbf{D}^{\prime}$, we can get the dual pair set $\bar{\mathbf{D}}=\{(x_i,x^{\prime}_i)\}$, and then calculate the neuron sensitivity $\sigma$ as follow:
\begin{equation}
\sigma(F_l^m,\bar{\mathbf{D}})=\frac{1}{\mathit{N}}\sum_{i=1}^\mathit{N}\frac{1}{dim(F_l^m(x_i))}\Vert F_l^m(x_i)-F_l^m(x_i^\prime)\Vert_1 ,\label{Neuron Sensitivity}
\end{equation}
where $F_l^m(x_i)$ and $F_l^m(x_i^\prime)$ respectively represents outputs of neuron $F_l^m$ towards benign sample $x_i$ and corresponding adversarial example $x_i^\prime$ during the forward process. $dim(\cdot)$ denotes the dimension of a vector.

\subsection{Sensitive Neuron}
Once we have defined \emph{Neuron Sensitivity}, we can further determine the most prominent neurons under this criterion and mark them as the \emph{Sensitive Neuron} $\Omega_l$ as shown below:
\begin{equation}
\Omega_l = \text{top-$k$}(F_{l}, \sigma),
\end{equation}
where top-$k$($\cdot$) represents top $k$ maximum instances of the input set based on a certain metric, such as neuron sensitivity $\sigma$ in this case, which means $\Omega_l$ is a subset of $F_l$.\
This can be easily accomplished by traversing the neurons in each layer and selecting $k$ neurons with the maximal neuron sensitivity for $\mathit{N}$ samples according to Equation \ref{Neuron Sensitivity}.

Obviously, sensitive neurons in each layer correspond to the vulnerable ones in a model towards adversarial examples, where more attention should be paid. Therefore, we use sensitive neurons to understand model behaviors in the adversarial setting for the rest of this paper. Figure \ref{fig:illustration of Sensitivity} demonstrates the basic procedure of computing neuron sensitivity and selecting sensitive neurons.

\section{Sensitive Neurons and Adversarial Robustness} \label{Section:Sensitive Neurons and Adversarial Robustness}
Apart from viewing DNNs as a black-box from a high-level viewpoint, it is natural for us to treat the model as a white-box and make deeper insights into model adversarial weaknesses from the perspective of sensitive neurons. In this section, we first explore the strong connections between sensitive neurons and adversarial robustness. With several empirical analyses, we surprisingly find that sensitive neurons make the most non-trivial contributions towards model robustness in the adversarial setting.

\subsection{Empirical Settings} \label{section:first setting}
\subsubsection{Datasets and models}

For empirical analyses, we adopt the widely used \textbf{CIFAR-10} \cite{krizhevsky2009learning} and \textbf{ImageNet} \cite{DBLP:journals/ijcv/RussakovskyDSKS15} datasets. CIFAR-10 consists of 60K natural scene color images with 10 classes of size $32 \times 32 \times 3 $. We use VGG-16 \cite{DBLP:journals/corr/SimonyanZ14a} \textcolor{black}{and Inception-V3 \cite{szegedy2016rethinking}} for CIFAR-10. ImageNet contains 14M images with more than 20k classes. For simplicity, we only choose 200 classes from 1000 in ILSVRC-2012 with 100K and 10k images for training set and validation set, respectively. The model we use for ImageNet is ResNet-18 \cite{DBLP:conf/cvpr/HeZRS16}.

Here, we further illustrate the name for each intermediate layer. For CIFAR-10, VGG-16 contains 13 convolutional layers and 1 full connected layer. Thus we respectively name them \emph{conv1-conv13} and \emph{fc1}. In contrast, \textcolor{black}{Inception-V3 and} ResNet-18 have more hierarchical architectures. \textcolor{black}{For Inception-V3 on CIFAR-10, we choose the implementation in torchvision\footnote{https://github.com/pytorch/vision/blob/master/torchvision/models/inception.py}. Specifically, it has 5 bottom convolutional layers named \emph{conv1-conv5} with 5 diverse inception layers followed by. The inception layers respectively contain 3, 1, 4, 1 and 2 serial inception blocks and
the block architecture differ in different layers. Here we use \emph{l1b1} to denote the first block of the first inception layer. The model also has 1 fully connected layer named \emph{fc1}. Unlike Inception-V3, ResNet-18 only uses one kind of basic architecture called \emph{basic block}.} Except for the bottom separated convolutional layer named \emph{conv1}, ResNet-18 has 4 layers and each of them contains 2 basic blocks with 2 convolutional layers. Hence we name them according to the hierarchy. In particular, \emph{l1b1c1} denotes the first layer, the first block, and the first convolutional layer. ResNet-18 only has 1 fully connected layer named \emph{fc1}.

\subsubsection{Adversarial Attack and Corruption}
We apply a diverse set of state-of-the-art \textbf{adversarial attack methods} \draft{which can be divided into two categories. The \textbf{white-box attacks} include} FGSM \cite{DBLP:journals/corr/GoodfellowSS14}, $\ell_2$ PGD, $\ell_\infty$ PGD \cite{DBLP:conf/iclr/MadryMSTV18} and C\&W attack \cite{carlini2017towards}. \draft{Apart from them, the \textbf{black-box attacks} are also applied including SPSA \cite{DBLP:conf/icml/UesatoOKO18} and NAttack \cite{DBLP:conf/icml/LiLWZG19}.} Parallel to that, we utilize the \textbf{corruption} \cite{DBLP:conf/iclr/HendrycksD19} to make further exploration. For adversarial attacks, we follow \textcolor{black}{the setting in} \cite{DBLP:journals/corr/GoodfellowSS14,DBLP:conf/iclr/MadryMSTV18} and \cite{carlini2017towards}. For corruption, we follow \textcolor{black}{the settings in} \cite{DBLP:conf/iclr/HendrycksD19}. The implementation details of these methods are shown as follows:

\begin{itemize}
\item \textbf{FGSM.}\quad In Section \ref{section:strong information} for analysis on CIFAR-10, we set $\epsilon = 2$. Afterwards, we assign $\epsilon = 8$ for testing adversarial robustness  for both CIFAR-10 and ImageNet datasets in Section \ref{section:sns}.
\item \textbf{$\ell_2$ PGD.}\quad In practice, we set $\epsilon = 80$, iteration $k=10$ and step size $\alpha = \epsilon / \sqrt{k}$ for analysis on CIFAR-10 in Section \ref{section:strong information}.
\item \textbf{$\ell_\infty$ PGD.}\quad In practice, we set $\epsilon = 2$, iteration $k=10$ and step size $\alpha = \epsilon / \sqrt{k}$ for analysis on CIFAR-10 in Section \ref{section:strong information}. In other cases, $\epsilon$ is fixed to 8 \textcolor{black}{unless otherwise stated}.
\item \textbf{C\&W.}\quad We utilize the $\ell_2$ form attack in experiments. We set initial const $c=0.01$ and confidence $k=0$. Besides, we perform 3 iterations of binary search over $c$ and run 300 iterations at each step.
\item \draft{\textbf{SPSA.}\quad We set perturbation size $\epsilon=8$, batch size $b=512$, learning rate $\eta=0.01$ and run 100 iterations at each step.}
\item \draft{\textbf{NAttack.}\quad We set perturbation size $\epsilon=8$, sample size $b=300$, initial mean $\mu_0=0$, standard deviation $\sigma^2=0.01$, learning rate $\eta=0.008$ and maximum number of iterations $T=100$.}
\item \textbf{Gaussian Noise}\quad We use the corresponding noises in CIFAR-10-C dataset proposed by \cite{DBLP:conf/iclr/HendrycksD19}. The corruption contains samples with 5 severities.
\end{itemize}

\subsection{Sensitive Neurons Contribute Most to Model Misclassification in the Adversarial Setting}\label{Section:Jaccard}
We first analyze the behaviors and contributions of sensitive neurons to model misclassification in adversarial setting. In this part, we utilize the targeted attack to control the model prediction and specifically, $\ell_\infty$ PGD method is applied. we generate adversarial examples on CIFAR-10 and ImageNet validation sets. On CIFAR-10, we conduct the experiments on whole validation set with VGG-16. \draft{Additionally, we also pick all the 200 classes from the standard ImageNet validation set with ResNet-18.} For each class ${y}$, we select $\mathbf{D}^y = \{x_i | x_i \in \mathbf{D} \wedge y_i \neq y \}$ from the validation set $\mathbf{D}$ for experiment. Then, we generate targeted adversarial set $\bar{\mathbf{D}}^y = \{x_i^\prime|x_i \in \mathbf{D}^y \wedge F(x_i^\prime)=y\}$ for the following analyses.

Since the linear layer can be viewed as a weighted sum process, we can easily calculate contributions of the penultimate layer $F_{\textrm{L}-1}$ to logit. Specifically, for a input adversarial example $x^\prime$ with prediction $y$, we measure the direct contribution of $F_{\textrm{L}-1}^{m}$ to target class as below:
\begin{equation}
\varphi(F_{\textrm{L}-1}^{m},x^{\prime},y) = F_{\textrm{L}-1}^{m}(x^{\prime})\cdot W_{m,y},
\label{equa:target_logit_contribution}
\end{equation}
where $W_{m,y}$ is the linear layer mapping from $m$-th representations to $y$-th logit.

On the basis of above metric, we further select a prominent portion of neurons as \emph{Important Neuron} $\Pi$ towards $x^\prime$:
\begin{equation}
\Pi^{y}(x^{\prime})=\text{top-$k$}(F_{\textrm{L}-1}, \varphi).
\label{equa:top_k important neurons}
\end{equation}

Obviously, these important neurons make the most important contributions to the model misclassification for adversarial example $x^{\prime}$.

As important neurons are designed for single adversarial image, we further extend the idea from one adversarial example image to one adversarial example set $\mathbf{D^{\prime}}$ for a specific target class $y$:
\begin{equation}
\Gamma^{y} = \text{top-$k$}(F_{\textrm{L}-1}, \rho),
\label{Most Contributive Neurons}
\end{equation}
where metric $\rho$ means the weighted voting method using important neurons $\Pi^{y}(x_1),...,\Pi^{y}(x_{\mathit{N}})$ computed with respect to each adversarial image in $\mathbf{D^{\prime}}$. Specifically, an Important Neuron sequence $\Pi^{y}(x_i)$ contains $k$ neurons ranked in descending order by metric $\varphi$. To emphasize their different importances, we give them different votes from $k$ to 1. After the voting process, votes for each neuron will be counted and the $k$ highest of them will be selected. In practice, we investigate the neurons from \emph{pool5} layer for VGG-16 model and \emph{global average pooling} layer for ResNet-18. Besides, we set $k=20$.

\draft{To quantify the correlations between neuron importance and sensitivity, we first exploit the Spearman's Rank Correlation Coefficient. This measurement aims to quantify the statistical dependence between the rankings of two variables, which takes both the overlap and the ranking order of the elements into consideration. Specifically, we select all the neurons at the penultimate layer and label them from $1$ to $n$ in order. We then respectively sort the neuron list by neuron sensitivity and importance in the descending order, respectively. For each neuron $i$, we use $I_i^{\sigma}$ and $I_i^{\varphi}$ to respectively denote the new ranking index in lists sorted by neuron sensitivity and importance. Thus, we can calculate the similarity between neuron importance and sensitivity via Spearman's Rank Correlation Coefficient as follows:}

\begin{equation}
\begin{small}
r_s(\sigma,\varphi) = 1-\frac{6\sum (I_i^{\sigma}-I_i^{\varphi})^2}{n^3-n}.
\end{small}
\label{Spearman rank correlation}
\end{equation}

\begin{figure*}[!tbp]
\centering
\begin{minipage}[t]{0.48\linewidth}
\centering
\subfigure[]{
\includegraphics[width=0.46\linewidth]{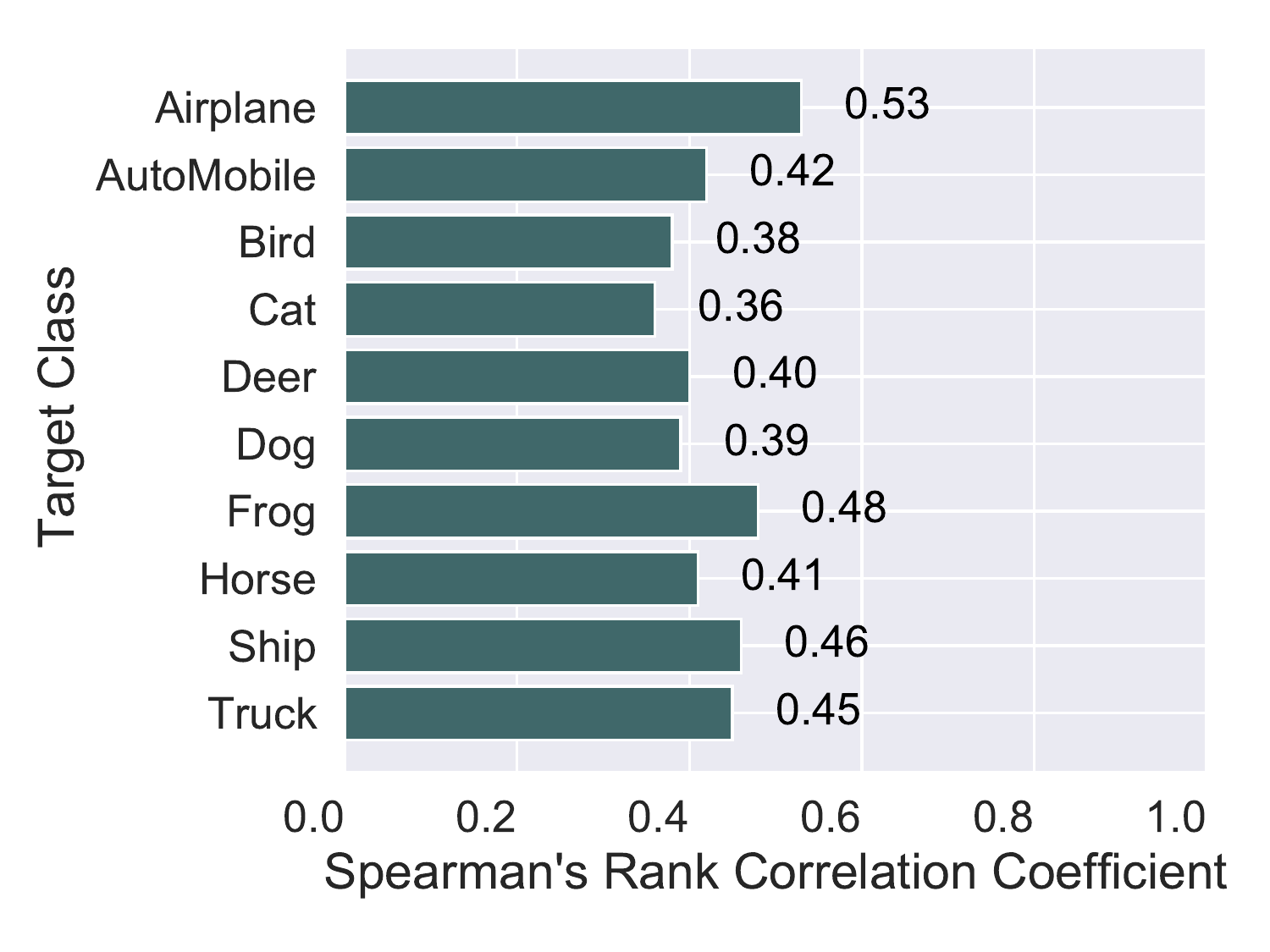}
}
\subfigure[]{
\includegraphics[width=0.46\linewidth]{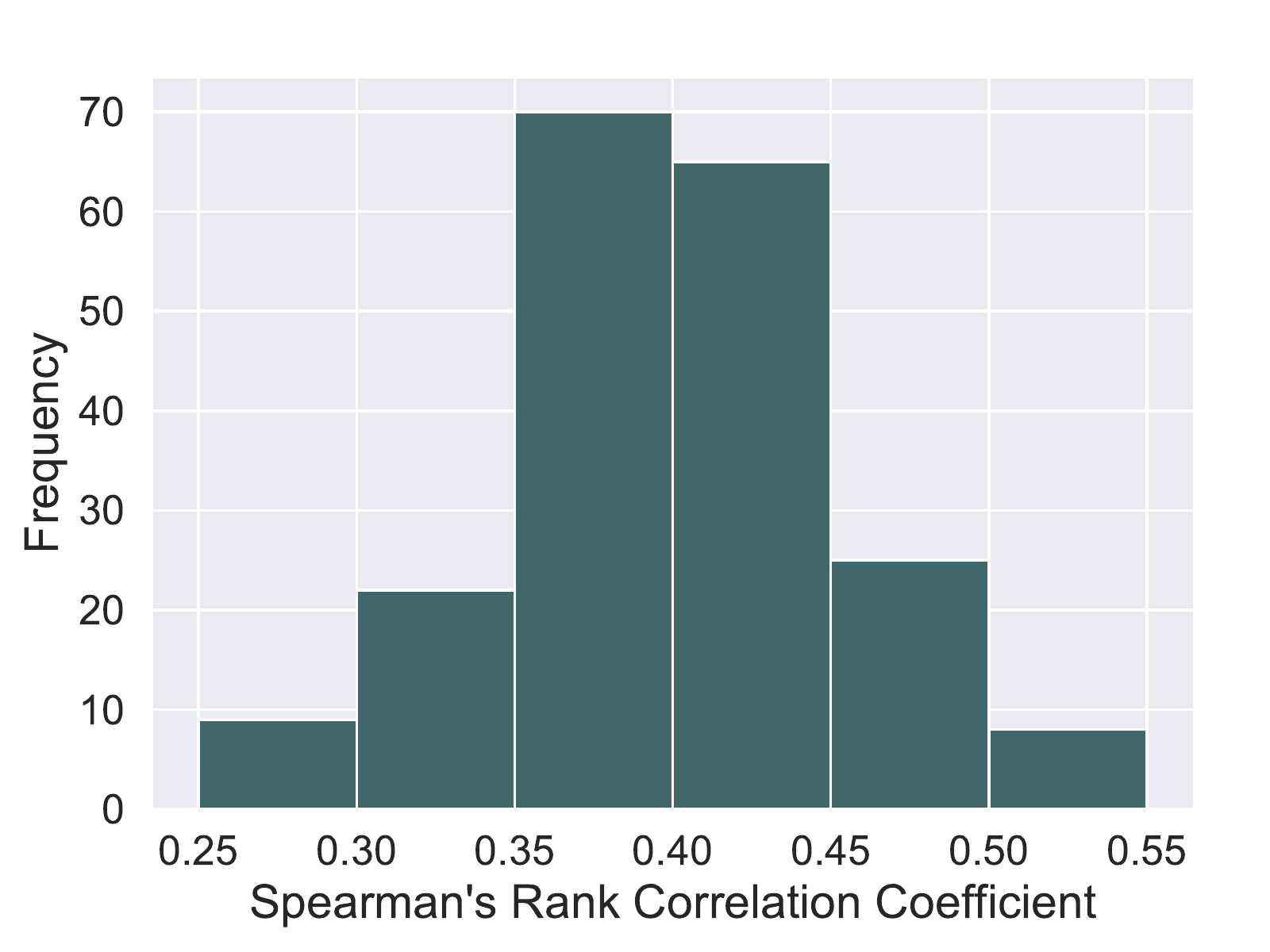}
}
\caption{\draft{The Spearman's Rank Correlation Coefficient of the reordered lists based on neuron sensitivity and neuron importance using PGD targeted attacks from different target labels. (a) records 10 classes similarities of VGG-16 Vanilla model on CIFAR-10 in a bar chart; (b) denotes the histogram of 200 classes similarities of ResNet-18 Vanilla model on ImageNet.}}% The scores are slightly lower than the Levenshtein Similarities, which proves that the more sensitive neurons are towards adversarial perturbations, the more likely they simultaneously make the most non-trivial contributions to model predictions.}}
\label{fig:Spearman correlation}
\end{minipage}
\hspace{0.015\linewidth}
\begin{minipage}[t]{0.48\linewidth}
\centering
\subfigure[]{
\includegraphics[width=0.46\linewidth]{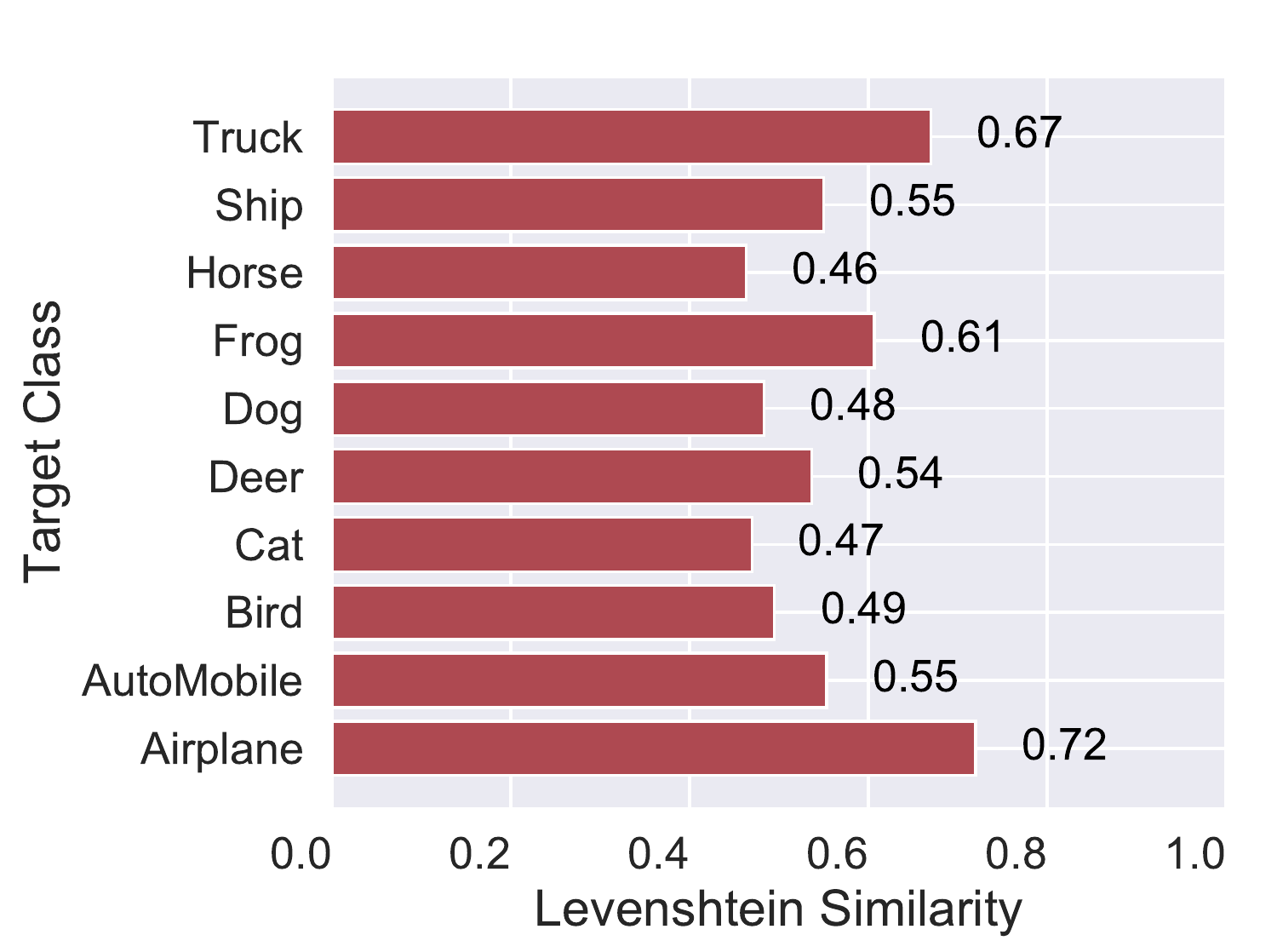}
}
\subfigure[]{
\includegraphics[width=0.46\linewidth]{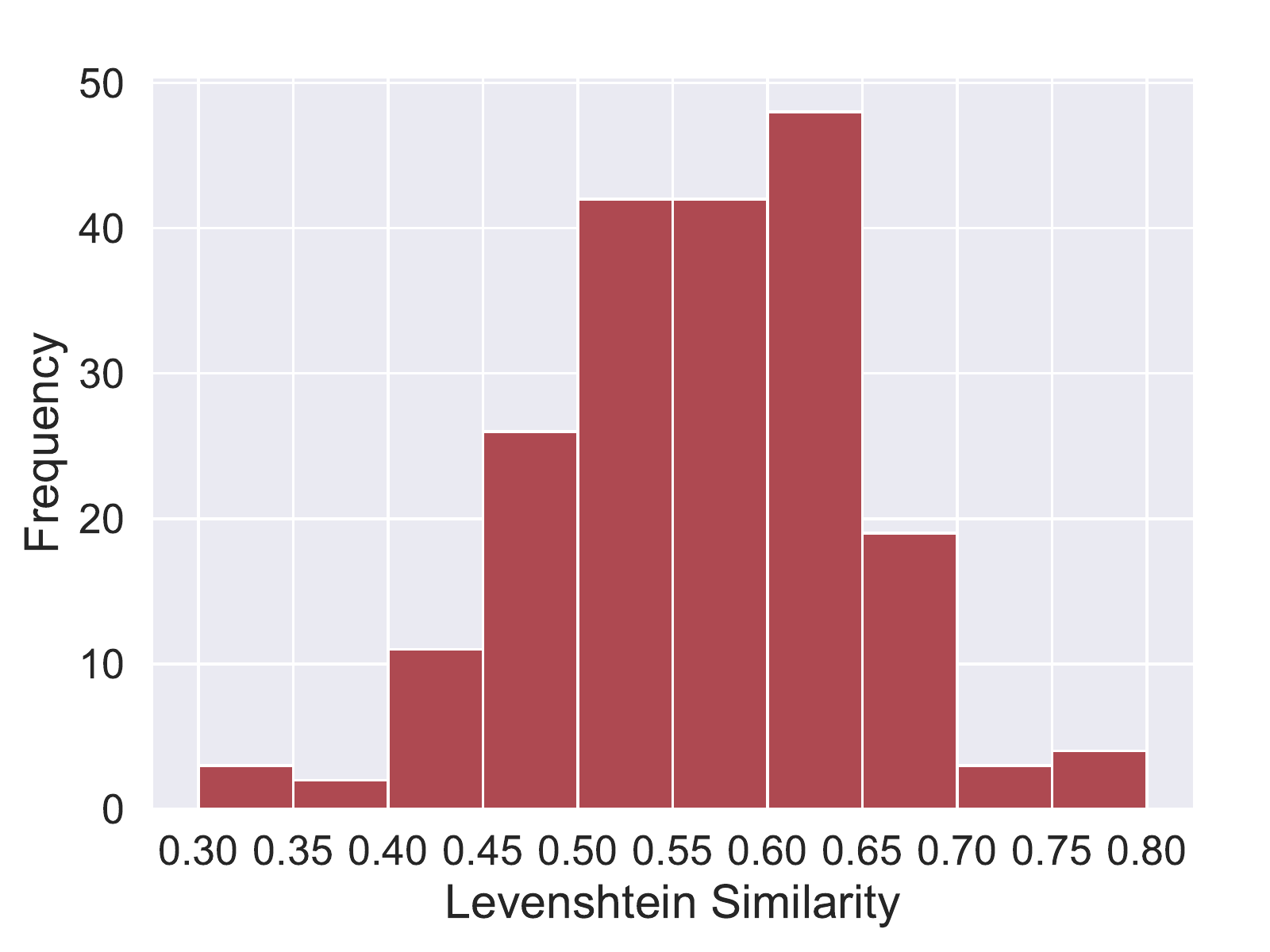}
}
\caption{\textcolor{black}{The Levenshtein Similarity of $\Gamma^{y}$ and $\Omega_{\textrm{L}-1}^{y}$ using PGD targeted attacks from different target labels. \draft{(a) records 10 classes similarities of VGG-16 Vanilla model on CIFAR-10 in a bar chart; (b) denotes the histogram of 200 classes similarities of ResNet-18 Vanilla model on ImageNet.}}} %We can see high similarities between sensitive neurons and important neurons in both cases, which indicate that sensitive neurons uncover the strongest weakness for deep models.}}}
\label{fig:Levenshtein Similarity}
\end{minipage}

\end{figure*}

\draft{The results of Spearman's Rank Correlation Coefficients between neuron importance and sensitivity on CIFAR-10 and ImageNet are shown in Figure \ref{fig:Spearman correlation}. For CIFAR-10, the similarity for each class ranges from 0.36 to 0.53 with the average value 0.428. For ImageNet, due to limited space, we provide the frequency of classes at different similarity score levels. More than 85\% classes obtain similarity score higher than 0.35; more than 50\% classes obtain similarity score higher than 0.40. Acording to the results, there exists a positive correlation between the neuron sensitivity and the neuron importance. In other words, the more sensitive the neurons behave, the more likely they make stronger contributions to the model prediction.}

\draft{The above mentioned Spearman's Rank Correlation Coefficient calculates the dependence between two whole sequences. However, the important neurons $\Gamma^{y}$ and sensitive neurons $\Omega_{\textrm{L}-1}^{y}$ are computed by choosing the most prominent portion via top-$k$. Thus, it's more appropriate to take the part of the whole sequence into consideration, which could better capture the similarity between sensitive and important neurons. Based on the above analysis, we further exploit the Levenshtein Distance to measure the similarity between the important neuron and sensitive neuron sequences. Levenshtein Distance measures the minimum number of single-element edit operations from one sequence to another, and it only allows the insertion, deletion and substitution operations. For two sequences $a$ and $b$, the calculation of Levenshtein Distance $\omega_{a,b}(|a|,|b|)$ can be described as below:}

\begin{equation}
\begin{small}
\omega_{a,b}(i,j)=
\left\{\begin{array}{ll} \hspace{-2mm} \max(i,j) & \hspace{-6mm} \text{if} \ \min(i,j) = 0,\\
    \hspace{-2mm} \min \left\{\begin{array}{l} \hspace{-2mm} \omega_{a,b}(i-1,j)+1\\
    \hspace{-2mm} \omega_{a,b}(i,j-1)+1\\
    \hspace{-2mm} \omega_{a,b}(i-1,j-1)+1_{(a_i\neq b_j)}\end{array}
    \right.
    & \hspace{-6mm} \text{otherwise,}\end{array}
\right.
\end{small}
\label{equa:Levenshtein Distance}
\end{equation}
\textcolor{black}{where $|a|$ denotes the lenth of sequence $a$ and $\omega_{a,b}(i,j)$ means the distance between the first $i$ elements of sequence $a$ and first $j$ elements of sequence $b$. Thus, we can calculate the Levenshtein Similarity as follow:}
\begin{equation}
S_{\omega}(a,b)=1-\frac{\omega_{a,b}(|a|,|b|)}{|a|+|b|}.
\label{equa:Levenshtein Similarity}
\end{equation}

Note that the range of $S_{\omega}(a,b)$ is $[0,1]$, where a higher value represents higher similarity. \draft{We respectively make statistics of per-class Levenshtein Similarities between $\Omega_{\textrm{L}-1}^{y}$ and $\Gamma^{y}$ on CIFAR-10 and ImageNet datasets. For CIFAR-10, the similarity for each class ranges from 0.46 to 0.72 with the average value 0.554. For ImageNet, we likewise provide the frequency of classes at different similarity score levels. More than 80\% classes obtain similarity score higher than 0.50; more than 36\% classes obtain similarity score higher than 0.60. We can still observe certain positive correlation in  this metric. However, the scores of Spearman's Rank Correlation Coefficients are slightly lower than the Levenshtein Similarities. The reasons behind might be the different calculation process: the Levenshtein Similarities focus on measuring similarity between $\Omega_{\textrm{L}-1}^{y}$ and $\Gamma^{y}$, while the Spearman's Rank Correlation Coefficients quantify the correlation of two complete ordered lists.}

\begin{figure*}[!tbp]
\centering
\begin{minipage}[c]{1\linewidth}
\subfigure[Neuron 6]{
\begin{minipage}[b]{0.32\linewidth}

\begin{minipage}[c]{1\linewidth}
\includegraphics[width=0.18\linewidth]{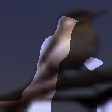}
\includegraphics[width=0.18\linewidth]{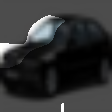}
\includegraphics[width=0.18\linewidth]{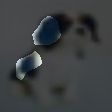}
\includegraphics[width=0.18\linewidth]{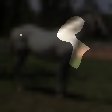}
\includegraphics[width=0.18\linewidth]{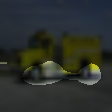}
\end{minipage}

\begin{minipage}[c]{1\linewidth}
\includegraphics[width=0.18\linewidth]{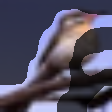}
\includegraphics[width=0.18\linewidth]{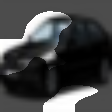}
\includegraphics[width=0.18\linewidth]{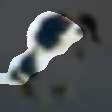}
\includegraphics[width=0.18\linewidth]{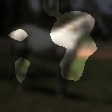}
\includegraphics[width=0.18\linewidth]{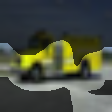}
\end{minipage}

\end{minipage}}\hspace{0.015\linewidth}
\subfigure[Neuron 53]{
\begin{minipage}[b]{0.32\linewidth}

\begin{minipage}[c]{1\linewidth}
\includegraphics[width=0.18\linewidth]{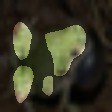}
\includegraphics[width=0.18\linewidth]{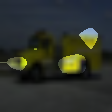}
\includegraphics[width=0.18\linewidth]{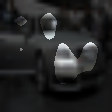}
\includegraphics[width=0.18\linewidth]{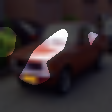}
\includegraphics[width=0.18\linewidth]{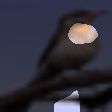}
\end{minipage}

\begin{minipage}[c]{1\linewidth}
\includegraphics[width=0.18\linewidth]{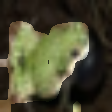}
\includegraphics[width=0.18\linewidth]{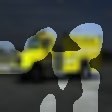}
\includegraphics[width=0.18\linewidth]{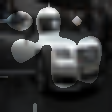}
\includegraphics[width=0.18\linewidth]{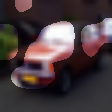}
\includegraphics[width=0.18\linewidth]{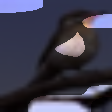}
\end{minipage}

\end{minipage}}\hspace{0.015\linewidth}
\subfigure[Neuron 74]{
\begin{minipage}[b]{0.32\linewidth}

\begin{minipage}[c]{1\linewidth}
\includegraphics[width=0.18\linewidth]{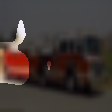}
\includegraphics[width=0.18\linewidth]{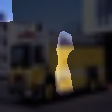}
\includegraphics[width=0.18\linewidth]{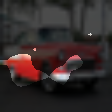}
\includegraphics[width=0.18\linewidth]{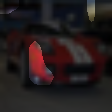}
\includegraphics[width=0.18\linewidth]{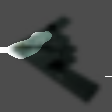}
\end{minipage}

\begin{minipage}[c]{1\linewidth}
\includegraphics[width=0.18\linewidth]{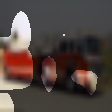}
\includegraphics[width=0.18\linewidth]{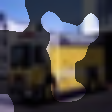}
\includegraphics[width=0.18\linewidth]{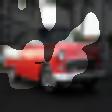}
\includegraphics[width=0.18\linewidth]{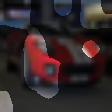}
\includegraphics[width=0.18\linewidth]{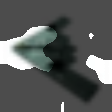}
\end{minipage}
\end{minipage}}
\end{minipage}

\caption{Image segmentation results of sensitive neurons on benign (top line) and adversarial examples (bottom line). (a), (b) and (c) represent sensitive neuron \emph{6}, \emph{53} and \emph{74} in the \emph{pool2} layer of VGG-16 Vanilla model on CIFAR-10, respectively.}
\label{RF_sen_cifar10}
\end{figure*}

\begin{figure*}[!tbp]
\centering
\begin{minipage}[c]{1\linewidth}
\subfigure[Neuron 5]{
\begin{minipage}[b]{0.32\linewidth}

\begin{minipage}[c]{1\linewidth}
\includegraphics[width=0.18\linewidth]{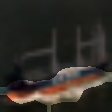}
\includegraphics[width=0.18\linewidth]{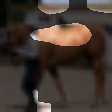}
\includegraphics[width=0.18\linewidth]{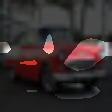}
\includegraphics[width=0.18\linewidth]{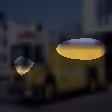}
\includegraphics[width=0.18\linewidth]{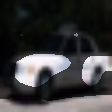}
\end{minipage}

\begin{minipage}[c]{1\linewidth}
\includegraphics[width=0.18\linewidth]{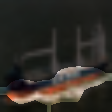}
\includegraphics[width=0.18\linewidth]{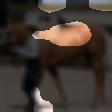}
\includegraphics[width=0.18\linewidth]{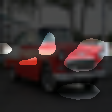}
\includegraphics[width=0.18\linewidth]{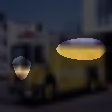}
\includegraphics[width=0.18\linewidth]{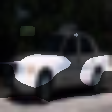}
\end{minipage}

\end{minipage}}\hspace{0.015\linewidth}
\subfigure[Neuron 73]{
\begin{minipage}[b]{0.32\linewidth}

\begin{minipage}[c]{1\linewidth}
\includegraphics[width=0.18\linewidth]{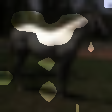}
\includegraphics[width=0.18\linewidth]{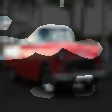}
\includegraphics[width=0.18\linewidth]{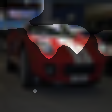}
\includegraphics[width=0.18\linewidth]{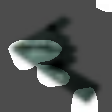}
\includegraphics[width=0.18\linewidth]{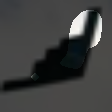}
\end{minipage}

\begin{minipage}[c]{1\linewidth}
\includegraphics[width=0.18\linewidth]{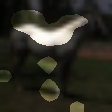}
\includegraphics[width=0.18\linewidth]{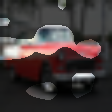}
\includegraphics[width=0.18\linewidth]{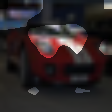}
\includegraphics[width=0.18\linewidth]{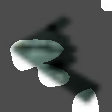}
\includegraphics[width=0.18\linewidth]{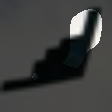}
\end{minipage}

\end{minipage}}\hspace{0.015\linewidth}
\subfigure[Neuron 45]{
\begin{minipage}[b]{0.32\linewidth}

\begin{minipage}[c]{1\linewidth}
\includegraphics[width=0.18\linewidth]{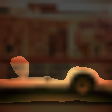}
\includegraphics[width=0.18\linewidth]{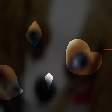}
\includegraphics[width=0.18\linewidth]{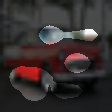}
\includegraphics[width=0.18\linewidth]{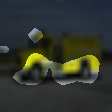}
\includegraphics[width=0.18\linewidth]{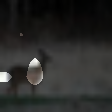}
\end{minipage}

\begin{minipage}[c]{1\linewidth}
\includegraphics[width=0.18\linewidth]{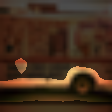}
\includegraphics[width=0.18\linewidth]{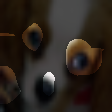}
\includegraphics[width=0.18\linewidth]{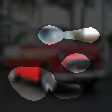}
\includegraphics[width=0.18\linewidth]{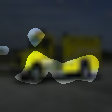}
\includegraphics[width=0.18\linewidth]{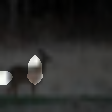}
\end{minipage}

\end{minipage}}
\end{minipage}

\caption{Image segmentation results of vanilla neurons on benign (top line) and adversarial examples (bottom line). (a), (b) and (c) represent vanilla neuron \emph{5}, \emph{73} and \emph{45} in the \emph{pool2} layer of VGG-16 Vanilla model on CIFAR-10, respectively.}
\label{RF_vanilla_cifar10}
\end{figure*}

\begin{figure*}[!tbp]
\centering
\begin{minipage}[c]{1\linewidth}
\subfigure[Neuron 430]{
\begin{minipage}[b]{0.32\linewidth}

\begin{minipage}[c]{1\linewidth}
\includegraphics[width=0.18\linewidth]{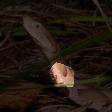}
\includegraphics[width=0.18\linewidth]{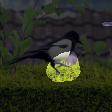}
\includegraphics[width=0.18\linewidth]{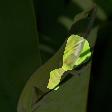}
\includegraphics[width=0.18\linewidth]{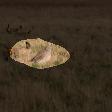}
\includegraphics[width=0.18\linewidth]{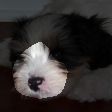}
\end{minipage}

\begin{minipage}[c]{1\linewidth}
\includegraphics[width=0.18\linewidth]{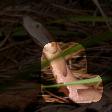}
\includegraphics[width=0.18\linewidth]{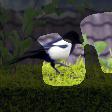}
\includegraphics[width=0.18\linewidth]{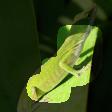}
\includegraphics[width=0.18\linewidth]{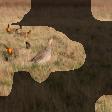}
\includegraphics[width=0.18\linewidth]{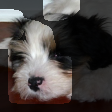}
\end{minipage}

\end{minipage}}\hspace{0.015\linewidth}
\subfigure[Neuron 321]{
\begin{minipage}[b]{0.32\linewidth}

\begin{minipage}[c]{1\linewidth}
\includegraphics[width=0.18\linewidth]{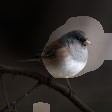}
\includegraphics[width=0.18\linewidth]{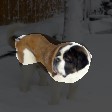}
\includegraphics[width=0.18\linewidth]{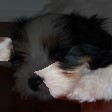}
\includegraphics[width=0.18\linewidth]{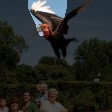}
\includegraphics[width=0.18\linewidth]{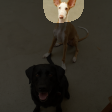}
\end{minipage}

\begin{minipage}[c]{1\linewidth}
\includegraphics[width=0.18\linewidth]{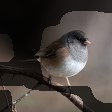}
\includegraphics[width=0.18\linewidth]{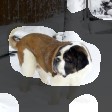}
\includegraphics[width=0.18\linewidth]{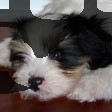}
\includegraphics[width=0.18\linewidth]{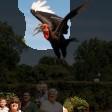}
\includegraphics[width=0.18\linewidth]{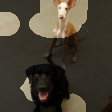}
\end{minipage}

\end{minipage}}\hspace{0.015\linewidth}
\subfigure[Neuron 114]{
\begin{minipage}[b]{0.32\linewidth}

\begin{minipage}[c]{1\linewidth}
\includegraphics[width=0.18\linewidth]{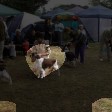}
\includegraphics[width=0.18\linewidth]{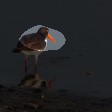}
\includegraphics[width=0.18\linewidth]{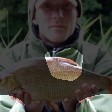}
\includegraphics[width=0.18\linewidth]{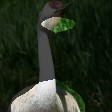}
\includegraphics[width=0.18\linewidth]{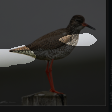}
\end{minipage}

\begin{minipage}[c]{1\linewidth}
\includegraphics[width=0.18\linewidth]{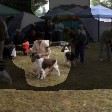}
\includegraphics[width=0.18\linewidth]{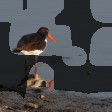}
\includegraphics[width=0.18\linewidth]{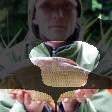}
\includegraphics[width=0.18\linewidth]{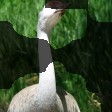}
\includegraphics[width=0.18\linewidth]{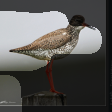}
\end{minipage}

\end{minipage}}
\end{minipage}

\caption{Image segmentation results of sensitive neurons on benign (top line) and adversarial examples (bottom line). (a), (b) and (c) represent sensitive neuron \emph{430}, \emph{321} and \emph{114} in the \emph{layer4\_output} layer of ResNet-18 Vanilla model on ImageNet, respectively. After adversarial attack, sensitive neurons tend to pay more attention to the noisy backgrounds and other meaningless regions, compared to their subtle detection regions on benign examples.}
\label{RF_sen_imagenet}
\end{figure*}

\begin{figure*}[!tbp]
\centering
\begin{minipage}[c]{1\linewidth}
\subfigure[Neuron 201]{
\begin{minipage}[b]{0.32\linewidth}

\begin{minipage}[c]{1\linewidth}
\includegraphics[width=0.18\linewidth]{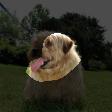}
\includegraphics[width=0.18\linewidth]{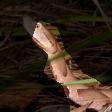}
\includegraphics[width=0.18\linewidth]{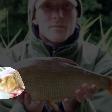}
\includegraphics[width=0.18\linewidth]{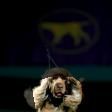}
\includegraphics[width=0.18\linewidth]{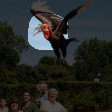}
\end{minipage}

\begin{minipage}[c]{1\linewidth}
\includegraphics[width=0.18\linewidth]{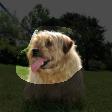}
\includegraphics[width=0.18\linewidth]{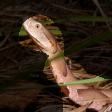}
\includegraphics[width=0.18\linewidth]{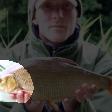}
\includegraphics[width=0.18\linewidth]{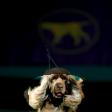}
\includegraphics[width=0.18\linewidth]{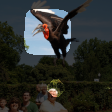}
\end{minipage}

\end{minipage}}\hspace{0.015\linewidth}
\subfigure[Neuron 32]{
\begin{minipage}[b]{0.32\linewidth}

\begin{minipage}[c]{1\linewidth}
\includegraphics[width=0.18\linewidth]{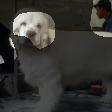}
\includegraphics[width=0.18\linewidth]{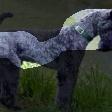}
\includegraphics[width=0.18\linewidth]{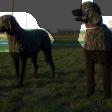}
\includegraphics[width=0.18\linewidth]{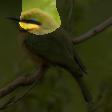}
\includegraphics[width=0.18\linewidth]{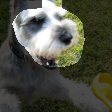}
\end{minipage}

\begin{minipage}[c]{1\linewidth}
\includegraphics[width=0.18\linewidth]{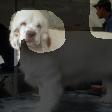}
\includegraphics[width=0.18\linewidth]{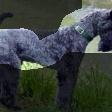}
\includegraphics[width=0.18\linewidth]{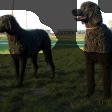}
\includegraphics[width=0.18\linewidth]{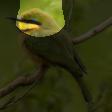}
\includegraphics[width=0.18\linewidth]{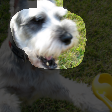}
\end{minipage}

\end{minipage}}\hspace{0.015\linewidth}
\subfigure[Neuron 282]{
\begin{minipage}[b]{0.32\linewidth}

\begin{minipage}[c]{1\linewidth}
\includegraphics[width=0.18\linewidth]{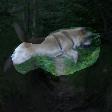}
\includegraphics[width=0.18\linewidth]{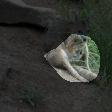}
\includegraphics[width=0.18\linewidth]{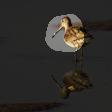}
\includegraphics[width=0.18\linewidth]{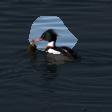}
\includegraphics[width=0.18\linewidth]{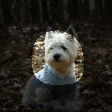}
\end{minipage}

\begin{minipage}[c]{1\linewidth}
\includegraphics[width=0.18\linewidth]{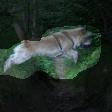}
\includegraphics[width=0.18\linewidth]{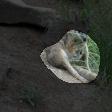}
\includegraphics[width=0.18\linewidth]{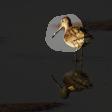}
\includegraphics[width=0.18\linewidth]{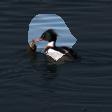}
\includegraphics[width=0.18\linewidth]{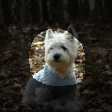}
\end{minipage}

\end{minipage}}
\end{minipage}

\caption{Image segmentation results of vanilla neurons on benign (top line) and adversarial examples (bottom line). (a), (b) and (c) represent vanilla neuron \emph{201}, \emph{32} and \emph{282} in the \emph{layer4\_output} layer of ResNet-18 Vanilla model on ImageNet, respectively. The Region of Interests for vanilla neurons show almost no differences between benign and adversarial examples.}
\label{RF_unsen_imagenet}
\end{figure*}

%\subsubsection{Sensitive Neurons are more susceptible to adversarial examples}
We further explore the behaviors of sensitive neurons in adversarial settings by showing what they detect during inference through visualization studies. Following the work in \cite{DBLP:journals/corr/ZhouKLOT14}, we investigate the region of interest for different neurons (e.g., sensitive neurons and vanilla ones) on CIFAR-10 with VGG-16 and on ImageNet with ResNet-18 using the image segmentation based method. Due to the limited space, we only present top-3 most typical neurons of each sequence, i.e., the sensitive neurons with the highest sensitivity and vanilla neurons with the lowest sensitivity. For each neuron, we first find top-5 images with the highest activation in the benign sample set and then visualize image segmentation results of them and their corresponding adversarial examples generated by $\ell_\infty$ PGD untargeted attack. According to visualization results in Figure \ref{RF_sen_cifar10} and \ref{RF_sen_imagenet}, after the adversarial attack, sensitive neurons tend to pay more attention to noisy backgrounds and other meaningless regions, compared to their subtle detection regions on benign examples. Instead, as illustrated in Figure \ref{RF_vanilla_cifar10} and \ref{RF_unsen_imagenet}, no evident variance can be observed between benign and adversarial examples in image segmentation results of vanilla neurons. With the above detection, we double confirm the conclusion that sensitive neurons are more sensitive to adversarial noises and play critical roles to model's final misclassification in the adversarial setting.

\begin{figure}[!h]
\centering
\subfigure[]{
\includegraphics[width=0.46\linewidth]{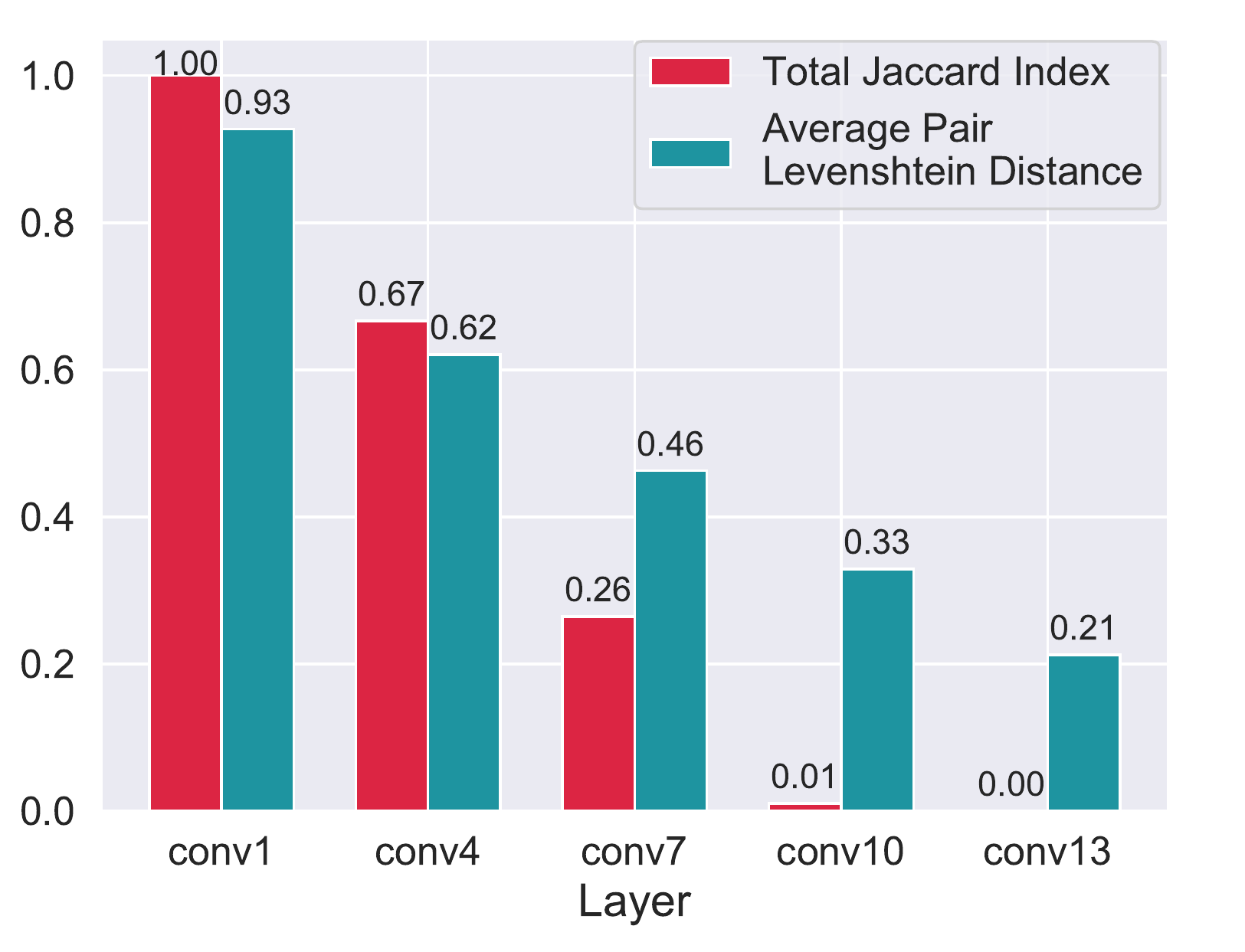}
}
\subfigure[]{
\includegraphics[width=0.46\linewidth]{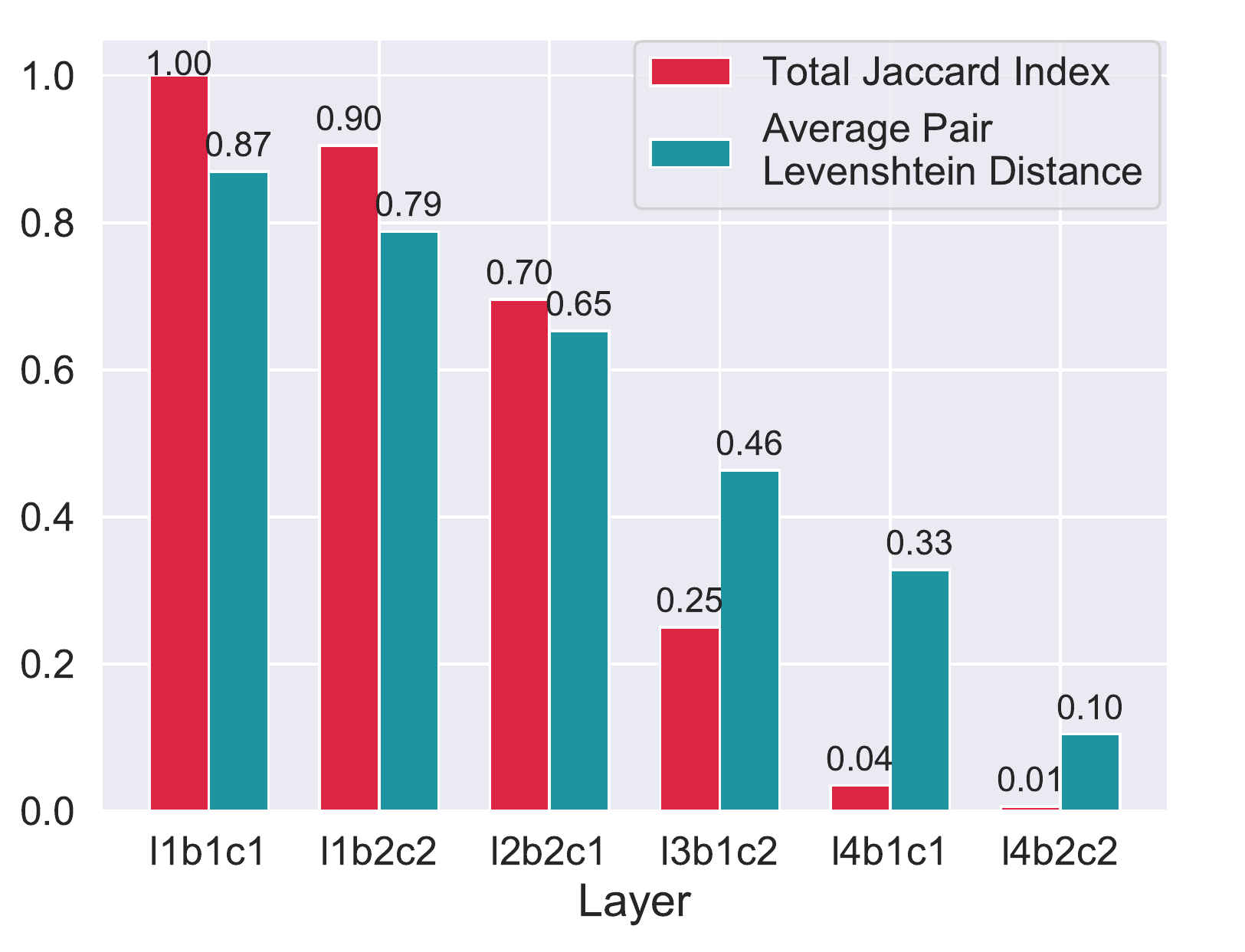}
}
\caption{The total Jaccard index and average pair Levenshtein Similarity of sensitive neuron sequences $\Omega_{l}^{1},\ldots,\Omega_{l}^{10}$ of PGD attack for 10 different target labels. Subfigure (a) and (b) separately demonstrate the situation of the VGG-16 Vanilla model on CIFAR-10 and ResNet-18 Vanilla model on ImageNet. Classes chosen on ImageNet follow the settings in Section \ref{Section:multiple jaccard}. The results indicate that adversarial examples with different target labels tend to share the same flaws of bottom layers, while utilizing different fragile neurons in the top layers.}
\label{fig:TenClassIoU}
\end{figure}

\subsection{Adversarial Attacks Exploit Sensitive Neurons Differently at Different Layers}\label{Section:multiple jaccard}

We further explore characteristics of sensitive neurons at layers of different depths. \draft{We basically follow the same setting in Section \ref{Section:Jaccard} except that we random pick 10 classes from ImageNet from a practical point of view. These classes include \emph{Academic Gown}, \emph{Black Stork}, \emph{Bucket}, \emph{Dumbbell}, \emph{Goldfish}, \emph{Ice Lolly}, \emph{Miniskirt}, \emph{Refrigerator}, \emph{Stick Insect} and \emph{Teapot}.} The sensitive neurons sequences $\Omega_{l}^{1},\ldots,\Omega_{l}^{\mathit{Y}}$ are obtained, where $l$ denotes the index of layer we use. We adopt the symbol $y$ to stand for the target label index, i.e., $y=1,\ldots,\mathit{Y}$, and $\mathit{Y}$ is 10 in this case. To measure the similarity of these sequences in one specific layer $l$, \draft{we adopt two metrics as follows:}

\draft{\emph{Average pair Levenshtein Similarity.} It is used to represent the average case of the similarity between two different sequences in layer $l$:}
\begin{equation}
%J_{avg}(\Omega_{l}^{1},\ldots,\Omega_{l}^{\mathit{Y}})=\frac{1}{\mathit{M}}\sum_{1\leq y,y'\leq \mathit{Y}}^{y\neq y'}\frac{|\Omega_{l}^{y} \cap \Omega_{l}^{y'}|}{|\Omega_{l}^{y} \cup \Omega_{l}^{y'}|},
S_{\omega}^{avg}(\Omega_{l}^{1},\ldots,\Omega_{l}^{\mathit{Y}})=\frac{1}{\mathit{M}}\sum_{1\leq y,y'\leq \mathit{Y}}^{y\neq y'} S_{\omega}(\Omega_{l}^{y},\Omega_{l}^{y'}),
\label{equa:average pair Levenshtein Similarity}
\end{equation}
where $\mathit{M}$ denotes the total pair number.

\draft{\emph{Total Jaccard index.} Levenshtein Similarity is not applicable to calculate the similarities of several sensitive neuron sequences, since it does not handle multiple inputs. By regarding the ordered sequences as sets, we can use Total Jaccard index to simultaneously measure the common similarity of all sensitive neuron sequences $\Omega_{l}^{y}$ in the specific layer $l$. Formally, Total Jaccard index can be characterized by the common overlap of these sets:}
\begin{equation}
S_{\delta}^{total}(\Omega_{l}^{1},\ldots,\Omega_{l}^{\mathit{Y}})=\frac{|\bigcap_{y=1}^{\mathit{Y}}\Omega_{l}^{y}|}{|\bigcup_{y=1}^{\mathit{Y}}\Omega_{l}^{y}|}.
\label{equa:total Jaccard index}
\end{equation}

Figure \ref{fig:TenClassIoU} (a) and (b) separately show the results on CIFAR-10 and ImageNet, from which we draw an important observation that the sensitive neurons in different target set vary a lot in the top layers, but have high similarities in the bottom layers, though they are attacked by adversarial examples with different target labels. The reason may lie in the hierarchical information processing structure of DNNs, since bottom layers focus on common low-level semantic features, e.g., edges and textures, while top layers care more about high-level semantic features to specific classes \cite{DBLP:conf/eccv/ZeilerF14}. This interesting finding indicates that different targeted adversarial examples tend to share the same flaws of bottom layers, while utilizing different fragile neurons in the top layers.

\begin{figure*}[!ht]
\centering
\subfigure[$\ell_\infty$ PGD]{
\includegraphics[width=0.225\linewidth]{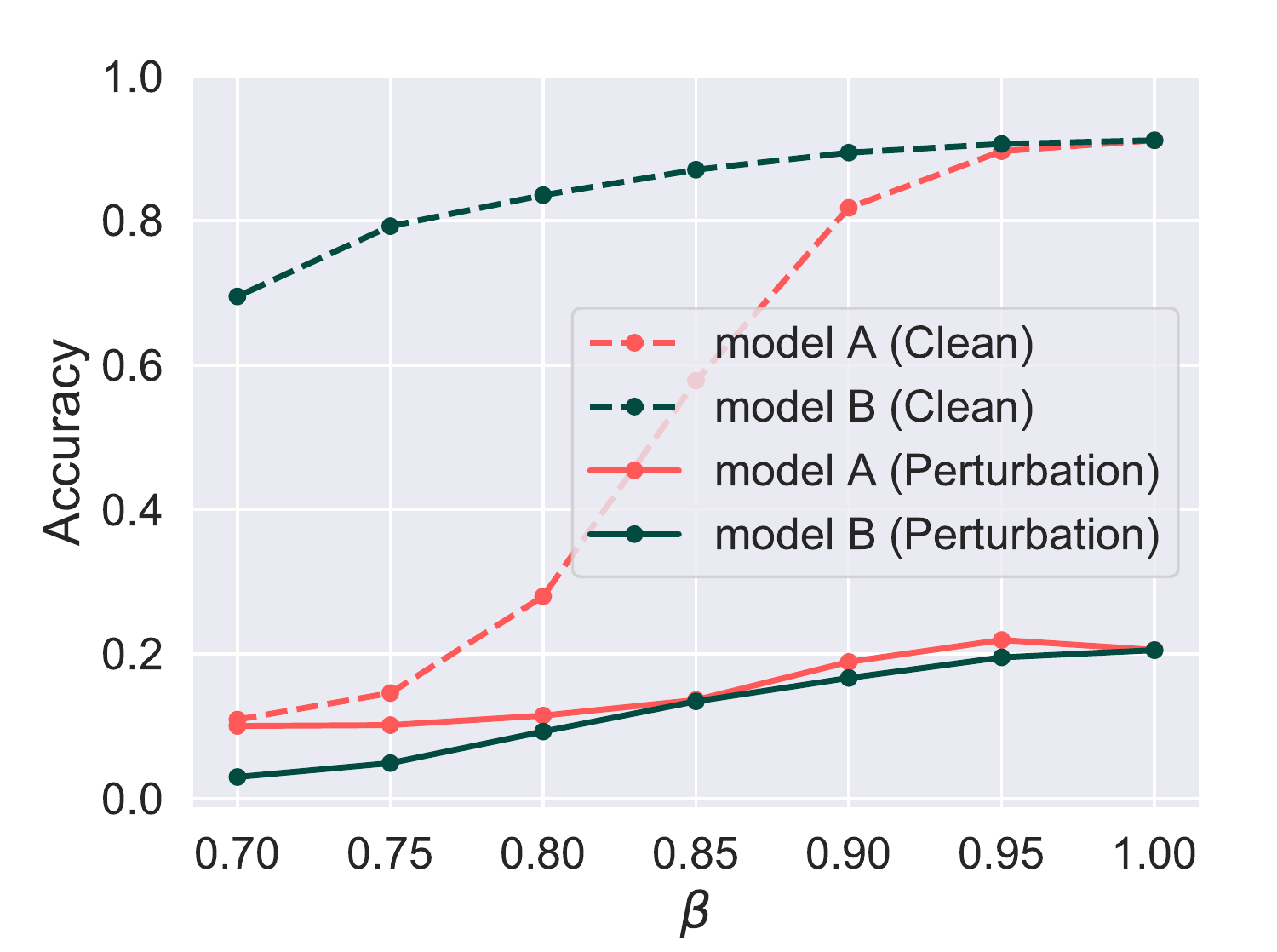}
}
\subfigure[$\ell_2$ PGD]{
\includegraphics[width=0.225\linewidth]{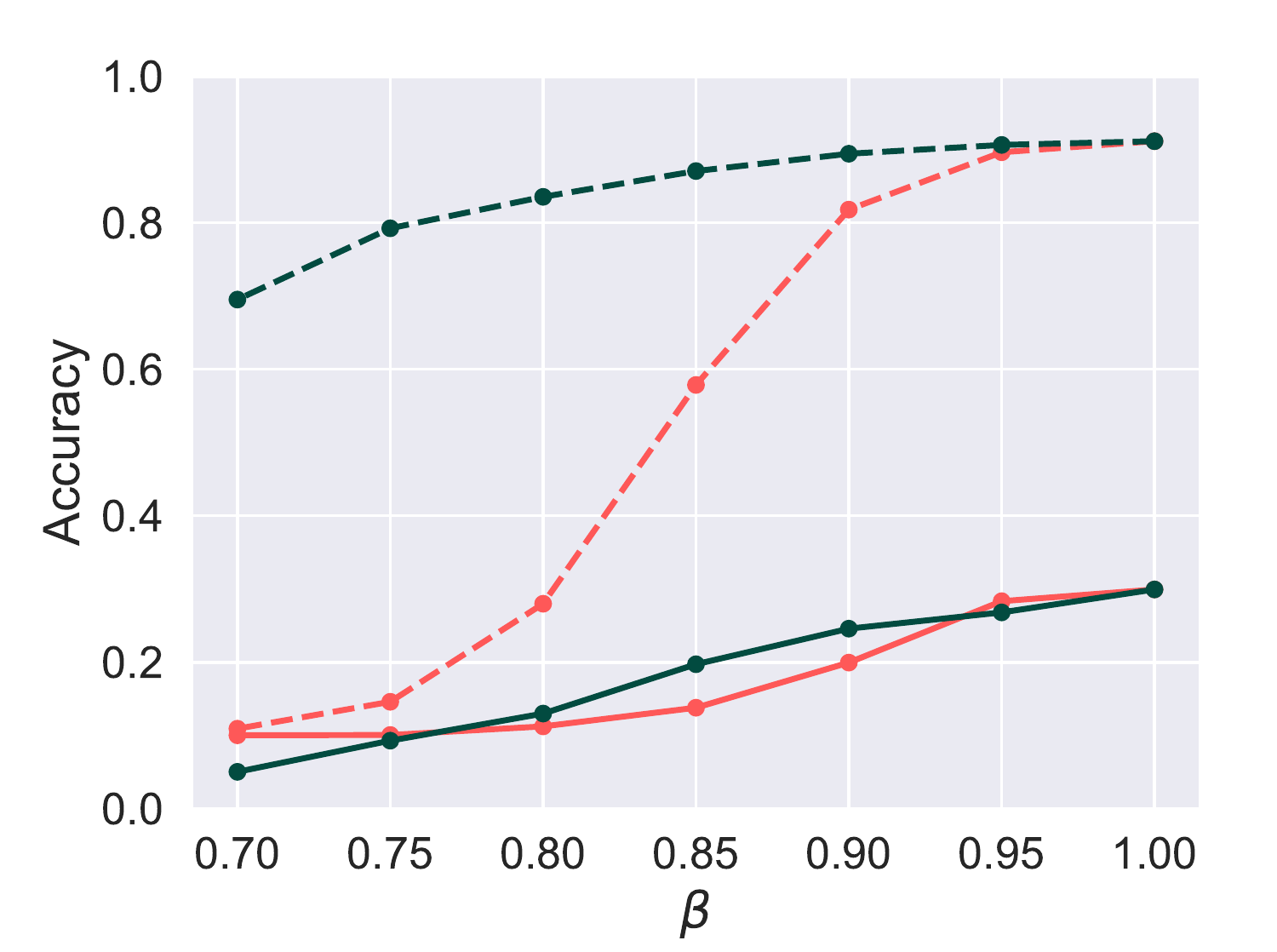}
}
\subfigure[FGSM]{
\includegraphics[width=0.225\linewidth]{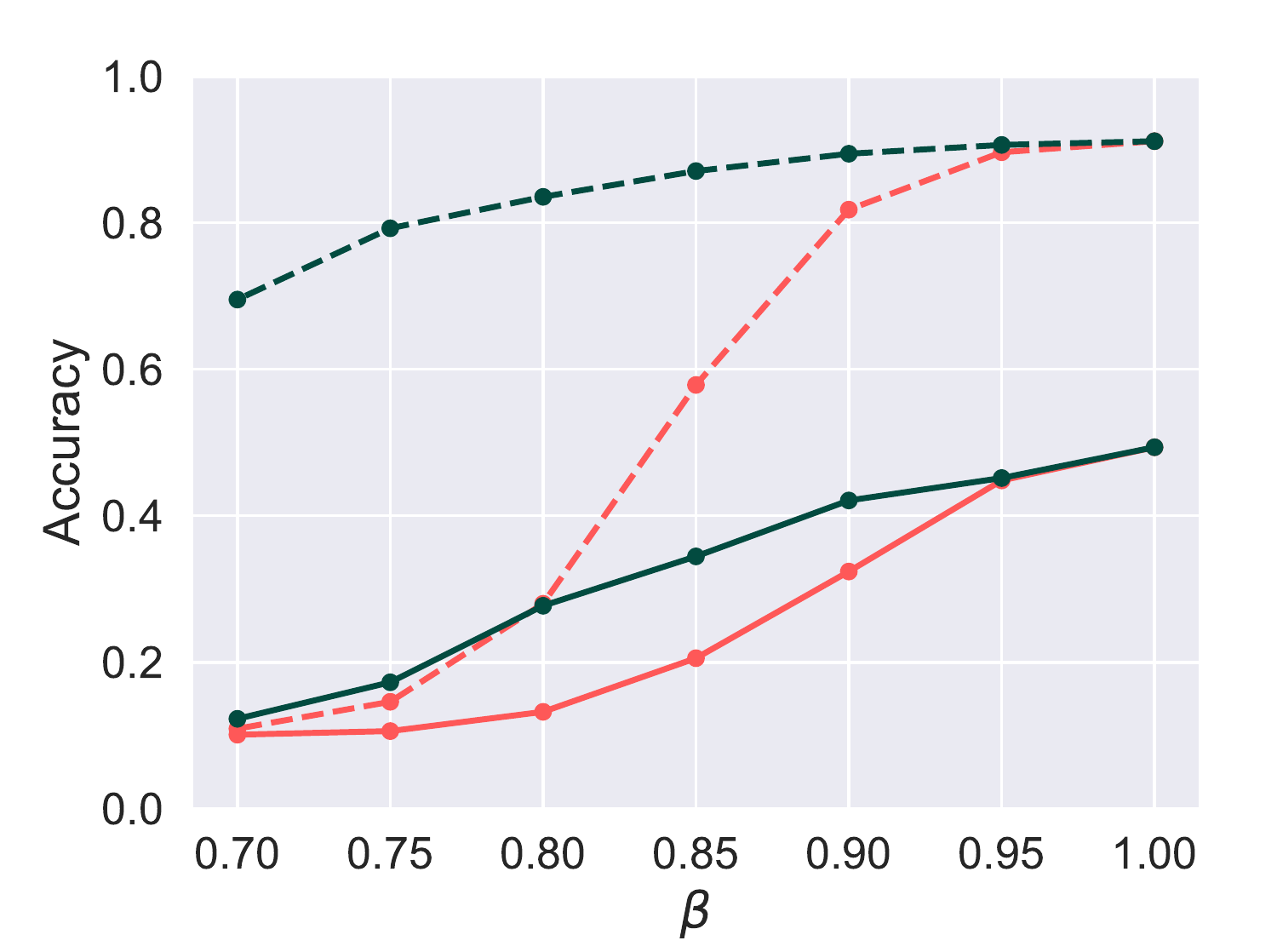}
}
\subfigure[Gaussian Noise]{
\includegraphics[width=0.225\linewidth]{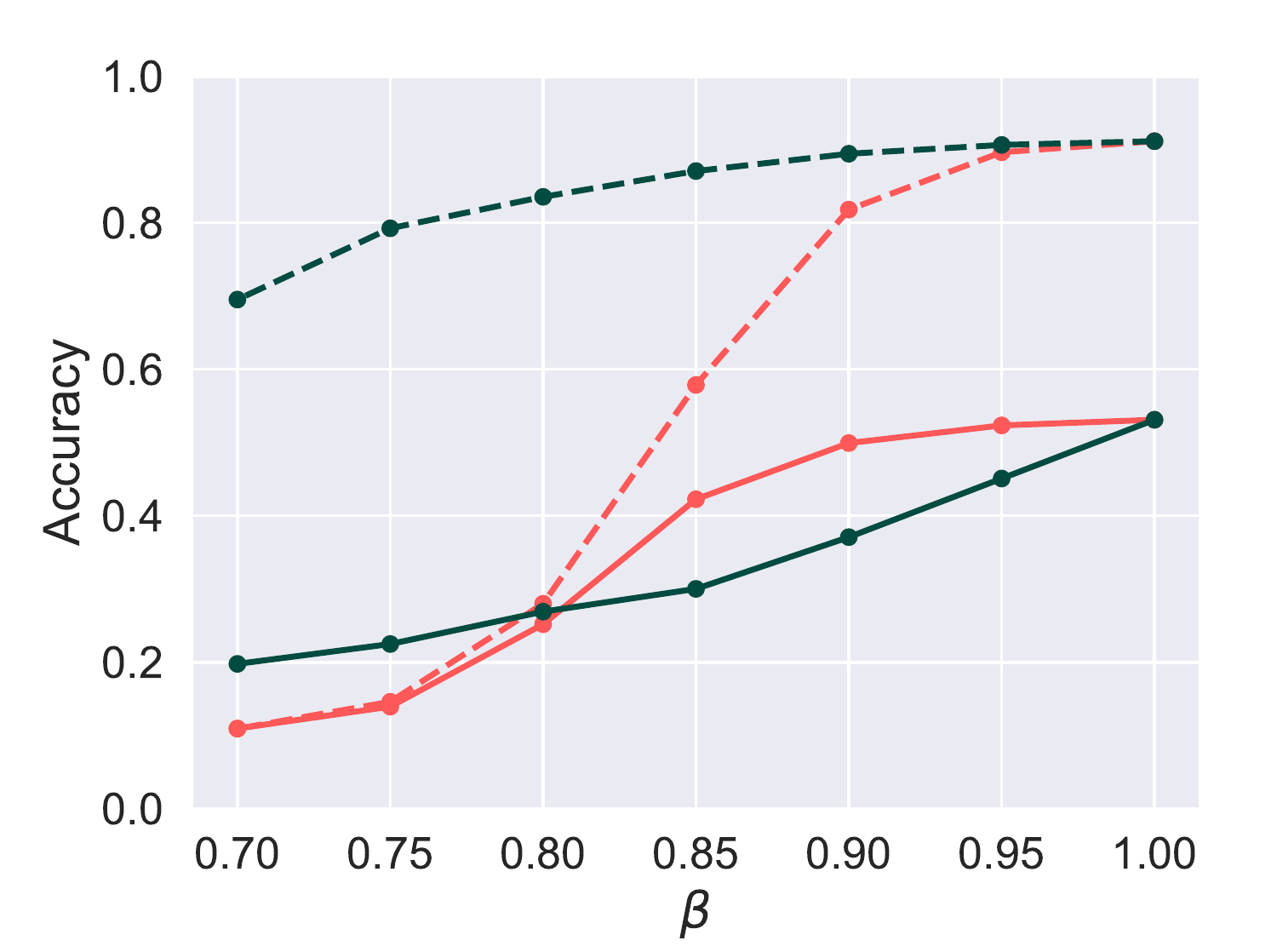}
}
\caption{Neurons behavior study by suppression on CIFAR-10 with VGG-16 Vanilla model. Specifically, we multiply the outputs of top-10\% sensitive neurons with a coefficient $\beta$ ($\beta \leq 1$). Model A and B respectively denote suppressing sensitive neurons and insensitive ones. We investigate how generalization and adversarial robustness are effected as each type of neurons are gradually suppressed. In Subfigure (a) to (d), we separately implement \emph{$\ell_\infty$ PGD}, \emph{$\ell_2$ PGD}, \emph{FGSM} and \emph{Gaussian Noise} for both models. Faster generalization degradation of sensitive neurons proves that they contain stronger semantic information. This indicates that sensitive neurons are responsible for both clean accuracy and robustness. There indeed exists a trade-off between robustness and accuracy.}
\label{fig:mix beta_suppression}
\end{figure*}

\subsection{Sensitive Neurons Convey Strong Semantic Information} \label{section:strong information}

As we have moved so far, we prefer to move further to give more insights about the roles of sensitive neurons. Prior studies have shown the ability to use suppression and ablation skills to study individual unit functions within a model \cite{DBLP:journals/corr/ZhouKLOT14,DBLP:conf/nips/ZhouLXTO14}. Inspired by them, we try to suppress the outputs of top-10\% sensitive neurons by multiplying them with a coefficient $\beta$ after activation (Vanilla model is obtained when $\beta$=1.0) with VGG-16 on CIFAR-10. For comparison, we randomly select the same amount of neurons from vanilla neurons set and do the same suppression operation as the control group. Additionally, we conduct experiments on the control group 3 times and compute the average of them as the final result to eliminate accidental errors. For the convenience of the narrative, we respectively name the experimental group and control group after \emph{model A} and \emph{model B}.

As demonstrated in Figure \ref{fig:mix beta_suppression}, focusing on the behaviors of clean examples, the performance of model A degenerates rapidly as $\beta$ decreases and soon collapses when $\beta$ reaches about $0.7$. In contrast, little impact on model B is observed and the model even has the ability of classification when $\beta=0.7$. Consequently, sensitive neurons extract strong semantic features for deep models, and model generalization will be significantly influenced if these neurons are suppressed. Meanwhile, there exists an interesting phenomenon in adversarial settings. When $\beta=0.95$, model A and B achieve similar performances on clean examples. At the same time, model A behaves better on adversarial robustness. This indicates that sensitive neurons truly respond more to adversarial robustness. However, the situation reverses as $\beta$ continues to drop. As the generalization difference of two models becomes more and more obvious, model A gradually loses the ability of classification and thus behaves terribly in the adversarial setting. Accordingly, there indeed exists a trade-off between robustness and accuracy \cite{DBLP:conf/iclr/TsiprasSETM19} and sensitive neurons are responsible for both clean accuracy and robustness.

\begin{table*}[!htb]
\centering
\caption{Model classification accuracy (\%) in clean and perturbed settings on CIFAR-10 with VGG-16.}
\begin{center}
\begin{small}
\begin{sc}
\setlength{\tabcolsep}{0.5mm}{
\begin{tabular}{c|c|cccc|ccccc}
\toprule
\multirow{2}*{Model} & \multirow{2}*{Clean} & \multicolumn{4}{c|}{PGD Attack} & \multicolumn{5}{c}{Gaussian Noise}\\
\cline{3-11}
~ & ~ & $\epsilon=2$ & $\epsilon=4$ & $\epsilon=6$ & $\epsilon=8$ & $s=1$ & $s=2$ & $s=3$ & $s=4$ & $s=5$\\ \hline
Vanilla & 91.1 & 20.6 & 2.5  & 0.6 & 0.1 & 82.7 & 68.5 & 53.1 & 45.8 & 39.9\\
PAT & 85.1  & 72.2 & 57.5 & 44.9 & 37.2 & 84.9 & 84.0 & 82.3 & 80.9 & 79.5\\
%Vanilla & 91.12 & 20.58 & 2.45  & 0.57 & 0.12 & 82.67 & 68.48 & 53.1 & 45.82 & 39.85\\
%PAT & 85.09  & 72.19 & 57.50 & 44.91 & 37.21 & 84.87 & 84.03 & 82.25 & 80.92 & 79.52\\
\bottomrule
\end{tabular}}
\end{sc}
\end{small}
\end{center}
\label{table:pgd_attack_acc}
\end{table*}

\begin{figure*}[!htb]
\centering
\subfigure[conv1]{
\includegraphics[width=0.225\linewidth]{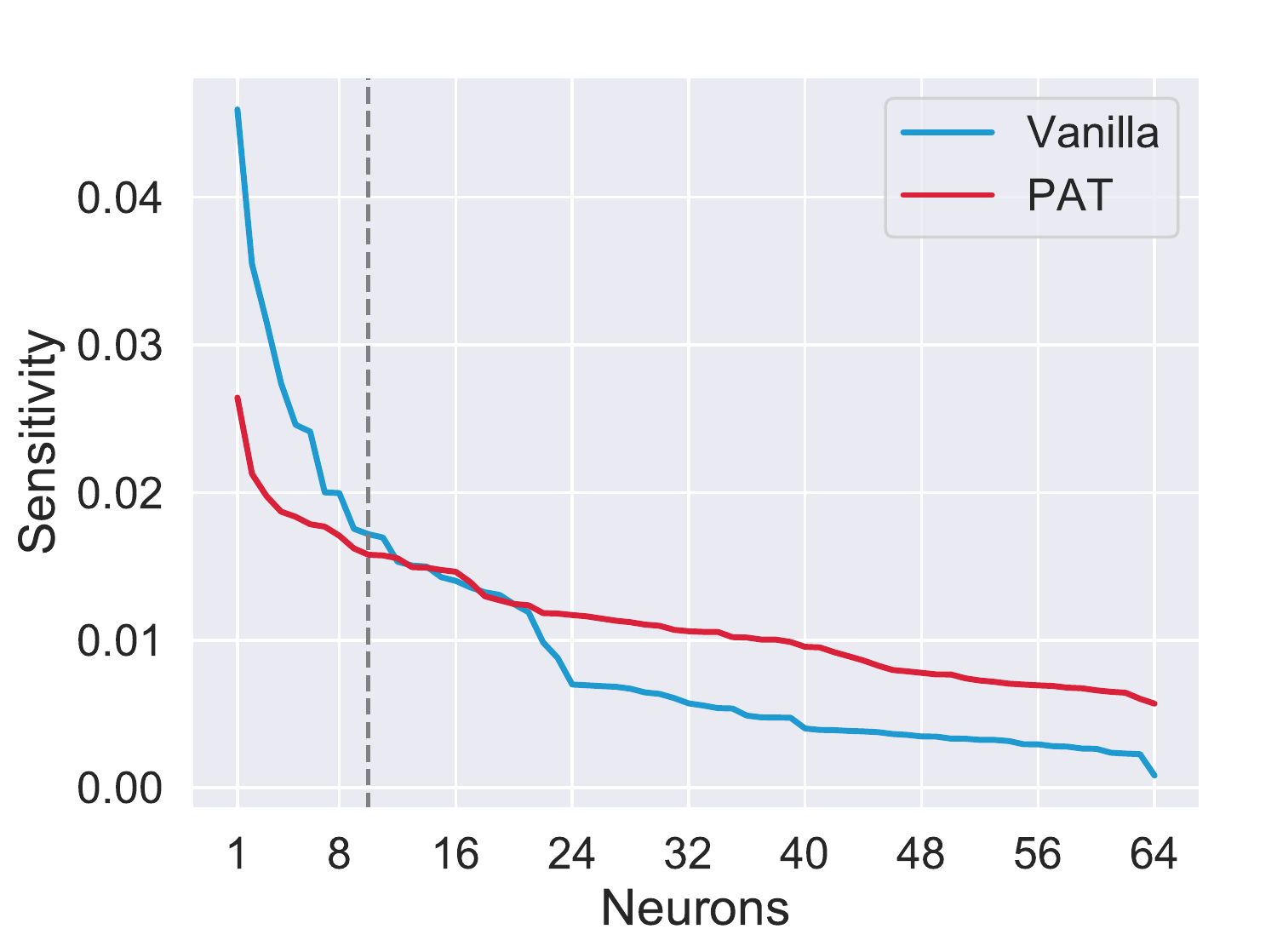}
}
\subfigure[conv4]{
\includegraphics[width=0.225\linewidth]{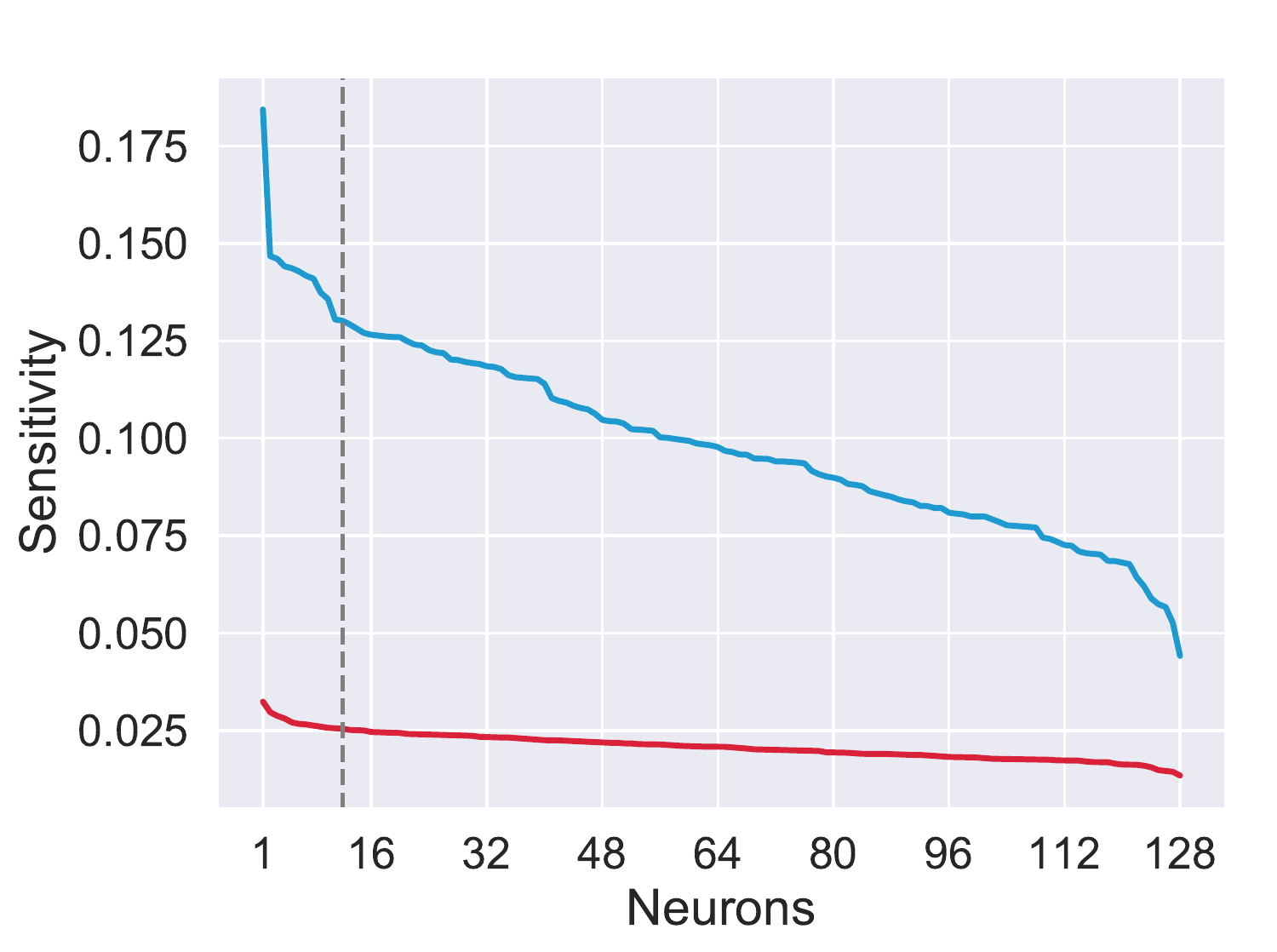}
}
\hspace{0.015\linewidth}
\subfigure[conv7]{
\includegraphics[width=0.225\linewidth]{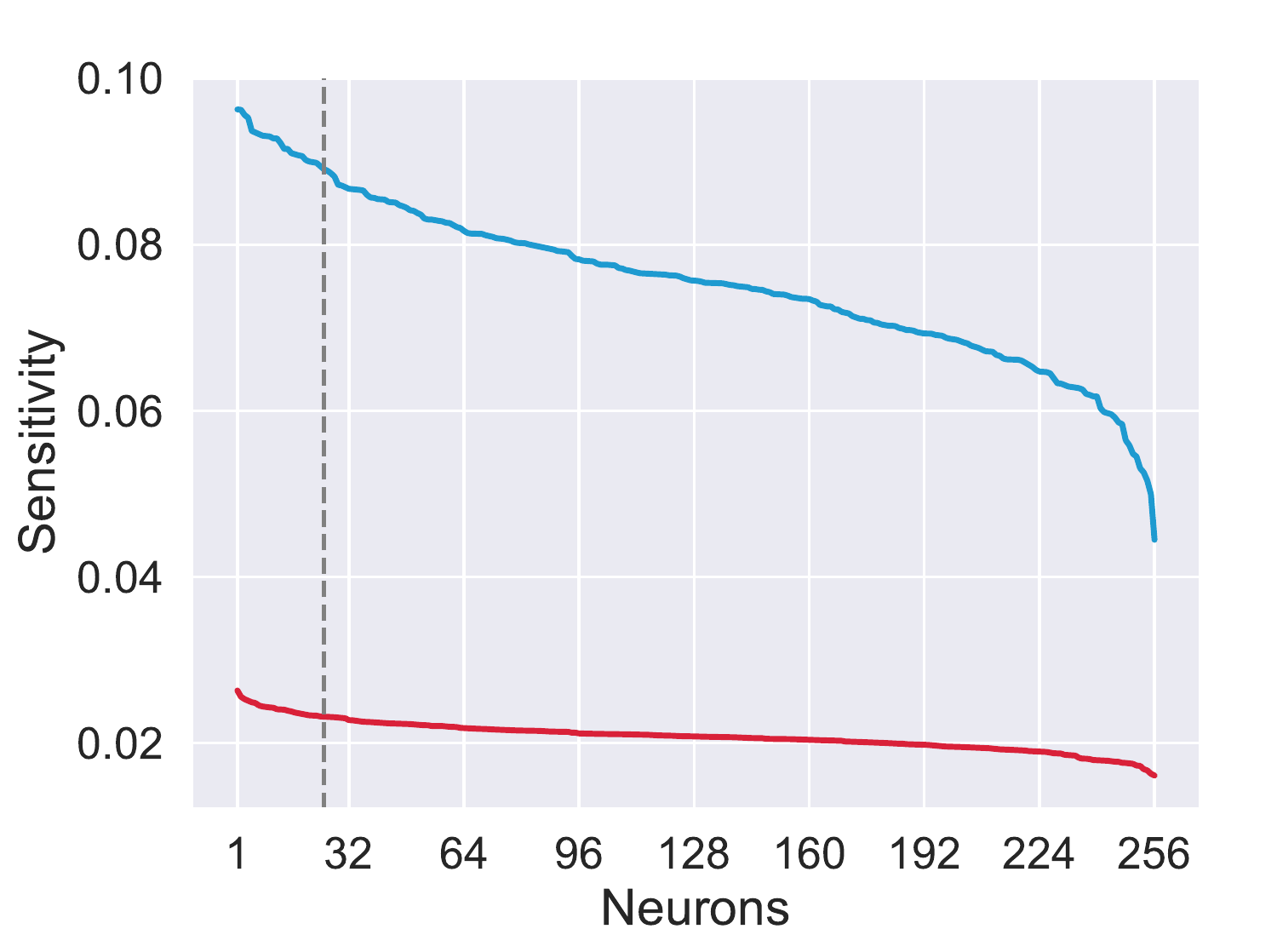}
}
\subfigure[conv13]{
\includegraphics[width=0.225\linewidth]{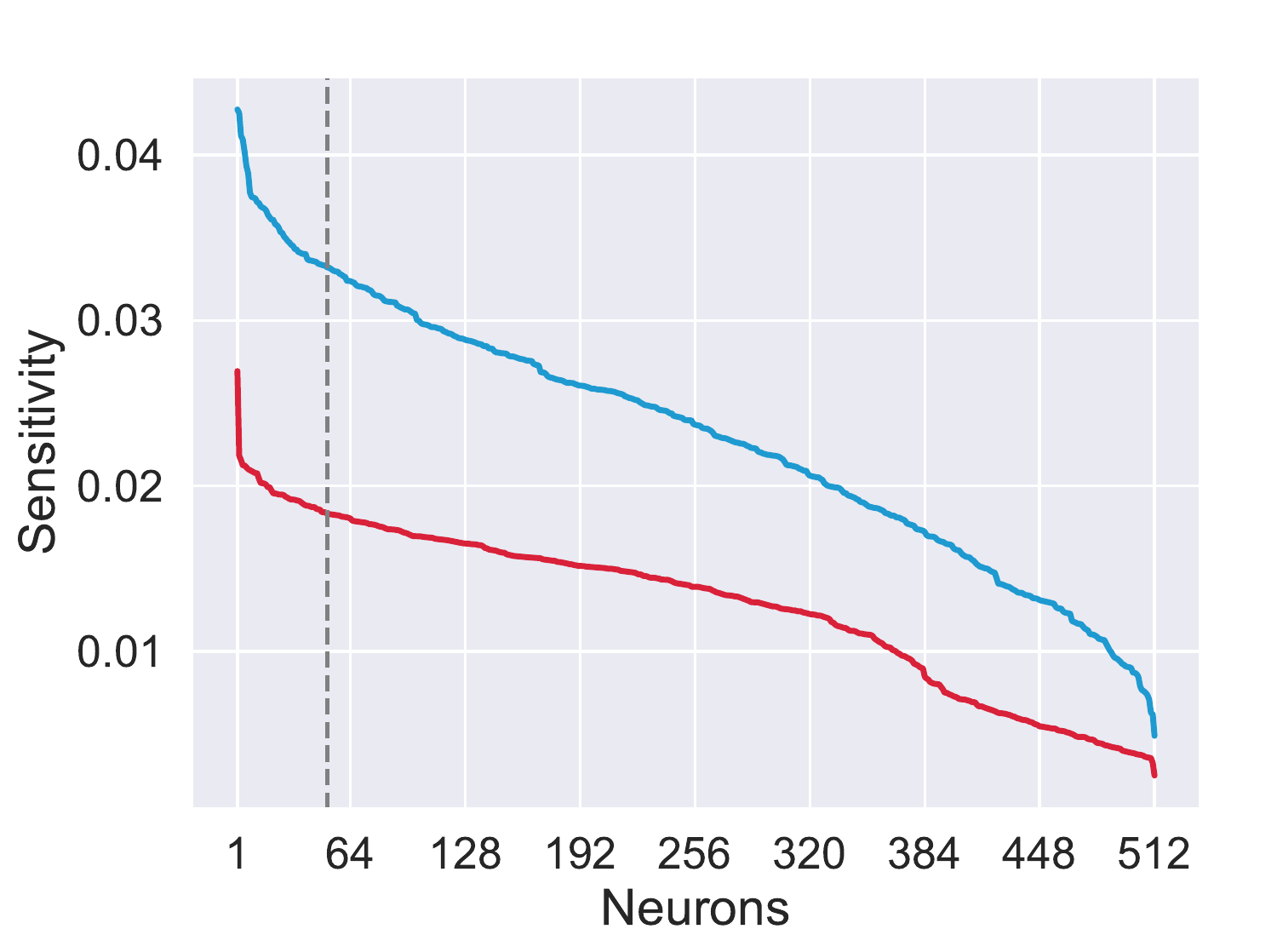}
}
\caption{Neuron sensitivity using $\ell_\infty$ PGD attack on different layers of VGG-16 on CIFAR-10. Subfigure (a) to (d) presents \emph{conv1}, \emph{conv4}, \emph{conv7} and \emph{conv13} respectively. Neurons at the left of the dotted gray line are selected as \emph{Sensitive Neurons}. Neurons of PAT show great insensitivity in contrast to the Vanilla model. Meanwhile, sensitive neurons lead others to a large extent.}
\label{fig:cifar_one_layer_sensitivity}
\end{figure*}

\begin{figure*}[!htb]
\centering
\subfigure[conv2]{
\includegraphics[width=0.225\linewidth]{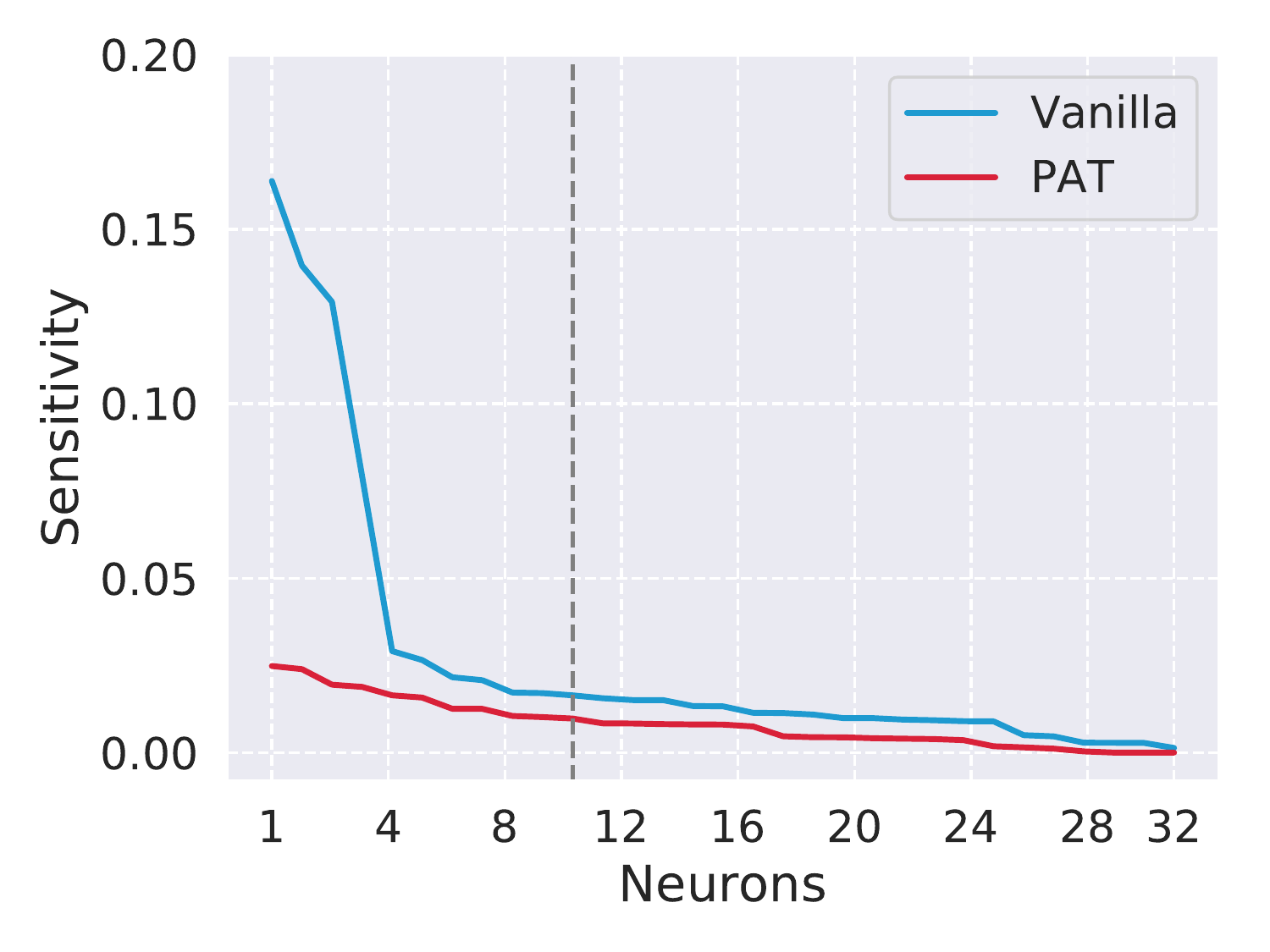}
}
\subfigure[conv5]{
\includegraphics[width=0.225\linewidth]{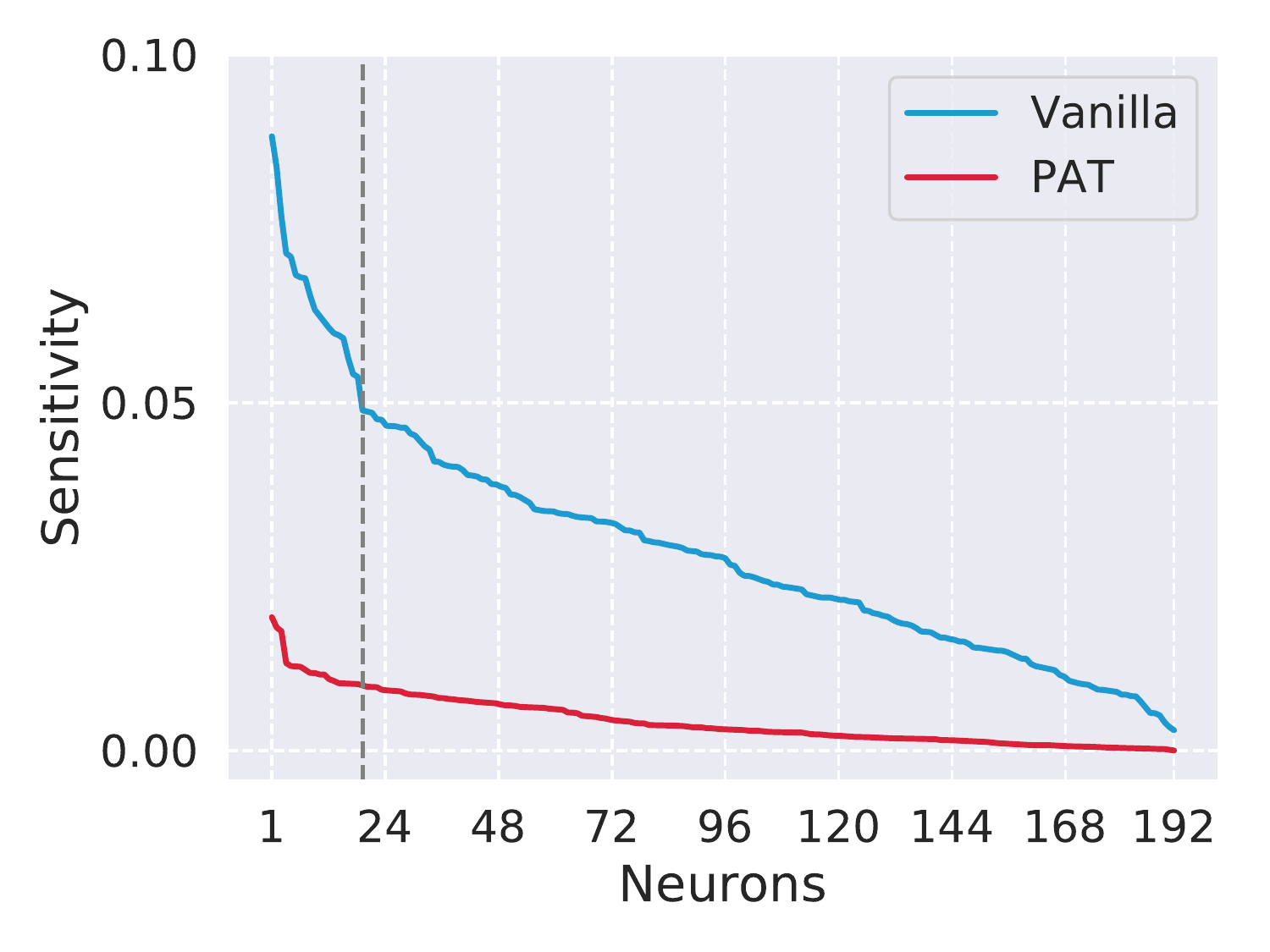}
}
\hspace{0.015\linewidth}
\subfigure[l2b1]{
\includegraphics[width=0.225\linewidth]{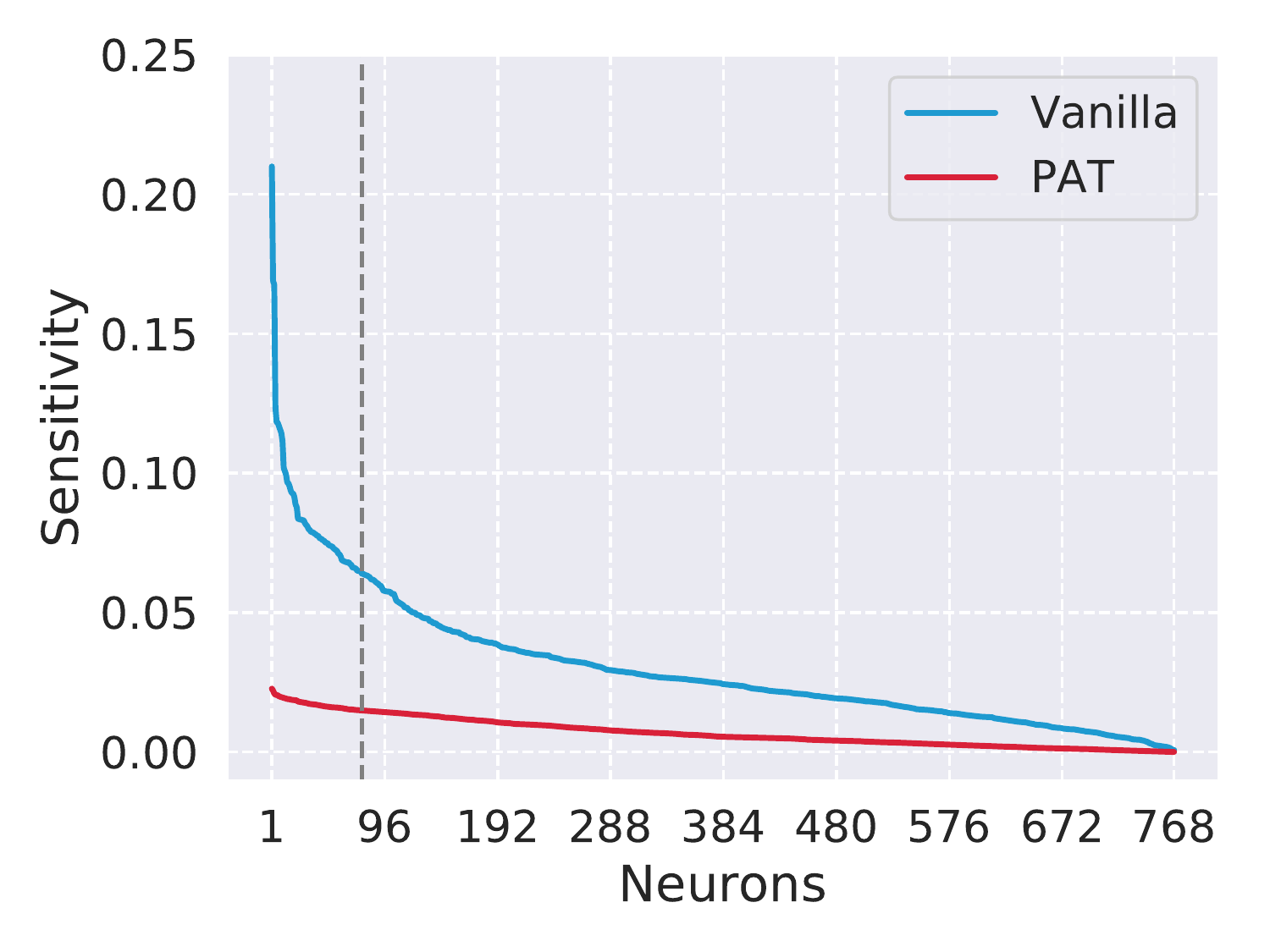}
}
\subfigure[l4b1]{
\includegraphics[width=0.225\linewidth]{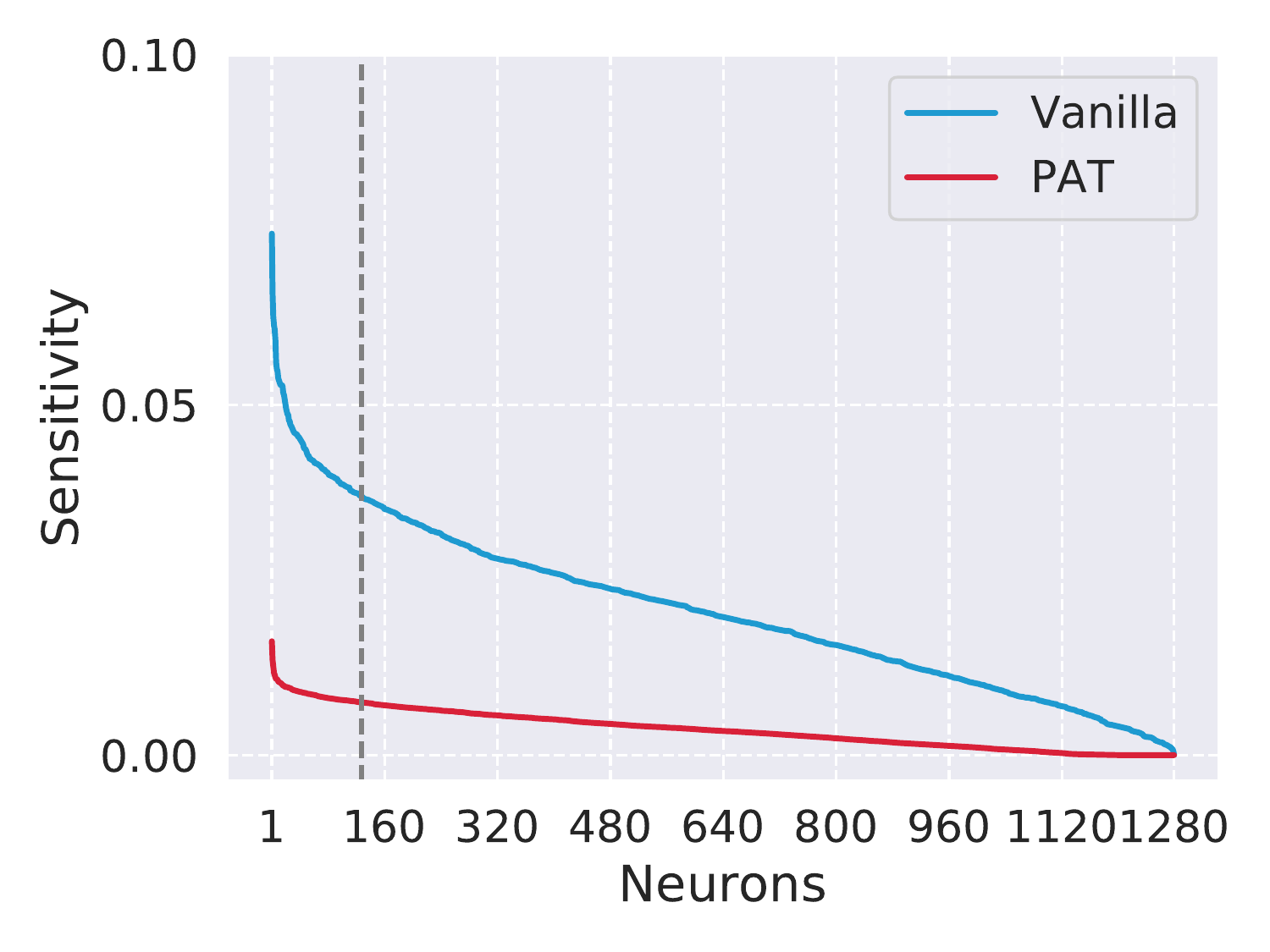}
}
\caption{\textcolor{black}{Neuron sensitivity using $\ell_\infty$ PGD attack on different layers of Inception-V3 on CIFAR-10. Subfigure (a) to (d) presents \emph{conv2}, \emph{conv5}, \emph{l2b1} and \emph{l4b1}, respectively. Neurons at the left of the dotted gray line are selected as \emph{Sensitive Neurons}. We can have same observations as Figure \ref{fig:cifar_one_layer_sensitivity}.}}
\label{fig:cifar_one_layer_sensitivity_inception}
\end{figure*}

\begin{figure*}[!htb]
\centering
\subfigure[l1b1c1]{
\includegraphics[width=0.225\linewidth]{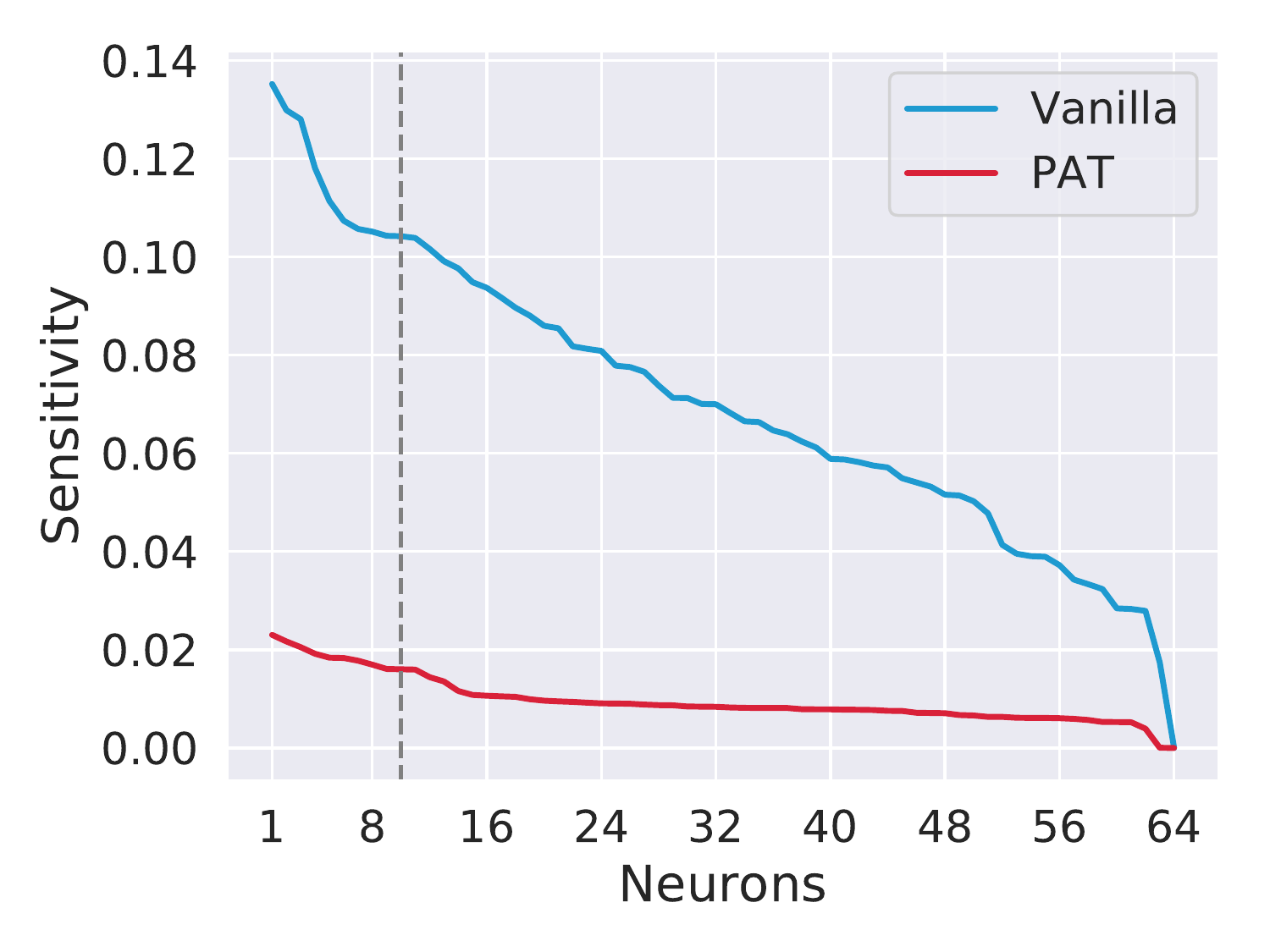}
}
\subfigure[l2b1c1]{
\includegraphics[width=0.225\linewidth]{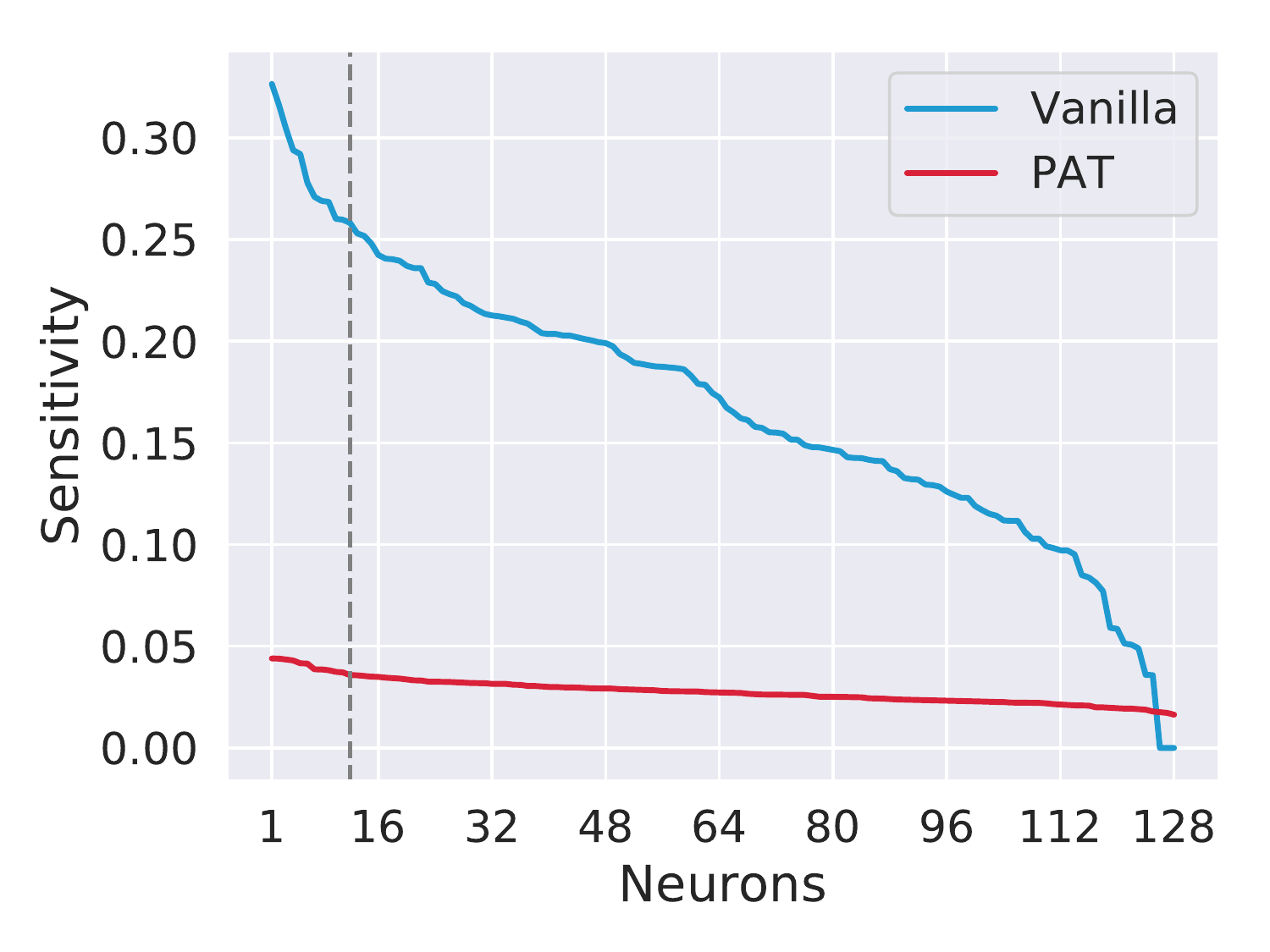}
}
\hspace{0.015\linewidth}
\subfigure[l3b1c2]{
\includegraphics[width=0.225\linewidth]{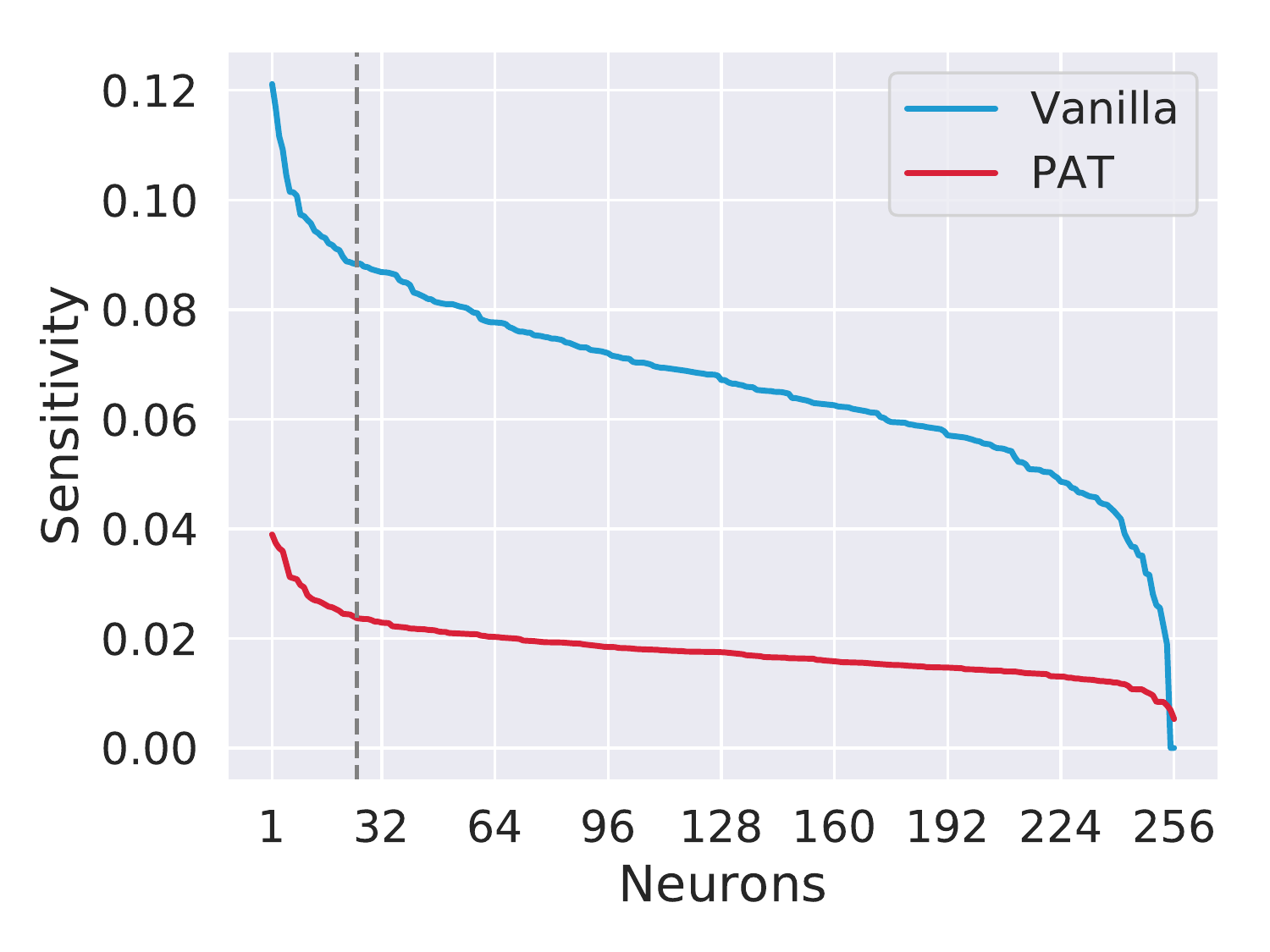}
}
\subfigure[l4b1c1]{
\includegraphics[width=0.225\linewidth]{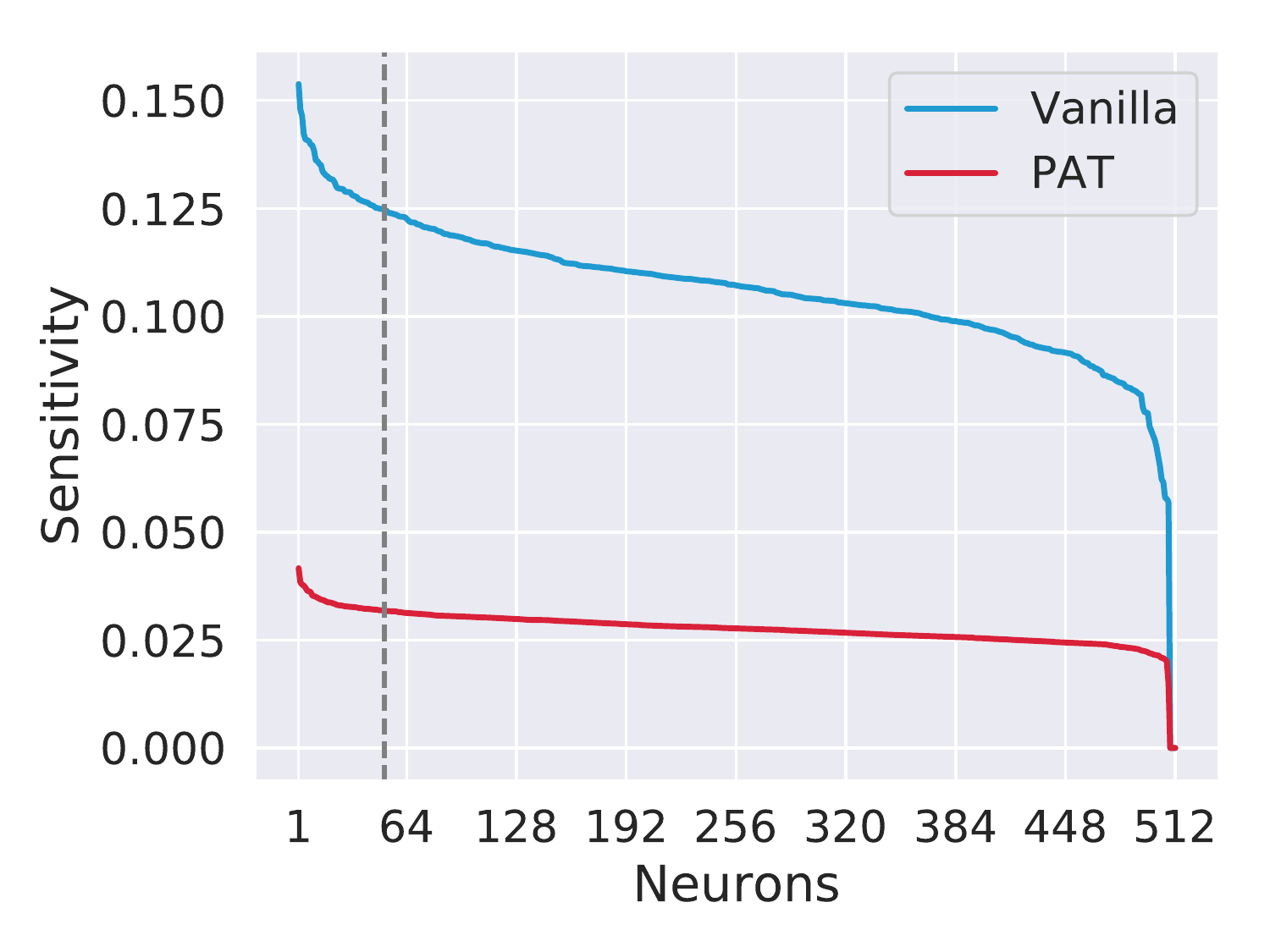}
}
\caption{Neuron sensitivity using $\ell_\infty$ PGD attack on different layers of ResNet-18 on ImageNet. Subfigure (a) to (d) presents \emph{l1b1c1}, \emph{l2b1c1}, \emph{l3b1c2} and \emph{l4b1c1} respectively. Neurons at the left of the dotted gray line are selected as \emph{Sensitive Neurons}. We can have same observations as Figure \ref{fig:cifar_one_layer_sensitivity}.}
\label{fig:imagenet_one_layer_sensitivity}
\end{figure*}

\begin{figure}[!htb]
\centering
\subfigure[$\ell_\infty$ PGD]{
\includegraphics[width=0.70\linewidth]{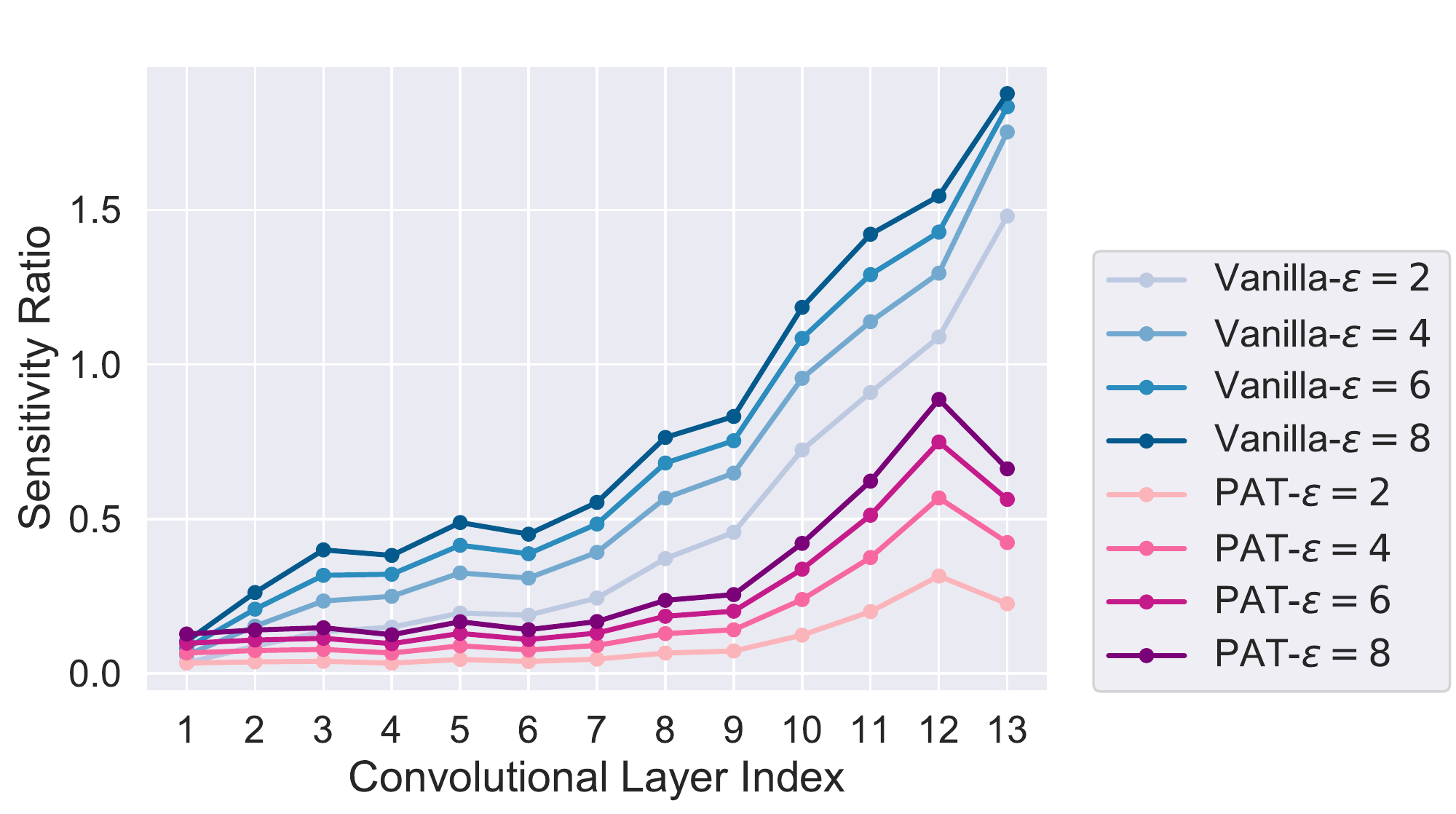}
}
\subfigure[Gaussian Noise]{
\includegraphics[width=0.70\linewidth]{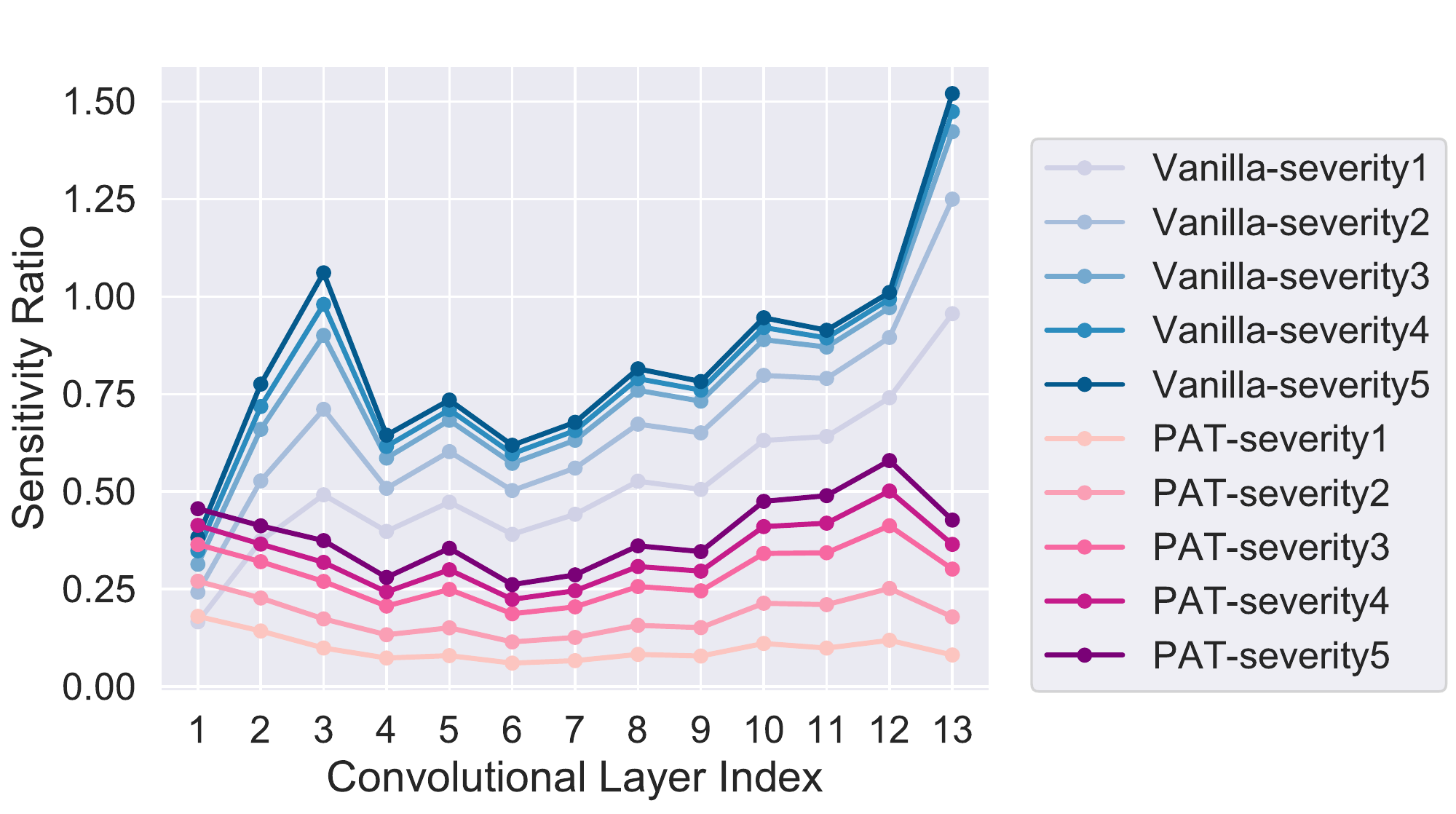}
}
\caption{Mean values of sensitivity ratios on sensitive neurons of all convolutional layers using VGG-16 on CIFAR-10. Subfigure (a) to (b) respectively represent the situation of \emph{PGD} and \emph{Gaussian Noise} with multiple attack power. Different error amplification effects can be seen in different situations.}
\label{fig:cifar-different-noise}
\end{figure}

\begin{figure*}[!htb]
\centering
\subfigure[conv3]{
\includegraphics[width=0.225\linewidth]{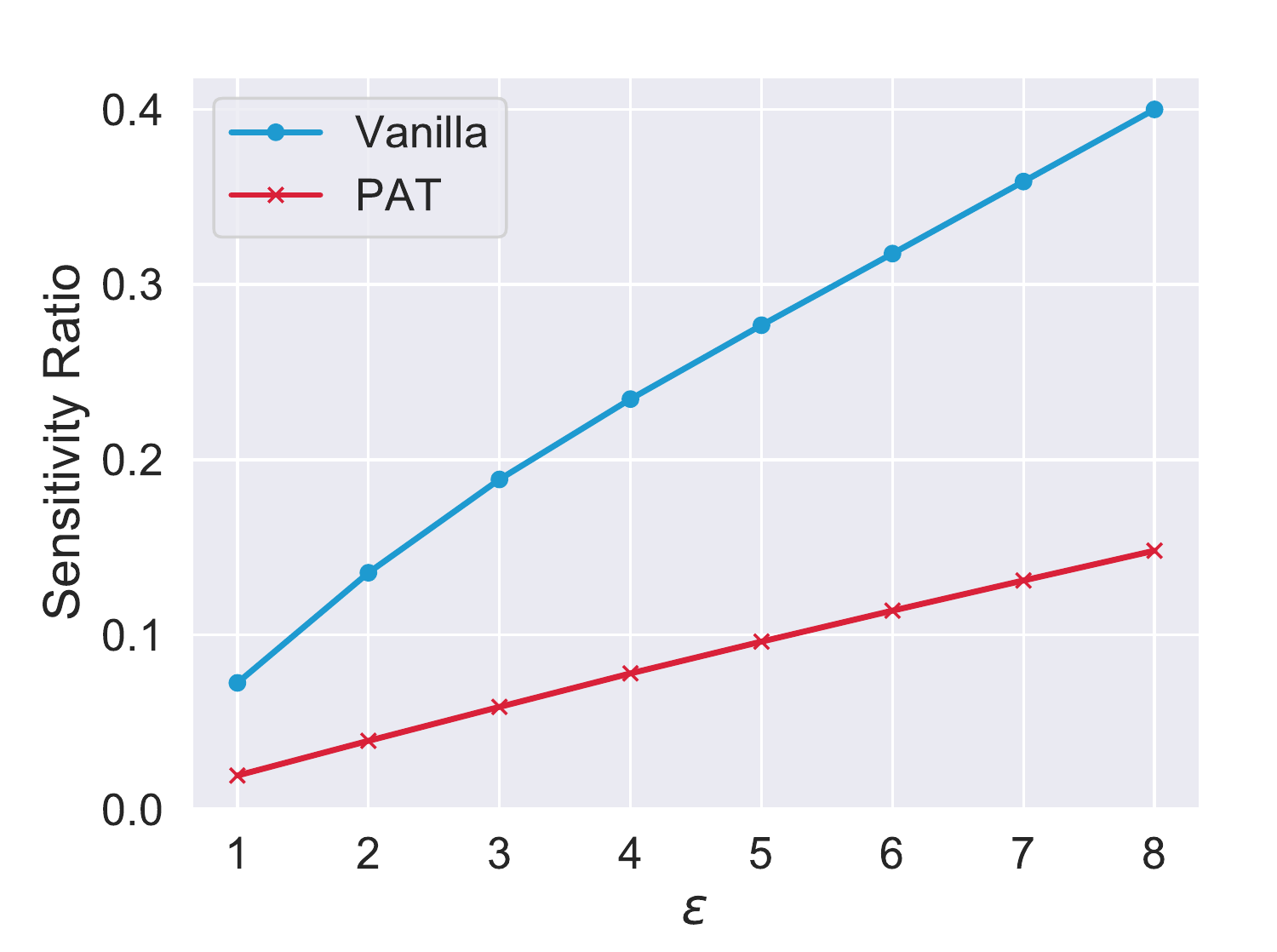}
}
\subfigure[conv7]{
\includegraphics[width=0.225\linewidth]{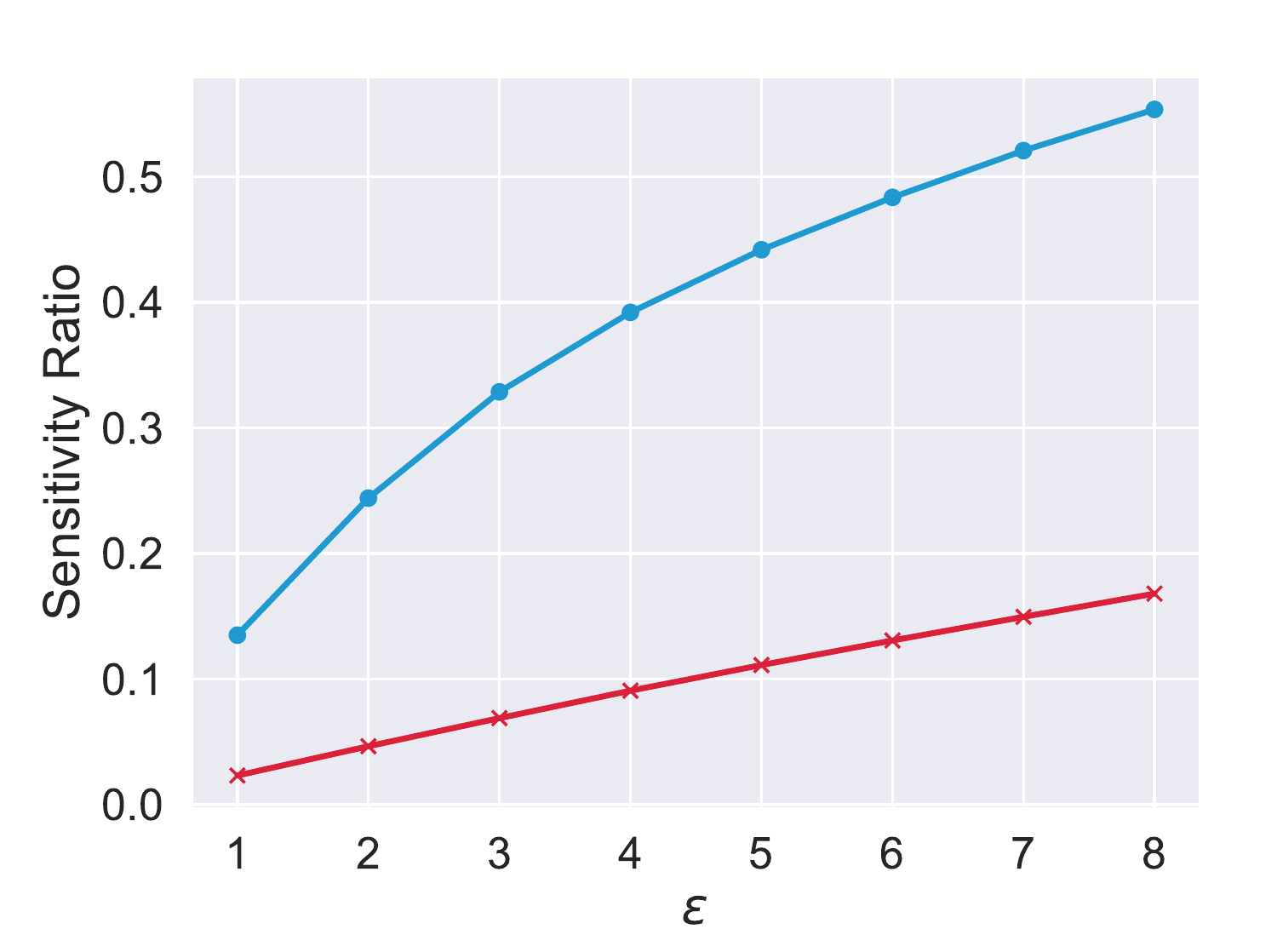}
}
\hspace{0.015\linewidth}
\subfigure[conv10]{
\includegraphics[width=0.225\linewidth]{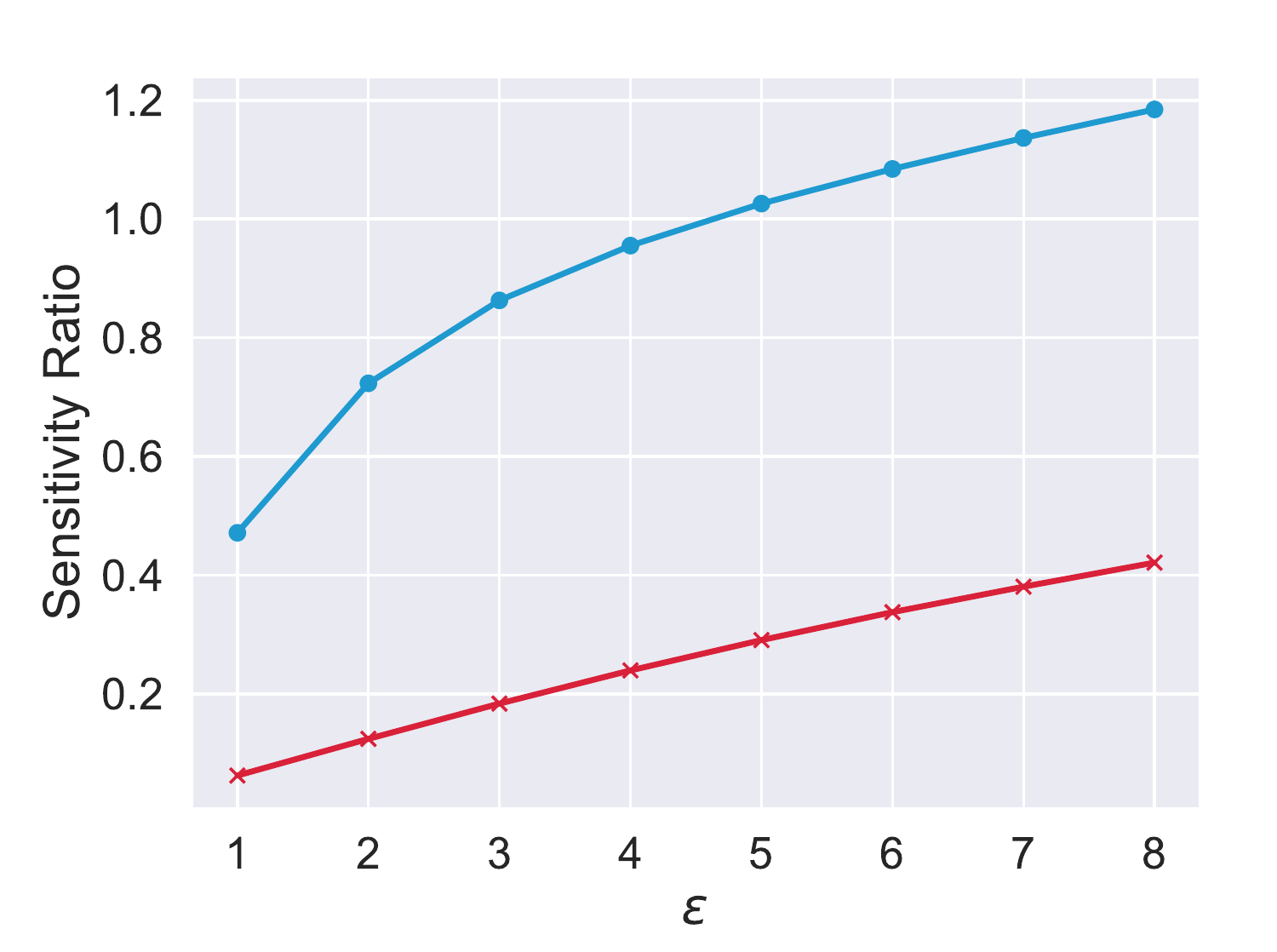}
}
\subfigure[conv13]{
\includegraphics[width=0.225\linewidth]{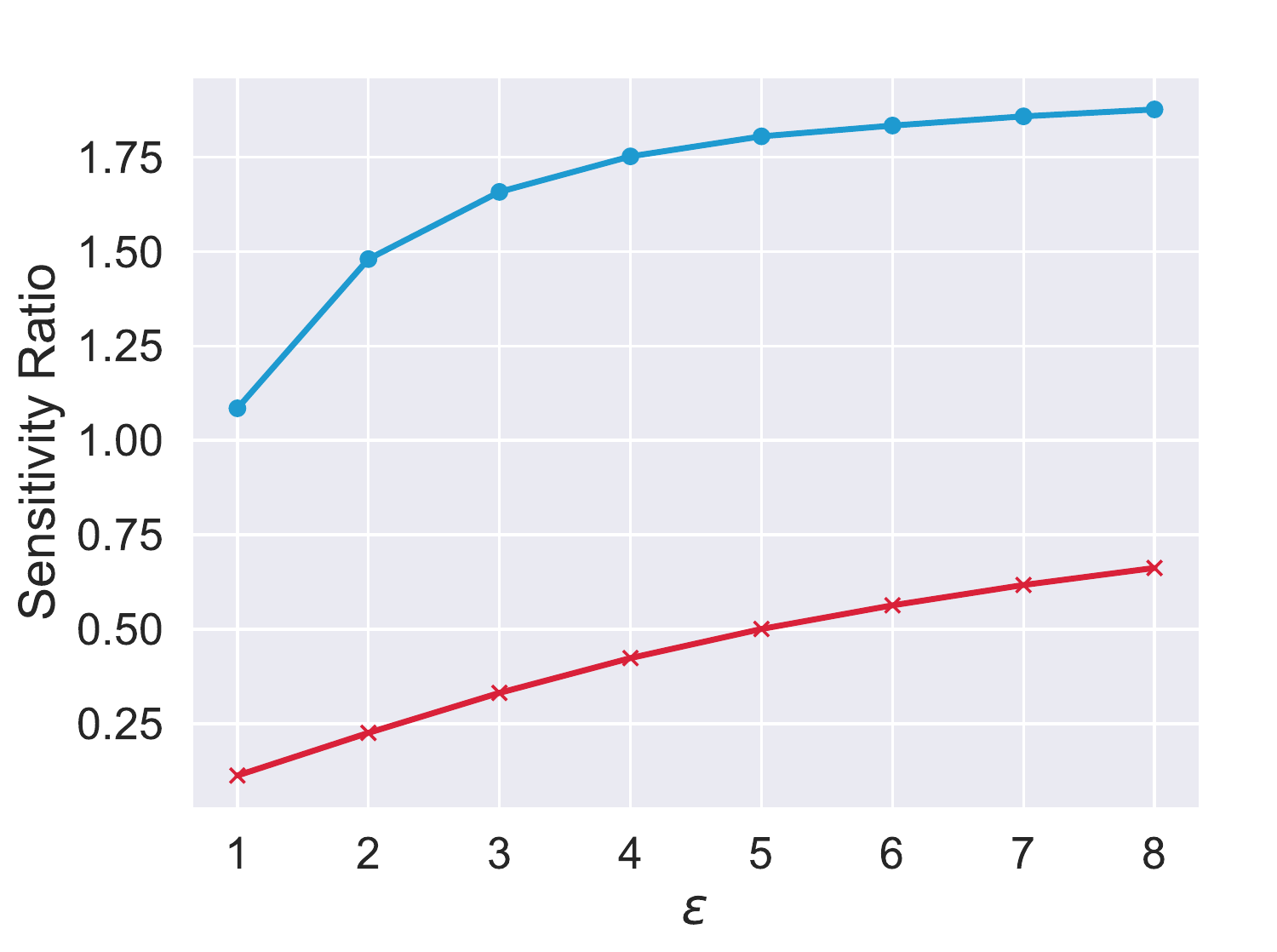}
}
\caption{Relationship between sensitivity ratios and the magnitude of perturbations. We use $\ell_\infty$ PGD attack with VGG-16 on CIFAR-10. Subfigure (a) to (d) respectively represent the situation of \emph{conv3}, \emph{conv7}, \emph{conv10} and \emph{conv13}. For the vanilla model, the positive correlation attenuates gradually as the perturbation propagates through layers, which indicates the denoising function that intermediate layers have. In contrast, correlation is an invariably directly proportional relationship for PAT. And the proportionality coefficient is much lower than vanilla.}
\label{fig:Sensitivity-Eps}
\end{figure*}

\begin{table}[!htb]
\centering
\caption{Sensitivity ratios of VGG-16 CIFAR-10 model using $\ell_\infty$ PGD with different magnitudes. We list the situation of layers from \emph{conv10} to \emph{ReLU13}.}
\begin{center}
\begin{small}
\begin{sc}
\setlength{\tabcolsep}{0.1mm}{
\begin{tabular}{c|c|cccccccc}
\toprule
Model & Layer & $\epsilon=1$ & $\epsilon=2$ & $\epsilon=3$ & $\epsilon=4$ & $\epsilon=5$ & $\epsilon=6$ & $\epsilon=7$ & $\epsilon=8$ \\\hline
\multirow{8}{*}{Vanilla} & conv10 & 0.47 & 0.72 & 0.86 & 0.96 & 1.03 & 1.08 & 1.14 & 1.18\\
~ & ReLU10 & 2.14 & 4.48 & 5.98 & 7.08 & 8.60 & 8.72 & 9.14 & 9.96\\
~ & conv11 & 0.65 & 0.91 & 1.04 & 1.14 & 1.22 & 1.29 & 1.36 & 1.42\\
~ & ReLU11 & 4.26 & 5.84 & 6.67 & 7.63 & 7.92 & 7.98 & 8.83 & 9.12\\
~ & conv12 & 0.83 & 1.09 & 1.21 & 1.30 & 1.37 & 1.43 & 1.49 & 1.54\\
~ & ReLU12 &  5.43 & 20.60 & 18.29 & 19.24 & 18.33 & 20.24 & 17.29 & 20.59\\
~ & conv13 & 1.08 & 1.48 & 1.66 & 1.75 & 1.80 & 1.83 & 1.86 & 1.88\\
~ & ReLU13 & 7.36 & 11.11 & 13.18 & 15.06 & 16.49 & 16.83 & 15.56 & 15.83\\\hline
\multirow{8}{*}{PAT} & conv10 & 0.06 & 0.12 & 0.18 & 0.24 & 0.29 & 0.34 & 0.38 & 0.42\\
~ & ReLU10 & 0.57 & 0.97 & 1.22 & 1.54 & 1.79 & 2.09 & 2.26 & 2.52\\
~ & conv11 & 0.10 & 0.20 & 0.29 & 0.38 & 0.45 & 0.51 & 0.57 & 0.62\\
~ & ReLU11 & 0.68 & 1.12 & 1.50 & 1.89 & 2.24 & 2.56 & 2.88 & 3.22\\
~ & conv12 & 0.16 & 0.32 & 0.45 & 0.57 & 0.67 & 0.75 & 0.82 & 0.89\\
~ & ReLU12 & 0.48 & 1.08 & 1.69 & 2.37 & 2.98 & 3.34 & 3.76 & 4.05\\
~ & conv13 & 0.11 & 0.23 & 0.33 & 0.42 & 0.50 & 0.56 & 0.62 & 0.66\\
~ & ReLU13 & 0.82 & 1.78 & 3.45 & 4.03 & 4.34 & 5.07 & 5.65 & 8.24\\
\bottomrule
\end{tabular}}
\end{sc}
\end{small}
\end{center}
\label{table:eps_ratio_layer}
\end{table}

\section{Improving Robustness with Sensitive Neurons} \label{Section:Improving Robustness with Sensitive Neurons}
We have demonstrated the close connections between neuron sensitivity and adversarial robustness, as well as the basic properties of sensitive neurons in deep models. With the above observations and conclusions, it is natural to improve model robustness by stabilizing those sensitive neurons. Therefore, in this section, we first try to explore the reasons why state-of-the-art adversarial training strategies achieve strong robustness from the view of sensitive neurons. Then, we \textcolor{black}{propose} a strategy called \emph{Sensitive Neuron Stabilizing} (SNS) to alleviate the hazard brought by adversarial examples and improve model robustness by reducing the sensitivity of sensitive neurons.

\subsection{Empirical Settings}.
For the adversarial attacks, we use FGSM, $\ell_2$ PGD, $\ell_\infty$ PGD, and C\&W. Their implementation details follow Section \ref{section:first setting}. As for \textbf{adversarial defense models}, we choose the state-of-the-art defense methods including PGD-based adversarial training (PAT) \cite{DBLP:conf/iclr/MadryMSTV18} and adversarial logit pairing (ALP) \cite{DBLP:journals/corr/abs-1803-06373}. The specific settings are listed as follow:

\begin{itemize}
\item \textbf{PAT.}\quad $\ell_\infty$ PGD attack on current training model is used for adversarial examples generating with $\epsilon = 8, k = 10, \alpha = \epsilon / \sqrt{k}$. The clean and adversarial examples are mixed in the ratio of 1 to 1 during the training phase.
\item \textbf{ALP.}\quad The loss terms of ALP consist of two main parts: adversarial training loss whose setting is consistent with PAT and logit pairing term which is implemented with $\ell_{2}$ loss. The ratio of these two terms during the training phase is 2 to 1, namely the coefficient $\lambda$ of logit pairing term is 0.5.
\end{itemize}

\subsection{Adversarial Training Builds Robust Models by Reducing Neuron Sensitivities}

With the increasing concerns of adversarial examples to model robustness, plenty of adversarial defense methods have been proposed including adversarial training, input transformation, etc. However, as discussed in \cite{athalye2018obfuscated}, most of these defense strategies just give a false sense of safety, which could be attacked easily through obfuscated gradient circumvent. Whereas, adversarial training based methods, which augment training data with adversarial examples, are relatively immune to these attacks and achieve the most robust models so far. Based on that, a number of works have been proposed to study and explain the essence of adversarial training to model robustness. \cite{DBLP:journals/corr/GoodfellowSS14} first introduced the adversarial training strategy to defense adversarial attack by somewhat reducing the linearity for high-dimensional DNNs. \cite{shaham2018understanding} tried to explain the performance of adversarial training from the view of robust optimization theory, which improves model robustness by increasing local stability.

Different from them, in this part, attempts have been made to interpret adversarial training from the perspective of neuron sensitivity. One important take-away is: \emph{adversarial training improves model robustness by embedding representation insensitivities}.

To demonstrate this point, in the beginning, we respectively trained a Vanilla and a PGD-based adversarial training (PAT) model using VGG-16 on CIFAR-10. Test results of trained models are illustrated in Table \ref{table:pgd_attack_acc}.
Then, we adopt $\ell_\infty$ PGD untargeted attack for its powerful strength and adversarial examples are generated on the validation set. Finally, we calculate the \emph{Neuron Sensitivity} and obtain the \emph{Sensitive Neuron} of each layer.

Figure \ref{fig:cifar_one_layer_sensitivity} presents neuron sensitivity on different layers. Compared with the Vanilla model, insensitivity towards adversarial examples can be captured in all neurons of PAT. This observation clarifies the reason why adversarial training methods are insensitive or robust to adversarial noises. Meanwhile, we observe prominent sensitivity gaps between the top 10\% neurons of each model, which proves the rationality of proposed sensitive neurons. In other words, sensitive neurons are significant indicators to represent model behaviors in the adversarial setting between benign and adversarial examples. Moreover, we also conduct experiment \textcolor{black}{on CIFAR-10 with Inception-V3 (Figure \ref{fig:cifar_one_layer_sensitivity_inception}) and} on ImageNet with ResNet-18 (Figure \ref{fig:imagenet_one_layer_sensitivity}), which conveys the same conclusion.

After neuron-wise analysis, investigation has been made in an overall view of different layers. It has been discussed that imperceptible perturbations will be amplified during the propagation \cite{liao2018defense}. Therefore, we decide to study the error amplification effect in DNNs using the sensitivity metric. To eliminate the magnitude difference among layers, we further propose a standardized version of sensitivity named \emph{Sensitivity Ratio}: %\aishan{duplicated notation with Eqn(1)}
\begin{equation}
\sigma_{Ratio}(F_l^m,\bar{\mathbf{D}})=\frac{1}{\mathit{N}}\sum_{i=1}^\mathit{N}\frac{\Vert F_l^m(x_i)-F_l^m(x_i^\prime)\Vert_1}{\Vert F_l^m(x_i)\Vert_1} .\label{Neuron Sensitivity Ratio}
\end{equation}

Actually, \emph{Sensitivity Ratio} measures the deviation proportion between $F_l^m(x_i)$ and $F_l^m(x_i^\prime)$.

The mean values of neuron sensitivity ratios of all convolutional layers are shown in Figure \ref{fig:cifar-different-noise}. For $\ell_\infty$ PGD adversarial attack, we could see an obvious error amplification effect, in which the sensitivity ratio increases layer by layer. This explains that adversarial examples lead to the wrong model decisions by elaborately making the intermediate features gradually deviate from clean examples during the forward propagation. In addition, the accelerating of the sensitivity ratio of the vanilla model is faster than PAT. As for Gaussian noises, we see a sensitivity ratio summit of the vanilla model at the 3-rd layer. Then, after a sudden drop, the sensitivity ratio increases continually during the propagation. Note that, the summit may indicate that bottom low-level feature extractors are sensitive to Gaussian noises (high-frequency random noise, \cite{yin2019fourier}), which is different from adversarial attacks. The above findings clearly demonstrate that adversarial noises and Gaussian noises are more easily absorbed by PAT during the forward propagation process due to the stable neurons leading to consistent model behaviors and strong robustness.

Additionally, we also investigate the relationship between neuron sensitivity ratios and the magnitude of adversarial perturbations. Results are shown in Figure \ref{fig:Sensitivity-Eps} and Table \ref{table:eps_ratio_layer}, from which we have the following conclusions: (1) There exists a positive correlation between neuron sensitivity and perturbation magnitude $\epsilon$ among all model intermediate layers. (2) For the vanilla model, this correlation attenuates gradually as the perturbation propagates through the layers. This indicates the denoising function that intermediate layers have. In contrast, the correlation is invariably directly proportional relationship among all layers of PAT. And the proportionality coefficient is much lower than vanilla. (3) A surprising phenomenon emerges in Table \ref{table:eps_ratio_layer} that sensitivity ratios of ReLU layers invariably boom from former convolutional layers and convolutional layers behave conversely. Hence we draw a conclusion that ReLU layers lead to the error amplification effect, though the embedded convolutional layers have the contrary ability.

\subsection{Training Adversarially Robust Models via Sensitive Neurons Stabilizing} \label{section:sns}
Motivated by our observations, a straightforward idea for improving model robustness is to force the sensitive neurons of benign and adversarial ones to behave similarly. In other words, we try to stabilize those sensitive neurons, and thus the whole model will be insensitive to adversarial examples. The \emph{Sensitive Neurons Stabilizing} (SNS for short) can be easily accomplished by directly adding a loss term to measure the similarities of sensitive neurons behaviors when inputting the clean and adversarial examples:
\begin{equation}
\mathcal{L}_{sns}(x,x^\prime;\theta) = \sum_{l}^{l\in \mathbf{S}}\sum_{m}^{F_l^m\in \Omega_{l}}\frac{1}{dim(F_l^m(x))}\Vert F_l^m(x)-F_l^m(x^\prime)\Vert_{1},\label{equa:sensitivity loss}
\end{equation}
where $\mathbf{S}$ denotes the indices of selected layers and $\Omega_{l}$ denotes \emph{sensitive neuron} set of layer $l$. Again, $dim(\cdot)$ denotes the dimension of a vector.

\begin{algorithm}[!htb]
   \caption{Improve model robustness with \emph{Sensitive Neurons Stabilizing}}
   \label{alg:Finetune Vanilla}
\hspace*{0.02in} {\bf Input:} training set $\mathbf{D}$ with $\mathit{N}$ samples, Vanilla model $F_{Vanilla}$\\
\hspace*{0.02in} {\bf Output:} robust model $F_{Robust}$\\
\hspace*{0.02in} {\bf Hyper-parameter:} $\lambda$, batchsize $\mathit{B}$ and epoch $\mathit{E}$\\
\begin{algorithmic}[1]
   \STATE Use PGD white-box attack to generate dual pair set $\bar{\mathbf{D}} = \{(x_i,x_i^\prime)\ |\ i=1,\ldots,\mathit{N}\}$ and select \emph{Sensitive Neurons}.\\
   \FOR{$\mathit{E}$ training epochs}
   %\FOR{$i$ in $\lfloor \frac{N}{m} \rfloor$}
   \FOR{$\lfloor \frac{\mathit{N}}{\mathit{B}} \rfloor$ mini-batch numbers}
   \STATE optimize the current model by \\ $ \mathcal{L}_{adv}(x,x^\prime,y;\theta)+ \lambda \mathcal{L}_{sns}(x,x^\prime;\theta)$
   \ENDFOR
   \ENDFOR
\end{algorithmic}
\end{algorithm}

\begin{table*}[!htb]
\centering
\caption{Experiments results (\%) of adversarial robustness test on CIFAR-10 with VGG-16. Our strategy outperforms all comparison methods towards adversarial examples generated by different attack methods. \draft{We use the bold format to highlight the best result in each case.}}
\begin{center}
\begin{small}
\begin{sc}
\setlength{\tabcolsep}{0.5mm}{
\begin{tabular}{c|c|cccc|cc}
\toprule
\multirow{2}{*}{Model} & \multirow{2}{*}{Clean} & \multicolumn{4}{c|}{White-box Attack} & \multicolumn{2}{c}{\draft{Black-box Attack}}\\ \cline{3-8}
~ & ~ & C\&W & $\ell_2$ PGD & $\ell_\infty$ PGD & FGSM & \draft{SPSA} & \draft{NAttack}\\ \hline
Vanilla & \textbf{91.1} & 0.0 & 28.9 & 0.1 & 29.2 & \draft{21.5} & \draft{0.0}\\
PAT & 85.1 & 13.4 & 68.3 & 37.4 & 49.6 & \draft{67.9} & \draft{42.6}\\
ALP & 84.4 & 13.8 & 69.1 & 36.5 & 48.5 & \draft{67.5} & \draft{43.1}\\ \hline
\textbf{SNS$_{dyn}^{adv}$} & 85.8 & 14.6 & \textbf{71.1} & 39.4 & \textbf{51.2} & \draft{69.3} & \draft{44.0}\\
\textbf{SNS$_{all}^{adv}$} & 85.2 & 14.3 & 70.0 & 37.9 & 50.4 & \draft{68.4} & \draft{43.4}\\
\textbf{SNS$_{rand}^{adv}$} & 83.5$\pm$0.2 & 13.8$\pm$0.1 & 68.7$\pm$0.4 & 35.5$\pm$0.3 & 49.5$\pm$0.4 & \draft{68.1$\pm$0.3} & \draft{43.3$\pm$0.3}\\
\textbf{SNS$_{sen}^{adv}$} & 86.0 & \textbf{15.3} & 71.0 & \textbf{39.6} & 51.0 & \textbf{\draft{69.4}} & \textbf{\draft{44.5}}\\
%Vanilla & 91.12\% & 0.00\% & 28.94\% & 0.14\% & 29.22\%\\
%PAT & 85.09\% & 13.44\% & 68.25\% & 37.36\% & 49.63\%\\
%ALP & 84.41\% & 13.83\% & 69.14\% & 36.50\% & 48.47\%\\
%\textbf{SNS$_{sen}^{adv}$} & 85.98\% & \textbf{15.29\%} & \textbf{71.04\%} & \textbf{39.57\%} & \textbf{51.02\%}\\
\bottomrule
%finetune hyperparameter:(optimize = conv8+conv9+conv10+conv13, use =conv7\_relu+conv8\_relu+conv9\_relu+conv10\_relu+conv11\_relu+conv12\_relu+conv13\_relu+pool5{[}top10{]}, alpha = 5.0)
\end{tabular}}
\end{sc}
\end{small}
\end{center}
\label{table:finetune vanilla}
\end{table*}

\begin{table*}[!htb]
\centering
\caption{Experiments results (\%) of adversarial robustness test on ImageNet with ResNet-18. Our strategy outperforms all comparison methods towards adversarial examples generated by different attack methods. \draft{We use the bold format to highlight the best result in each case.}}
\begin{center}
\begin{small}
\begin{sc}
\setlength{\tabcolsep}{0.5mm}{
\begin{tabular}{c|c|cccc|cc}
\toprule
\multirow{2}{*}{Model} & \multirow{2}{*}{Clean} & \multicolumn{4}{c|}{White-box Attack} & \multicolumn{2}{c}{\draft{Black-box Attack}}\\ \cline{3-8}
~ & ~ & C\&W & $\ell_2$ PGD & $\ell_\infty$ PGD & FGSM & \draft{SPSA} & \draft{NAttack}\\ \hline
Vanilla & \textbf{73.0} & 0.0 & 31.4  & 0.0 & 2.3 & \draft{11.1} & \draft{0.0}\\
PAT & 63.4  & 5.8 & 58.4 & 11.4 & 19.4 & \draft{36.4} & \draft{29.7}\\
ALP & 62.7  & 6.1 & 58.6 & 11.2 & 18.9 & \draft{35.9} & \draft{29.6}\\ \hline
\textbf{SNS$_{sen}^{adv}$} & 64.0 & \textbf{6.9}  & \textbf{59.4} & \textbf{13.0} & \textbf{21.3} & \textbf{\draft{37.2}} & \textbf{\draft{31.5}}\\
%Vanilla & 72.95\% & 0.00\% & 31.39\%  & 0.04\% & 2.32\%\\
%PAT & 63.37\%  & 5.79\% & 58.38\% & 11.38\% & 19.37\%\\
%ALP & 62.74\%  & 6.06\% & 58.60\% & 11.15\% & 18.93\%\\
%\textbf{SNS$_{sen}^{adv}$} & 63.95\% & \textbf{6.93\%}  & \textbf{59.42\%} & \textbf{12.97\%} & \textbf{21.26\%}\\
\bottomrule
%finetune hyperparameter:(optimize = conv8+conv9+conv10+conv13, use =conv7\_relu+conv8\_relu+conv9\_relu+conv10\_relu+conv11\_relu+conv12\_relu+conv13\_relu+pool5{[}top10{]}, alpha = 5.0)
\end{tabular}}
\end{sc}
\end{small}
\end{center}
\label{table:finetune ResNet}
\end{table*}

%\begin{table}[!htb]
%\centering
%\caption{\textcolor{black}{Ablation study results (\%) of effectiveness of selecting sensitive neurons for neurons stabilization in adversarial robustness improvement. All models are trained on CIFAR-10 with VGG-16.}}
%\begin{center}
%\begin{small}
%\begin{sc}
%\setlength{\tabcolsep}{0.5mm}{
%\begin{tabular}{c|c|cccc}
%\toprule
%Model & Clean & C\&W & $\ell_2$ PGD & $\ell_\infty$ PGD & FGSM\\ \hline
%Vanilla & 91.1 & 0.0 & 28.9 & 0.1 & 29.2\\
%PAT & 85.1 & 13.4 & 68.3 & 37.4 & 49.6\\
%SNS$_{sen}^{adv}$ & 86.0 & \textbf{15.3} & \textbf{71.0} & \textbf{39.6} & \textbf{51.0}\\
%SNS$_{all}^{adv}$ & 85.2 & 14.3 & 70.0 & 37.9 & 50.4\\
%SNS$_{rand}^{adv}$ & 83.5$\pm$0.2 & 13.8$\pm$0.1 & 68.7$\pm$0.4 & 35.5$\pm$0.3 & 49.5$\pm$0.4\\
%Model & Clean & C\&W & $\ell_2$ PGD & $\ell_\infty$ PGD & FGSM\\ \hline
%Vanilla & 91.12 & 0.00 & 28.94 & 0.14 & 29.22\\
%PAT & 85.09 & 13.44 & 68.25 & 37.36 & 49.63\\
%SNS$_{sen}^{adv}$ & 85.98 & \textbf{15.29} & \textbf{71.04} & \textbf{39.57} & \textbf{51.02}\\
%SNS$_{all}^{adv}$ & 85.16 & 14.34 & 69.97 & 37.85 & 50.39\\
%SNS$_{rand}^{adv}$ & 83.52$\pm$0.23 & 13.81$\pm$0.14 & 68.74$\pm$0.35 & 35.52$\pm$0.27 & 49.54$\pm$0.39\\
%\bottomrule
%finetune hyperparameter:(optimize = conv8+conv9+conv10+conv13, use =conv7\_relu+conv8\_relu+conv9\_relu+conv10\_relu+conv11\_relu+conv12\_relu+conv13\_relu+pool5{[}top10{]}, alpha = 5.0)
%\end{tabular}}
%\end{sc}
%\end{small}
%\end{center}
%\label{table:Ablation VGG}
%\end{table}

Given clean examples $x$ and adversarial examples $x^\prime$, our SNS method minimizes the following loss:
\begin{equation}
\mathcal{L}_{adv}(x,x^\prime,y;\theta) + \lambda  \mathcal{L}_{sns}(x,x^\prime;\theta),\label{finetune_adv}
\end{equation}
where $\mathcal{L}_{adv}$ denotes the adversarial training loss and $\mathcal{L}_{sns}$ represents the similarity of sensitive neurons' behaviors in some specific layers for the dual pair $(x,x^\prime)$. Here, $\lambda$ is a hyper-parameter balancing the two loss terms. The whole training process is demonstrated in Algorithm \ref{alg:Finetune Vanilla}.

\begin{figure}[!htbp]
	\centering
	%\vspace{-0.3in}
	\includegraphics[width=0.60\linewidth]{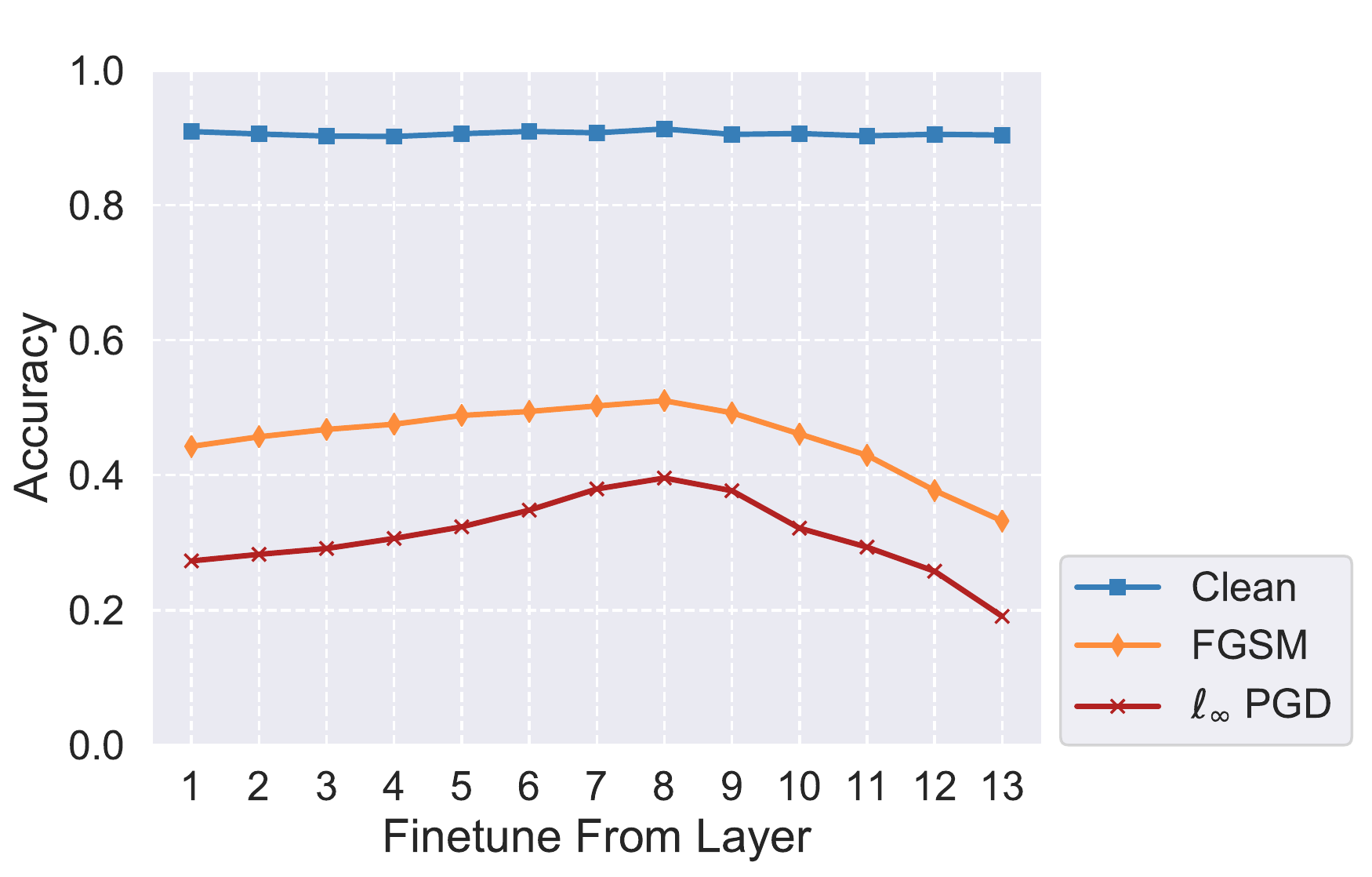}
	%\vspace{-.4in}
	\caption{VGG-16 model trained using Equation \ref{finetune_adv} on CIFAR-10 when applying top-$k$ layers. SNS training with layers from $conv8$ to $conv13$ enjoys the strongest model against all kinds of attacks.}
    \label{fig:top-k layer}
%\vspace{-.15in}
\end{figure}

As there are numerous layers in the architecture of deep models, a question emerges: \emph{do we need to select all layers when performing SNS training?} Since different hidden layers behave variously from each other \cite{DBLP:journals/corr/abs-1902-01996}, it seems necessary to come up with a strategy for choosing the desired hidden layers. As discussed before that sensitive neurons that are more close to predictions reveal more adversarial weaknesses towards adversarial attacks. Meanwhile, they contain more high-level semantic information contributing to the model's final decisions. Thus, we conduct experiments of training models with top-$k$ hidden layers to figure out the choices for layer selection. Namely, parameters of the bottom several layers are locked during finetuning. As illustrated in Figure \ref{fig:top-k layer}, using layers from $conv8$ to $conv13$ reaches the most adversarially robust model for VGG-16 on CIFAR-10. Similarly, we find using \emph{l3b2c1} to \emph{l4b2c2} reaches the best result. Thus, the rest of the paper follows this guidance.

Here we clarify the experiment settings of SNS. $\ell_\infty$ PGD attack is used for generating adversarial example set from the training set, based on which sensitive neurons are selected. Then we train models with different hyper-parameters on different datasets. On CIFAR-10 with VGG-16, we use the sensitive neurons from \emph{conv8} to \emph{conv13} with $\lambda = 5.0$ in the training set. On ImageNet with ResNet-18, we use the sensitive neurons from \emph{l3b2c1} to \emph{l4b2c2} with $\lambda = 6.0$ in the training set.

\begin{figure}[!htb]
\centering
\subfigure[]{
\includegraphics[width=0.46\linewidth]{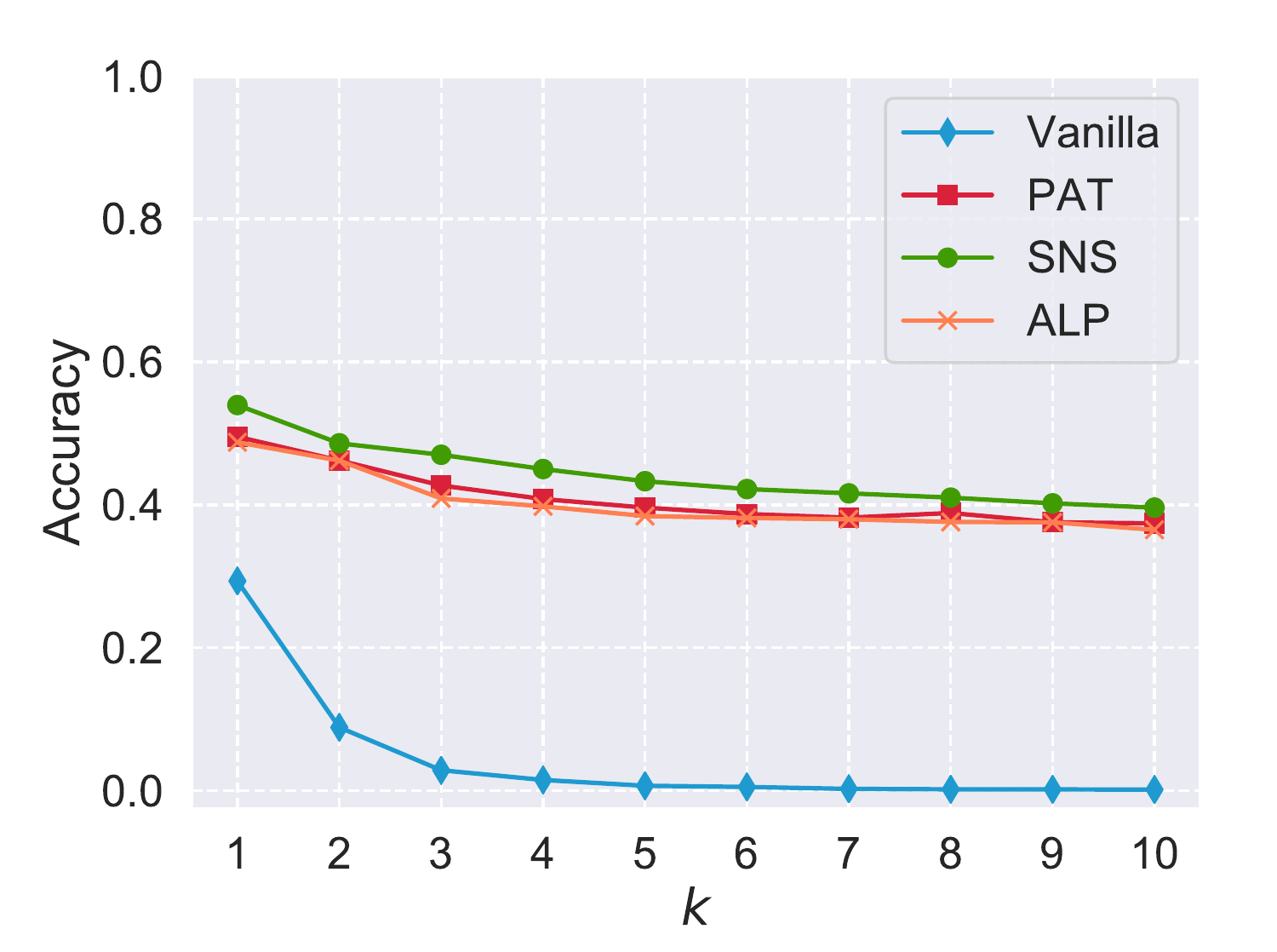}
}
\subfigure[]{
\includegraphics[width=0.46\linewidth]{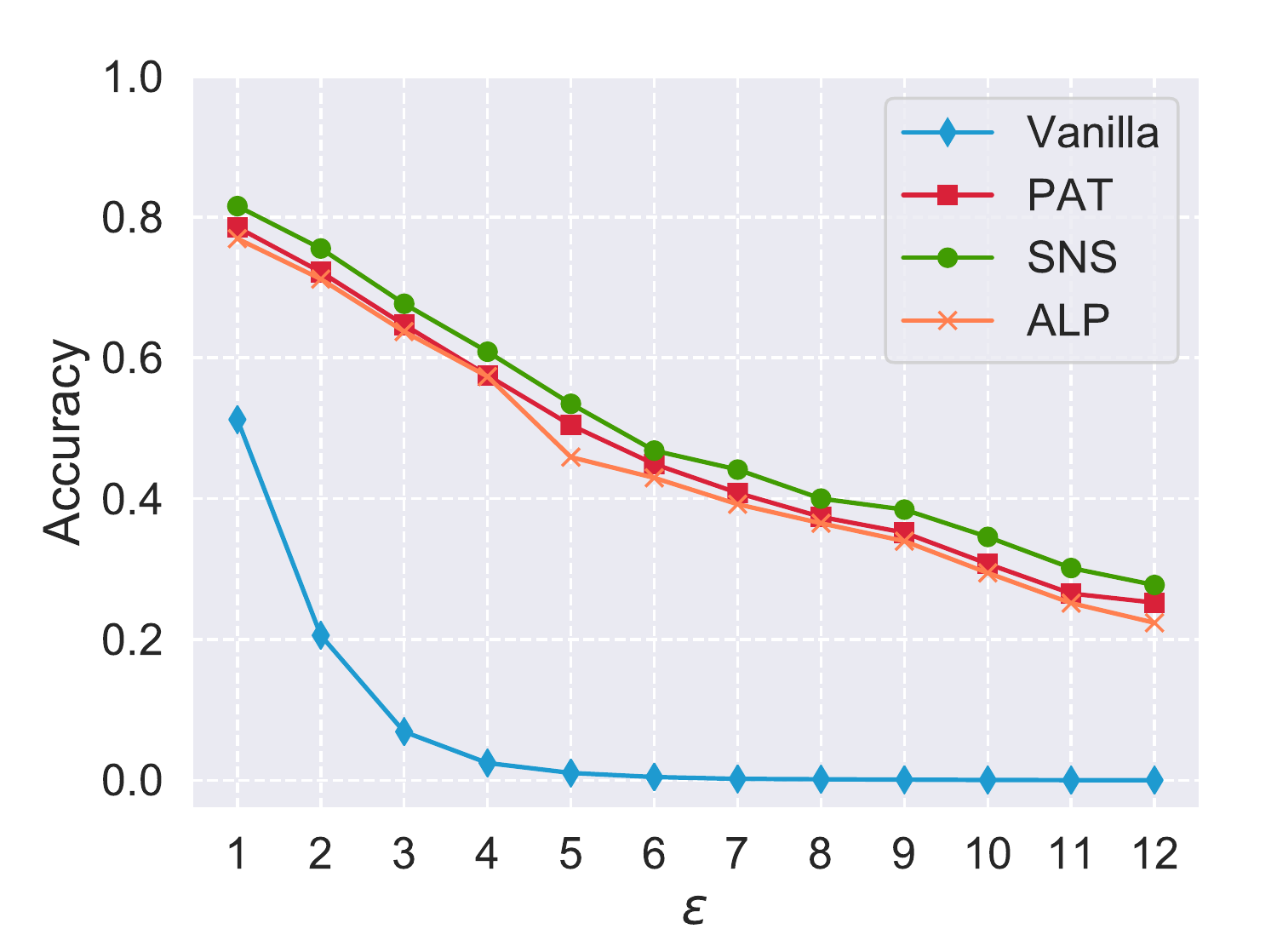}
}
\caption{\textcolor{black}{Experiment results of $\ell_\infty$ PGD attack with different parameters on CIFAR-10 using VGG-16 model. Subfigure (a) to (b) respectively represent model robust accuracy against $\ell_\infty$ PGD with different iteration $k$ and varying perturbation magnitude $\epsilon$.}}
\label{fig:PGD_parameter}
\end{figure}

\begin{figure}[!htb]
\centering
\subfigure[]{
\includegraphics[width=0.46\linewidth]{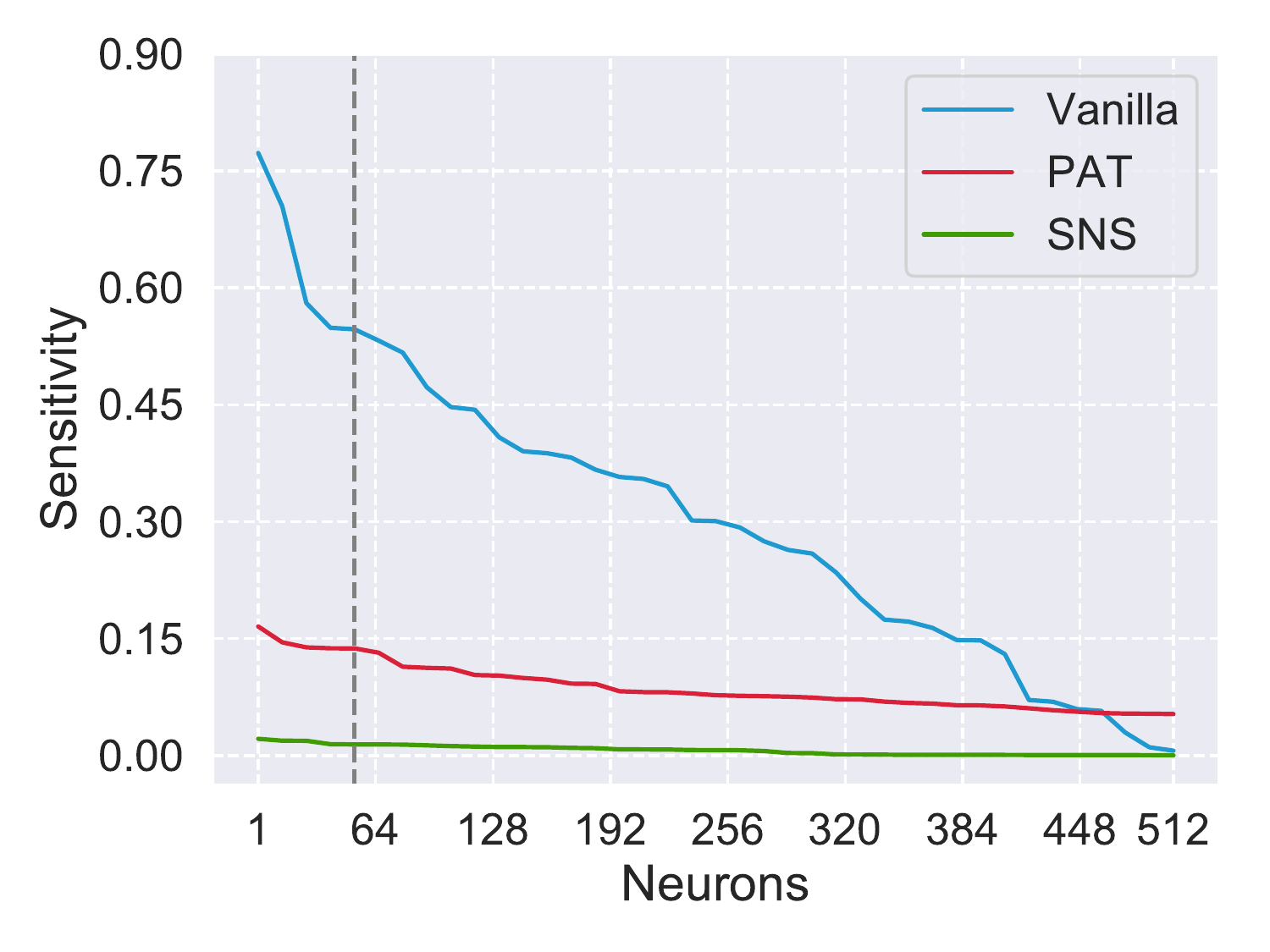}
}
\subfigure[]{
\includegraphics[width=0.46\linewidth]{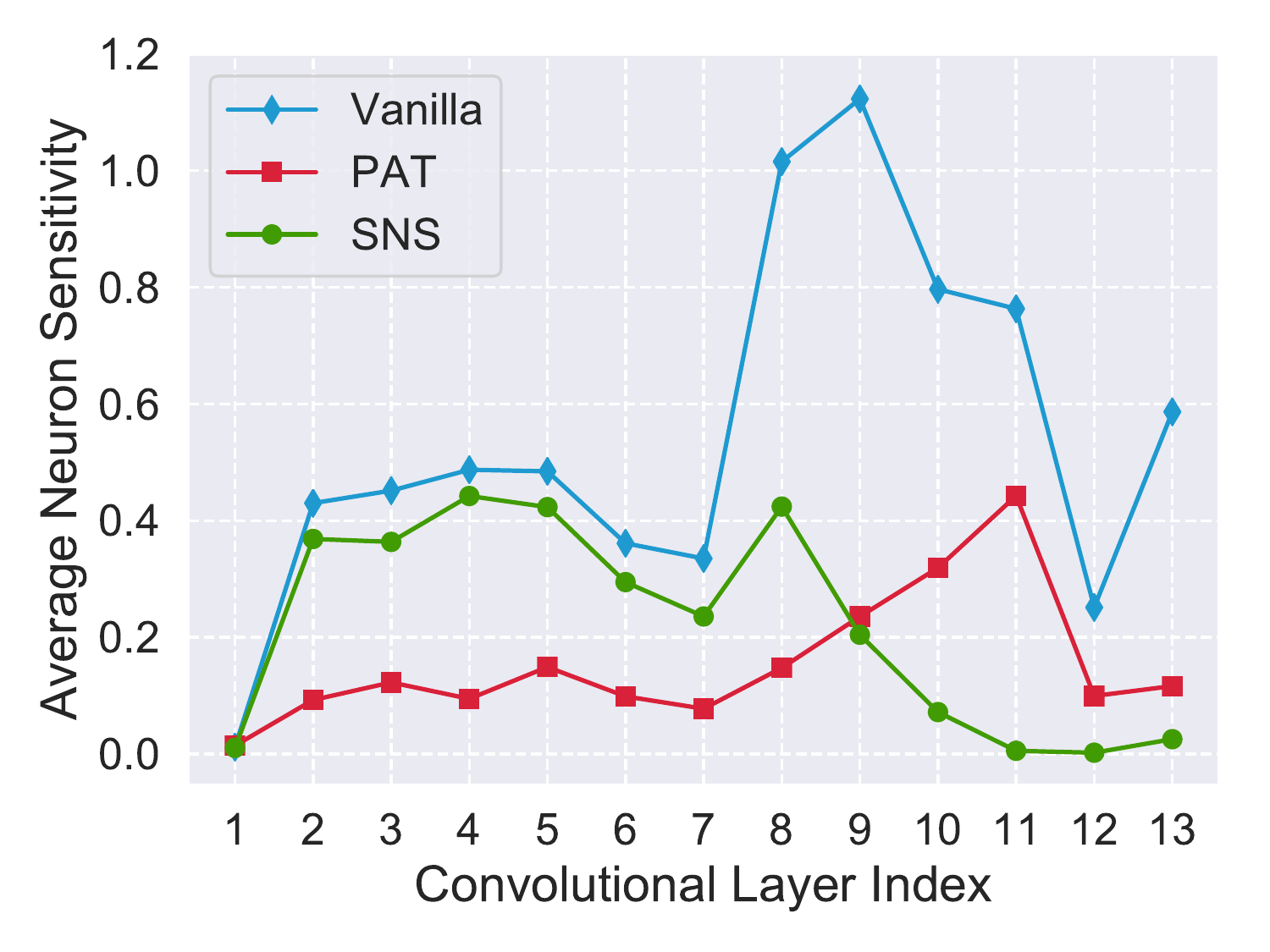}
}
\caption{Neuron sensitivity under PGD attack of sensitive neurons with Vanilla, PAT and SNS$_{sen}^{adv}$ trained VGG-16 models on CIFAR-10. Subfigure (a) to (b) represent neuron sensitivities of all neurons on \emph{conv10} layer and the mean values of neuron sensitivities for all convolutional layers, respectively. Our method dramatically decreases the neuron sensitivity and achieves better results than PAT.}
\label{fig:Sensitivity after finetune revision}
\end{figure}

We try to test model robustness by using our proposed SNS on CIFAR-10 with VGG-16 first. \draft{To fully demonstrate the adversarial defense ability of each trained model, we respectively adopt white-box and black-box attacks in this part. White-box attacks get full access to the models and thus generally provide convincing results of models' adversarial robustness. Here we adopt FGSM, C\&W, $\ell_2$ PGD and $\ell_\infty$ PGD. Moreover, SPSA and NAttack are used as black-box attacks since they have the ability to further examine whether obfuscated gradient is introduced of all models according to \cite{athalye2018obfuscated,DBLP:conf/icml/UesatoOKO18}.} We train model with the loss in Equation \ref{finetune_adv} using top-10\% sensitive neurons. According to \cite{DBLP:journals/corr/abs-1803-06373}, ALP is also implemented on the basis of PGD-based adversarial training and achieves good performance. Thus we choose this method as a contrast.

\draft{According to the result (SNS$_{sen}^{adv}$) in Table \ref{table:finetune vanilla}, our method outperforms all other comparison methods including PAT and ALP, indicating that SNS builds strong models towards adversarial examples.} Meanwhile, ALP also shows weak adversarial defense ability in most cases compared with PAT. These observations prove the importance of using sensitive neurons for improving model robustness compared with logit (ALP). Furthermore, stabilizing sensitive neurons also serves as a regularization term when used with adversarial training loss to alleviate clean example accuracy drops, since most adversarial training strategies build models with relatively low clean example accuracy. The corresponding experimental results on ImageNet with ResNet-18 are shown in Table \ref{table:finetune ResNet}. \draft{It is worth noting that SNS keeps behaving well towards black-box attacks (e.g., SPSA and NAttack) on different datasets, which prove that our SNS model does not achieve security via obscurity and provides true benefits in adversarial robustness.}

\textcolor{black}{To fully evaluate the model robustness, we further conduct an experiment with $\ell_\infty$ PGD attack using different parameters (i.e., $k$ and $\epsilon$) on CIFAR-10 using a VGG-16 model. In particular, we test model accuracy against $\ell_\infty$ PGD attack with different iteration numbers $k$ and perturbation magnitudes $\epsilon$. We first apply $\ell_\infty$ PGD attack using iteration $k=1,...,10$ with fixed $\epsilon=8$ and then perform attacks with $\epsilon=1,...,12$ and fixed the $k=10$. As plotted in Figure \ref{fig:PGD_parameter}, SNS model constantly outperforms PAT and ALP models in adversarial accuracy, which further confirms the superiority of our proposed method.}

\textcolor{black}{By simply computing and extracting sensitive neurons from a vanilla model with fixed parameters, we could achieve considerable improvements in adversarial robustness with or without PAT (as the white-box \draft{and black-box} attack results shown in Table \ref{table:finetune vanilla} and Table \ref{table:finetune ResNet}). However, during training, the sensitive neurons for a specified model might be different. Thus, it's intuitive for us to further conduct an experiment by training models with sensitive neurons updated dynamically during training. Specifically, we update the sensitive neurons at the end of each epoch based on 200 randomly chosen samples. We respectively use SNS$_{sen}^{adv}$ and SNS$_{dyn}^{adv}$ to denote the original SNS model and SNS model with dynamic sensitive neurons. As shown in Table \ref{table:finetune vanilla}, the performance of SNS$_{sen}^{adv}$ and SNS$_{dyn}^{adv}$ are quite similar. The results revealed the effectiveness of using sensitive neurons from vanilla models with fixed parameters to perform SNS training. Though updating sensitive neurons dynamically improves model robustness, it introduce more extra computation cost than original SNS if sensitive neurons are dynamically updated during each training iteration. In our future work, we will concentrate on this challenge and try to propose a better way to reduce the computation cost.}

Since our model SNS$_{sen}^{adv}$ reaches the best performance of adversarial robustness, we make further analyses on how the neuron sensitivity changes after SNS training. Figure \ref{fig:Sensitivity after finetune revision} (a) shows the neuron sensitivity of all neurons in \emph{conv10} layer after SNS training. It is obvious that our method dramatically suppresses the neuron sensitivity and achieves better results than PAT, which demonstrates that the SNS loss term significantly contributes to better adversarial robustness. Further, our method requires much fewer epochs compared to PAT, leading to less time consumption. As illustrated in Figure \ref{fig:Sensitivity after finetune revision} (b), interestingly, the sensitivities of neurons on layer \emph{conv8} to \emph{conv13}  have a significant change, which exactly match layers we use in SNS to train the model. Meanwhile, the neuron sensitivities in bottom layers stay high since they were fixed and only \emph{conv8} to \emph{conv13} were finetuned during training.

\textcolor{black}{We also conduct an ablation study to demonstrate the effectiveness of sensitive neurons by randomly selecting neurons or choosing all neurons and then stabilize them. We respectively train VGG-16 models by Equation \ref{finetune_adv} using all neurons and randomly selected neurons (denoted by SNS$_{all}^{adv}$ and SNS$_{rand}^{adv}$ respectively). For the strategy of random neuron selection, we randomly select 10\% of neurons in each layer and repeat for three times to show the
mean value. As shown in Table \ref{table:finetune vanilla}, the adversarial robustness of the SNS$_{all}^{adv}$ model decreases to some extent, which means that sensitive neurons are more critical for adversarial robustness. Constraining all neurons may somewhat lead to the drop of adversarial robustness. Meanwhile, SNS$_{rand}^{adv}$ behaves even worse than the original PAT model, which shows the negative effect of selecting random neurons for neurons stabilizing to adversarial robustness. The reason might be that using vanilla neurons will introduce meaningless gradients, which is harmful to the model robustness.}

\section{Conclusion} \label{Section:conclusion}
This paper tries to interpret and improve adversarial robustness for deep models from a new perspective of neuron sensitivity, which is measured by neuron behavior variation intensity against benign and adversarial examples. We first draw the close relationship between adversarial robustness and neuron sensitivities, as sensitive neurons make the most non-trivial contribution to model predictions in the adversarial setting. Based on that conclusion, we further propose to improve model robustness against adversarial examples by stabilizing the sensitive neurons, which constrains the behaviors of sensitive neurons between benign and adversarial examples. Moreover, our experimental results reveal that state-of-the-art adversarial training strategies achieve strong robustness by reducing neuron sensitivities, which in turn confirms the importance of sensitive neurons to the adversarial robustness. Extensive experiments on various datasets demonstrate that our algorithm effectively achieves excellent results.

Currently, our SNS method uses the same coefficient $\lambda$ to combine sensitivities in different layers together. However, since each layer has a different contribution towards the model robustness, it is better for us to use adaptive coefficients that consider heterogeneous behaviors of different layers. Thus, it would make full use of the efficacy of each intermediate layer and build stronger models. Meanwhile, the layer selection method we adopt in SNS is a naive top-$k$ layers choosing procedure. We will develop a better strategy that adapts to various model architectures in future work.

%\section{Acknowledgement} \label{Section:ack}
%This work was supported by The National Key Research and Development Plan of China (2020AAA0103502), and National Natural Science Foundation of China (62022009 and 61872021).

\bibliographystyle{IEEEtran}
\bibliography{ref}

\begin{IEEEbiography}[{\includegraphics[width=1in,height=1.25in,clip,keepaspectratio]{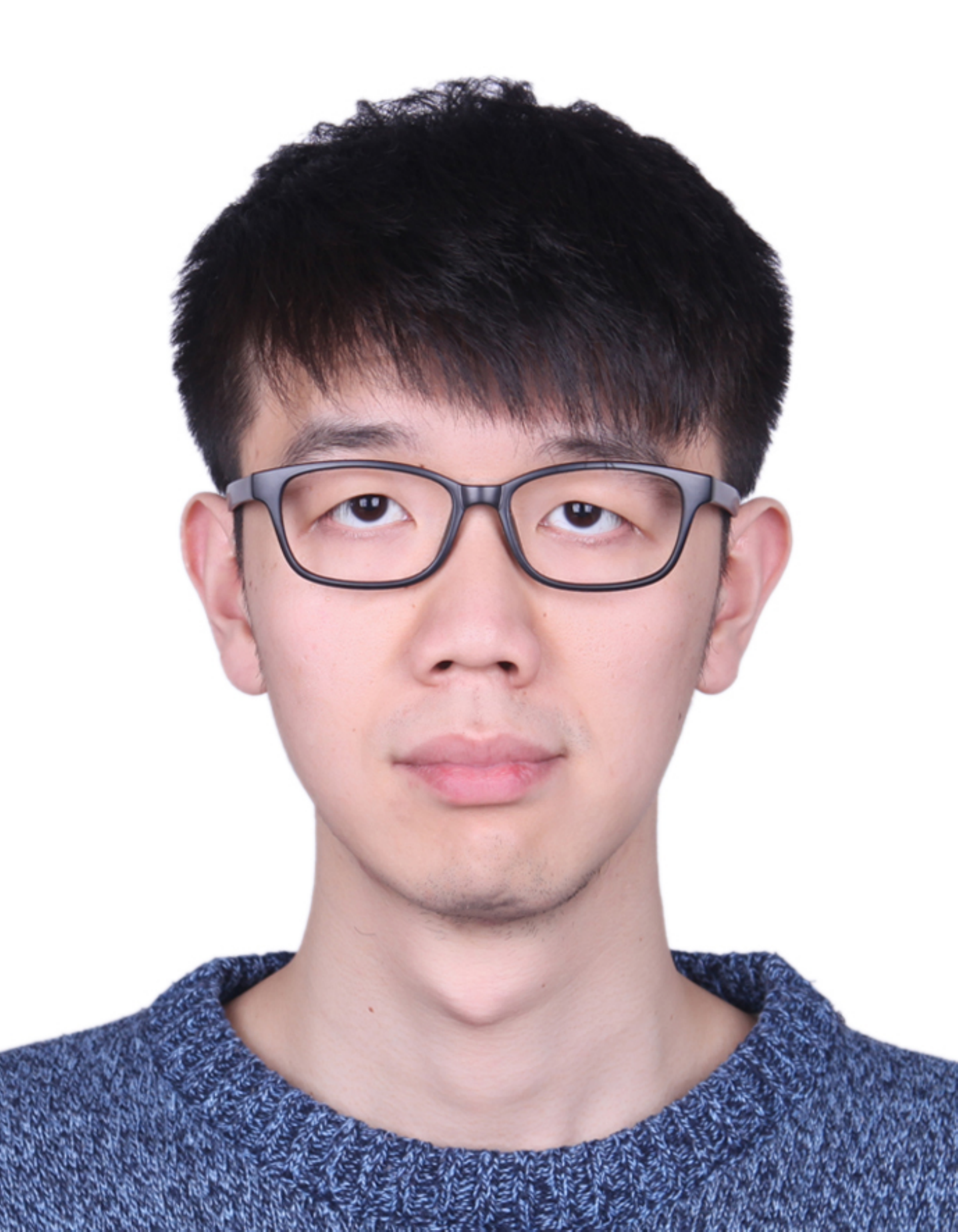}}]{Chongzhi Zhang}
received his BS in 2019 in computer science from Beihang University. He is currently working toward the Master degree at the school of Computer Science and Engineering, Beihang University. His current research interests include adversarial examples, object detection, multi-camera tracking, visual relationship detection and domain adaptation.
\end{IEEEbiography}

\begin{IEEEbiography}[{\includegraphics[width=1in,height=1.25in,clip,keepaspectratio]{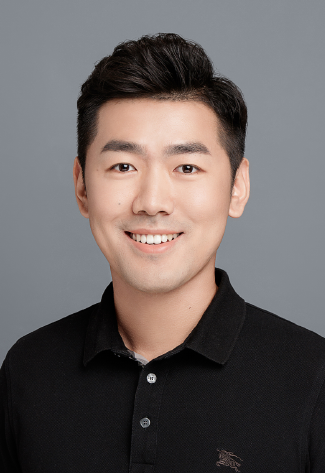}}]{Aishan Liu}
received his BS and MS in 2013 and 2016 in computer science from Beihang University. He is currently working toward the Ph.D. degree at the school of Computer Science and Engineering, Beihang University. His current research interests include adversarial examples and robust deep learning models.
\end{IEEEbiography}

\begin{IEEEbiography}[{\includegraphics[width=1in,height=1.25in,clip,keepaspectratio]{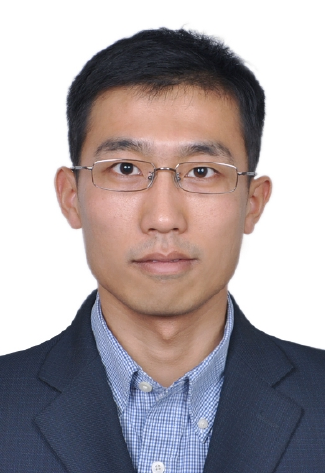}}]{Xianglong Liu}
received the BS and Ph.D degrees in computer science from Beihang University, Beijing, China, in 2008 and 2014. From 2011 to 2012, he visited the Digital Video and Multimedia (DVMM) Lab, Columbia University as a joint Ph.D student. He is currently an Associate Professor with the School of Computer Science and Engineering, Beihang University. He has published over 40 research papers at top venues like the IEEE TRANSACTIONS ON IMAGE PROCESSING, the IEEE TRANSACTIONS ON CYBERNETICS, the Conference on Computer Vision and Pattern Recognition, the International Conference on Computer Vision, and the Association for the Advancement of Artificial Intelligence. His research interests include machine learning, computer vision and multimedia information retrieval.
\end{IEEEbiography}

\begin{IEEEbiography}[{\includegraphics[width=1in,height=1.25in,clip,keepaspectratio]{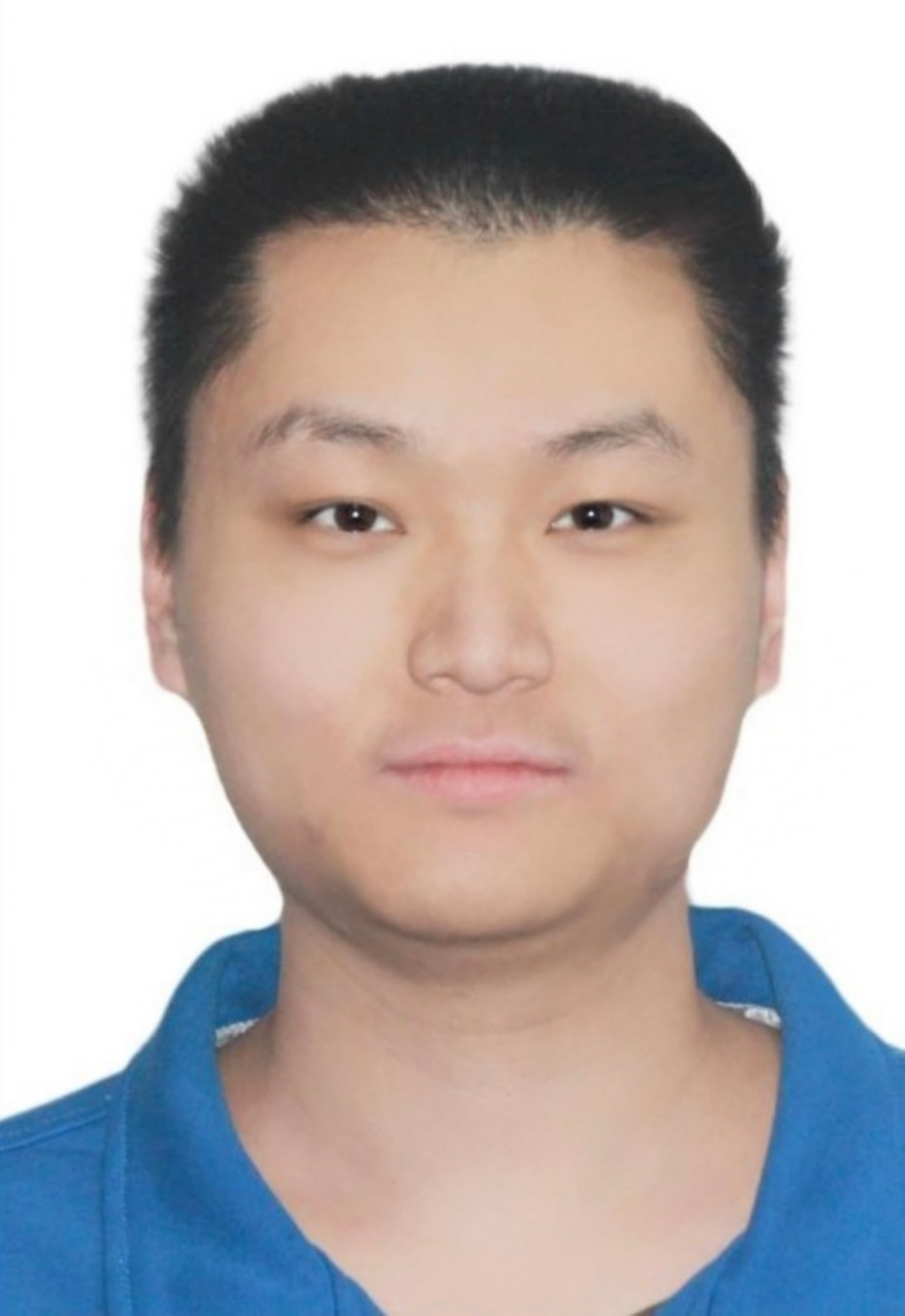}}]{Yitao Xu}
is currently a senior student in Beihang University at the School of Computer Science and Engieering. He is working towards his B.Eng, majoring in Computer Science and Technology. His research interests include deep learning interpretation, adversarial examples and cognitive science.
\end{IEEEbiography}

\begin{IEEEbiography}[{\includegraphics[width=1in,height=1.25in,clip,keepaspectratio]{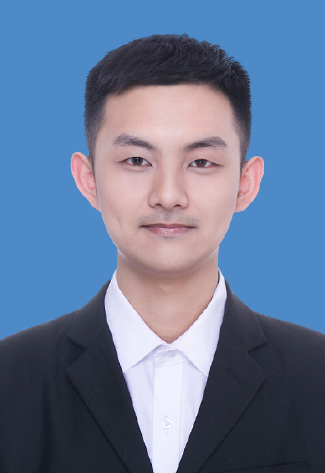}}]{Hang Yu}
received the BS degree in Tang Ao-qing honors program in computer science from Jilin University. He is currently a Master candidate at the School of Computer Science and Engineering, Beihang University. His current research interests include adversarial examples, statistical deep learning and computer vision.
\end{IEEEbiography}

\begin{IEEEbiography}[{\includegraphics[width=1in,height=1.25in,clip,keepaspectratio]{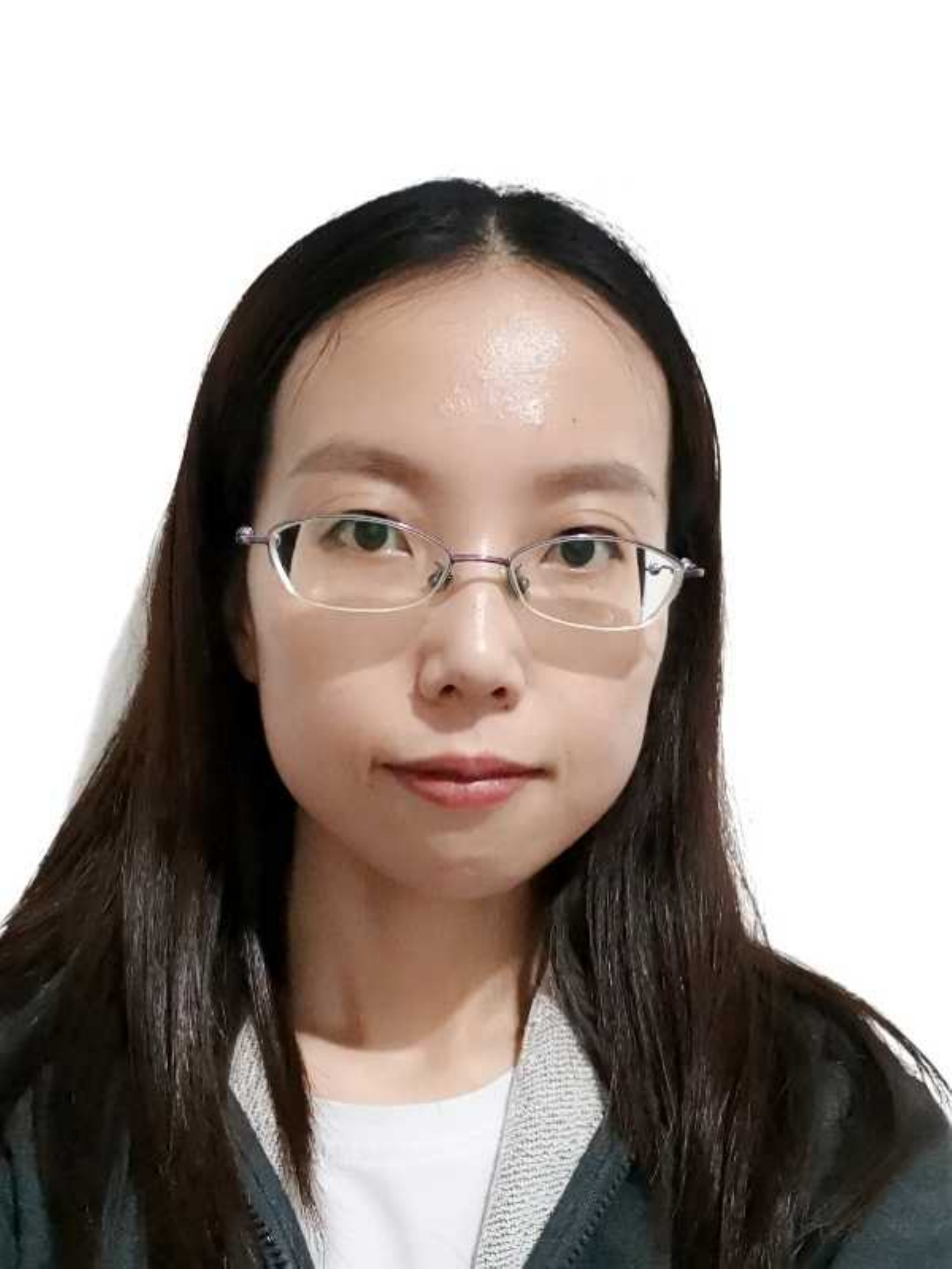}}]{Yuqing Ma}
received the BS in 2015 from Shandong University, China. She is currently working toward the Ph.D. degree at the school of Computer Science and Engineering, Beihang University. Her current research interests include computer vision, generative models, and few shot learning.
\end{IEEEbiography}

\begin{IEEEbiography}[{\includegraphics[width=1in,height=1.25in,clip,keepaspectratio]{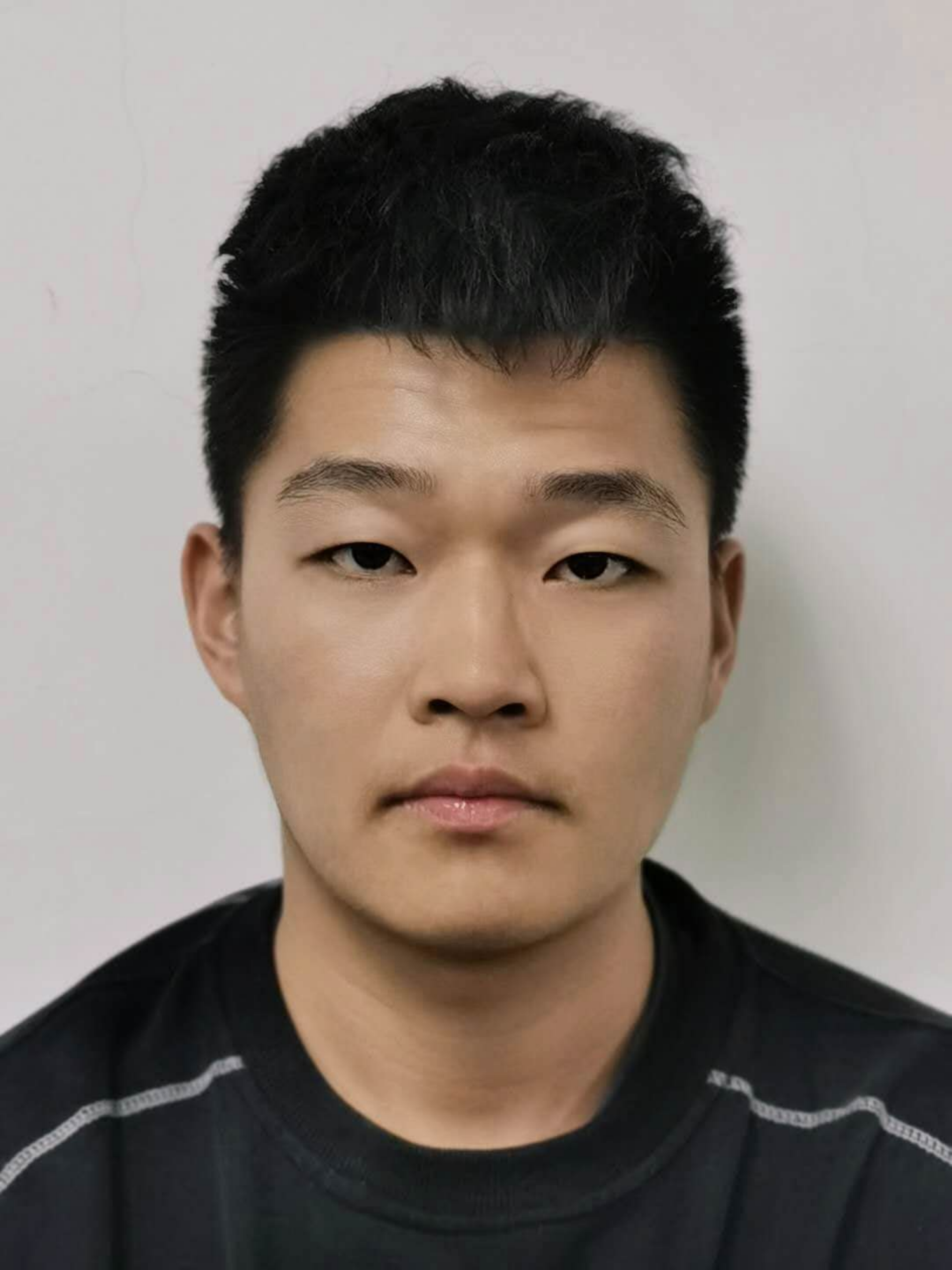}}]{Tianlin Li}
received his B.Eng  and M.Eng from Beihang University. He is pursuing a PHD degree. His current research interests include AI security and deep learning interpretation, especially modeling neural networks for explaining.
\end{IEEEbiography}

\end{document}